\documentclass[11pt]{article}

\usepackage{amsmath,amssymb,amsthm}
\usepackage{mathtools}
\usepackage{graphicx}
\usepackage{booktabs}
\usepackage{subcaption}
\usepackage{changepage}
\usepackage{placeins}
\usepackage{needspace}

\usepackage{algorithm}
\usepackage{algorithmic}

\usepackage{tikz}
\usetikzlibrary{arrows.meta,calc,positioning}

\usepackage{arxiv}
\hypersetup{
  pdftitle={Soft Fitted Q-Iteration without Bellman Completeness: Occupancy Reweighting and Temperature Annealing},
  pdfauthor={Lars van der Laan and Nathan Kallus}
}

\DeclareMathOperator*{\argmin}{arg\,min}

\DeclareMathOperator{\KL}{KL}
\DeclareMathOperator{\Var}{Var}

\theoremstyle{plain}
\newtheorem{theorem}{Theorem}
\newtheorem{lemma}{Lemma}

\newcommand{\E}{\mathbb{E}}

\newcommand{\CartPoleControlPrimaryGain}{5.030}
\newcommand{\CartPoleControlPrimaryCILower}{3.979}
\newcommand{\CartPoleControlPrimaryCIUpper}{6.137}
\newcommand{\CartPoleControlPrimaryWins}{29/30}

\newcommand{\CartPoleControlMedianESS}{0.779}
\newcommand{\CartPoleControlMedianMaxWeight}{2.901}

\newcommand{\CartPoleSensitivityKEightyTfiveGain}{5.145}
\newcommand{\CartPoleSensitivityKEightyTfiveCILower}{4.583}
\newcommand{\CartPoleSensitivityKEightyTfiveCIUpper}{5.699}

\newcommand{\CartPoleSensitivityKEightyTfiveSoftGain}{4.801}
\newcommand{\CartPoleSensitivityKEightyTfiveSoftCILower}{4.233}
\newcommand{\CartPoleSensitivityKEightyTfiveSoftCIUpper}{5.362}

\newcommand{\CartPoleSensitivityHardOneGain}{4.287}

\newcommand{\CartPoleSensitivityHardHundredGain}{6.310}

\title{Soft Fitted Q-Iteration without Bellman Completeness:\\
Occupancy Reweighting and Temperature Annealing}

\author{%
  Lars van der Laan \\
  Department of Statistics, University of Washington \\
  \texttt{lvdlaan@uw.edu} \\
  \And
  Nathan Kallus \\
  Netflix and Cornell University
}

\begin{document}

\maketitle

\begin{abstract}
Fitted \(Q\)-iteration (FQI) is a standard regression-based method for optimal
control in offline reinforcement learning, but its stability under function
approximation often relies on Bellman completeness, which requires Bellman
images of the fitted class to remain in the class. We study Kullback--Leibler
(KL)-regularized, or soft, FQI relative to a fixed reference policy without
this assumption. Our
key insight is that soft control locally inherits the contraction of policy
evaluation in a discounted-occupancy norm. At the soft-optimal fixed point,
the linearization of the soft Bellman operator is exactly the Bellman operator
for the soft-optimal policy, which contracts in its discounted-occupancy norm;
projection in the same norm preserves this contraction. Standard soft FQI
instead projects under the offline state-action distribution and need not
preserve this property. Motivated by this observation, we propose \emph{occupancy-reweighted soft
FQI}, which retains standard Bellman targets and least-squares updates while
reweighting regressions by discounted-occupancy ratios induced by the current
soft policy. Under \(Q\)-function realizability and local regularity, we
establish local contraction and finite-sample convergence with estimated
ratios, without Bellman completeness. We then use temperature annealing to
convert the local result into global convergence from arbitrary
initialization: sufficiently high temperature provides a globally contractive
starting regime, while gradual cooling connects successive local contraction
regions to any prescribed positive target temperature. Under an action-gap
margin condition, switching at a fixed positive temperature to hard FQI with
refreshed occupancy weights also yields population and finite-sample
convergence to the unregularized optimum. Controlled experiments isolate the
roles of occupancy reweighting and temperature annealing. 
\end{abstract}

\section{Introduction}

Fitted \(Q\)-iteration (FQI) is a standard regression-based method for
learning an optimal policy through value-based control in reinforcement
learning. Each iteration forms bootstrapped Bellman targets from the current
\(Q\)-function and regresses them onto a fitted function class, making FQI
simple to implement and compatible with flexible supervised-learning methods
\citep{ernst2005tree,munos2005error,lazaric2012finite,
voloshin2019empirical,fujimoto2019off,le2019batch,agarwal2021deep}.
Entropy regularization smooths the Bellman optimality operator by replacing
the hard maximum with a log-sum-exp, inducing a softmax policy. More generally,
KL regularization relative to a fixed reference policy yields a
reference-weighted log-sum-exp and corresponding reference-weighted softmax
policy; the uniform reference recovers standard entropy regularization
\citep{ziebart2008maximum,ziebart2010causal,haarnoja2017reinforcement,
haarnoja2018soft}, while a behavior-policy reference encourages the learned
policy to remain close to the data-generating policy, mitigating distribution
shift and coverage requirements. The resulting \emph{soft} FQI preserves the
regression structure of FQI while replacing the nonsmooth optimality
operator with a differentiable one.

Differentiability, however, does not by itself stabilize fitted Bellman
iteration. At the population level, soft FQI applies the soft Bellman operator
and projects the resulting target onto the fitted class in the \(L^2\) norm of
the offline state-action distribution. Although the soft Bellman operator is a
\(\gamma\)-contraction in the supremum norm, projection in the offline-data
norm need not preserve this contraction. Contractivity, and hence geometric
convergence, is commonly recovered through \emph{Bellman completeness}, which
requires Bellman images of the fitted class to remain in the class. Projection
is then inactive on Bellman images, so fitted iteration inherits the stability
of exact soft value iteration. Related analyses replace exact completeness
with stability conditions on the projected Bellman operator or approximate
completeness conditions that control the \emph{inherent Bellman error}, namely,
how far Bellman images of functions in the fitted class lie outside that class
\citep{gordon1995stable,tsitsiklis1996analysis,munos2008finite,
farahmand2010error,scherrer2014approximate,fan2020theoretical}.

Bellman completeness is a particularly strong approximation requirement for
FQI. Unlike ordinary approximation error, it is non-monotone in the fitted
class: enlarging the class can improve approximation of the optimal
\(Q\)-function while worsening completeness, because the Bellman images of all
newly added functions must also be represented
\citep{zhan2022offline,zanette2022bellman}. Thus, richer function classes
increase statistical complexity without necessarily reducing inherent Bellman
error. Moreover, realizability and coverage alone do not generally ensure
stable or sample-efficient offline value learning
\citep{chen2019information,foster2021offline,amortila2020variant,
wang2021exponential}. Minimax and adversarial \(Q\)-learning provide one
alternative by replacing successive fitted Bellman regressions with
saddle-point problems over auxiliary critic or test-function classes
\citep{dai2018sbeed,uehara2020minimax,uehara2021finite,
jin2021bellman,xie2021batch,uehara2023offline}. For soft control, such methods
can avoid Bellman completeness under realizability of the soft \(Q\)-function
and a corresponding dual function together with partial coverage
\citep{uehara2023offline}, but require solving an auxiliary adversarial
problem.

A complementary perspective comes from off-policy evaluation. For a fixed
target policy, the Bellman evaluation operator is a \(\gamma\)-contraction in
the \(L^2\) norm of the target policy's stationary state-action distribution,
a fact underlying classical projected temporal-difference analysis
\citep{tsitsiklis1996analysis}. Stationary-distribution corrections reweight
off-policy data toward this target-policy distribution, restoring contraction
in the target-policy norm
\citep{hallak2017consistent,geladaBellemare2019CovariateShift}, while
occupancy-weighted fitted \(Q\)-evaluation applies the same principle to
successive fitted Bellman updates \citep{vdLaanKallus2025FQE}. This suggests a
different route for control: rather than impose Bellman completeness or replace
fitted regression by a minimax problem, adjust the regression norm to
preserve the contraction structure of the Bellman update. The challenge is
that, in control, the Bellman operator is nonlinear and the soft-optimal
occupancy is unknown.

Our key insight is that soft control locally inherits the contraction of
policy evaluation. We show that, at the soft-optimal fixed point
\(Q_\tau^\star\), the linearization of the soft Bellman operator is exactly
the Bellman operator for the soft-optimal policy. In its discounted-occupancy
norm, this operator is contractive, and projection in the same norm preserves
the contraction. We therefore obtain local contraction of the projected soft
Bellman update around \(Q_\tau^\star\). Standard soft FQI need not inherit
this property because it instead projects under the offline state-action
distribution. The size of the contraction region depends on temperature: as
\(\tau\) increases, the induced soft policy varies more slowly with \(Q\),
reducing the nonlinear remainder in the first-order approximation and
extending the region over which the local contraction persists.

Motivated by this norm choice, we propose \emph{occupancy-reweighted soft
FQI}. The method changes only the weights in each least-squares Bellman
update, leaving the targets and regression objective otherwise unchanged. At
each temperature, it estimates the discounted-occupancy ratio induced by the
current initializer and holds these weights fixed across the subsequent
regressions at that temperature. When the initializer is sufficiently close
to \(Q_\tau^\star\), the resulting occupancy is close enough to the
soft-optimal occupancy for the local contraction to persist.

Local contraction alone does not guarantee convergence from arbitrary
initialization. We show that, at sufficiently high temperature, the
population update satisfies a contraction bound over the entire fitted class.
Starting in this regime and gradually annealing the temperature allows the
iterates to track the soft-optimal solutions through successive local
contraction regions, yielding convergence from arbitrary initialization to
any prescribed positive target temperature. For unregularized control, we
stop annealing at a fixed positive switch temperature and use the resulting
iterate to initialize hard FQI, refreshing the occupancy weights after each
hard update.

\textbf{Contributions.}
Our main contributions are as follows.
\begin{enumerate}[leftmargin=1.5em,itemsep=2pt,topsep=2pt]

\item \textbf{Local contraction without Bellman completeness.}
We show that projected soft FQI is locally contractive in the
soft-optimal discounted-occupancy norm under \(Q\)-function realizability and
mild regularity, without requiring Bellman completeness.

\item \textbf{Occupancy-reweighted soft FQI.}
We propose an offline fitted \(Q\)-iteration method that reweights
least-squares updates by discounted occupancy ratios. At each temperature, a
single ratio estimate is reused across multiple regression updates.

\item \textbf{Global convergence by annealing and hard continuation.}
We show that sufficiently high temperature removes the local initialization
requirement and that gradual temperature annealing yields convergence from
arbitrary initialization to any prescribed positive target temperature.
Realizability is required only at the target for exact convergence. Under an
action-gap margin condition, switching at a positive temperature to hard FQI
yields convergence to the unregularized optimum.

\item \textbf{Finite-sample guarantees.}
We establish finite-sample guarantees for annealed occupancy-reweighted soft
FQI with estimated occupancy ratios, including the optional hard continuation.
The bounds separate optimization, regression, ratio-estimation, and
approximation errors.

\end{enumerate}

\subsection{Related Work}

\textbf{Fitted value iteration and FQI.}
Classical analyses of fitted value iteration and FQI obtain stability under
additional structure linking the Bellman operator, fitted class, and sampling
distribution, such as Bellman completeness, small inherent Bellman error,
concentrability, or structured representations
\citep{gordon1995stable,tsitsiklis1996analysis,munos2005error,
munos2008finite,farahmand2010error,scherrer2014approximate,
chen2019information,amortila2020variant,wang2021exponential,
wang2021statistical,chang2022learning}.
Difficulties persist even under linear approximation: recent work establishes
boundedness or convergence to sets, while realizable problems can admit
multiple projected Bellman solutions
\citep{meyn2024projected,liu2025linear,mehta2026optimistic}.
Our approach instead changes the projection norm: occupancy reweighting yields
local contraction around the soft-optimal solution without Bellman
completeness, while temperature annealing extends this result to convergence
from arbitrary initialization.
Appendix~\ref{app:related-work} gives an extended discussion and additional
references.

\textbf{Minimax and adversarial \(Q\)-learning.}
Minimax methods avoid Bellman completeness by testing Bellman residuals against
auxiliary critic or test-function classes
\citep{dai2018sbeed,uehara2020minimax,uehara2021finite,
jin2021bellman,xie2021batch,uehara2023offline}.
They can obtain guarantees under \(Q\)-function realizability, including under
partial coverage for soft control \citep{uehara2023offline}, but introduce an
auxiliary function class and saddle-point optimization.
Occupancy-reweighted soft FQI instead retains standard least-squares Bellman
regressions and obtains stability by changing the projection norm.

\textbf{Distribution correction and occupancy weighting.}
Stationary and emphatic weighting have long been used to stabilize off-policy
policy evaluation, while occupancy-ratio methods such as DualDICE and FORE
estimate related distribution corrections
\citep{mahmood2015emphatic,sutton2016emphatic,yu2015convergence,
hallak2017consistent,geladaBellemare2019CovariateShift,
nachum2019dualdice,zhang2020gendice,van2026fitted}.
For fitted policy evaluation, the closest work is occupancy-weighted FQE,
which reweights each Bellman regression by the occupancy ratio of a fixed
evaluation policy and obtains global contraction in that policy's occupancy
norm \citep{vdLaanKallus2025FQE}. The closest control precedent is
\citet{sinha2022experience}, who use estimates of the current-policy occupancy
ratio to reweight critic updates in an actor--critic algorithm, motivated by
contraction in fitted policy evaluation.

These policy-evaluation arguments do not establish convergence for control,
where the policy and its occupancy change with \(Q\). We extend the
occupancy-weighting principle to soft control by establishing local
contraction around the soft-optimal solution and using temperature annealing
to obtain convergence from arbitrary initialization. KL regularization
controls the size of the local contraction region
\citep{ziebart2008maximum,neu2017unified,geist2019theory,
haarnoja2018soft}; this role of temperature differs from the bias--variance
tuning studied by \citet{uehara2023offline}.

\section{Preliminaries}

\subsection{Soft Control with a Fixed Reference Policy}

Consider a discounted Markov decision process (MDP) with standard Borel state
space \(\mathcal S\), finite action space \(\mathcal A\) with
\(|\mathcal A|\ge2\), transition kernel \(P\), discount factor
\(\gamma\in(0,1)\), and one-step reward \(R\) satisfying
\(|R|\le B_R<\infty\) almost surely. Write
\(\bar r(s,a):=\E[R\mid S=s,A=a]\) for the conditional mean reward.
We study KL-regularized control relative to a known reference policy
\(\pi_{\mathrm{ref}}\), with temperature \(\tau>0\). Larger values of
\(\tau\) penalize deviations from the reference policy more strongly, whereas
the limit \(\tau\downarrow0\) recovers ordinary optimal control.

Assume the reference policy is uniformly positive:
\[
L_{\mathrm{ref}}
:=
\sup_{(s,a)\in\mathcal S\times\mathcal A}
\log\{1/\pi_{\mathrm{ref}}(a\mid s)\}<\infty.
\]
Let \(\nu_b\) denote the offline state-action distribution on
\(\mathcal S\times\mathcal A\), and write
\(\nu_b=\nu_0\otimes\pi_b\), where \(\nu_0\) is its state marginal and
\(\pi_b\) is the associated behavior policy.
Define
\(\nu_{\mathrm{ref}}:=\nu_0\otimes\pi_{\mathrm{ref}}\), the state-action
distribution obtained by drawing \(S\sim\nu_0\) and then
\(A\sim\pi_{\mathrm{ref}}(\cdot\mid S)\).
A uniform reference policy yields standard entropy regularization up to an
additive constant, with \(L_{\mathrm{ref}}=\log|\mathcal A|\).

For any \(Q\)-function \(Q\), define the induced soft policy
\[
\pi_{\tau,Q}(a\mid s)
:=
\frac{
\pi_{\mathrm{ref}}(a\mid s)e^{Q(s,a)/\tau}
}{
\sum_{a'\in\mathcal A}
\pi_{\mathrm{ref}}(a'\mid s)e^{Q(s,a')/\tau}
},
\qquad
a\in\mathcal A,
\]
and the reference-weighted soft state value
\[
V_{\tau,Q}(s)
:=
\tau\log
\sum_{a\in\mathcal A}
\pi_{\mathrm{ref}}(a\mid s)e^{Q(s,a)/\tau}.
\]
The soft Bellman optimality operator is
\[
(\mathcal T_\tau Q)(s,a)
:=
\bar r(s,a)
+\gamma\E_{S'\sim P(\cdot\mid s,a)}[V_{\tau,Q}(S')].
\]
Let \(Q_\tau^\star\) denote the unique bounded solution to
\[
Q_\tau^\star=\mathcal T_\tau Q_\tau^\star,
\]
the \emph{soft-optimal \(Q\)-function}, and let
\(\pi_\tau^\star:=\pi_{\tau,Q_\tau^\star}\) denote the associated
soft-optimal policy.

For a fixed policy \(\pi\), write
\[
(P_\pi f)(s,a)
:=
\E\!\left[
f(S',A')
\,\middle|\,
S=s,A=a,\;
S'\sim P(\cdot\mid s,a),\;
A'\sim\pi(\cdot\mid S')
\right].
\]
For later use, let \(\mu_{\tau,Q}\) denote the normalized
discounted occupancy under \(\pi_{\tau,Q}\) with state-action restart
distribution \(\nu_{\mathrm{ref}}\), characterized by
\begin{equation}
\mu_{\tau,Q}
=
(1-\gamma)\nu_{\mathrm{ref}}
+
\gamma\mu_{\tau,Q}P_{\pi_{\tau,Q}}.
\label{eq:stage-frozen-occupancy}
\end{equation}
Thus, at a restart the state is drawn from \(\nu_0\) and the action from
\(\pi_{\mathrm{ref}}\), while all subsequent actions follow
\(\pi_{\tau,Q}\). Write
\[
\mu_\tau^\star
:=
\mu_{\tau,Q_\tau^\star}.
\]
Whenever \(\mu_{\tau,Q}\ll\nu_b\), its occupancy ratio relative to the
offline distribution is \(\mathrm d\mu_{\tau,Q}/\mathrm d\nu_b\).

For a fixed policy \(\pi\), define its KL-regularized Bellman evaluation
operator by
\[
(\mathcal T_\tau^\pi f)(s,a)
:=
\bar r(s,a)
+\gamma(P_\pi f)(s,a)
-\gamma\tau
\E_{S'\sim P(\cdot\mid s,a)}
\left[
\KL\!\bigl(
\pi(\cdot\mid S')
\big\|
\pi_{\mathrm{ref}}(\cdot\mid S')
\bigr)
\right].
\]
For a probability measure \(\mu\) and measurable \(f\), write
\[
\|f\|_{2,\mu}
:=
\left\{
\E_{\mu}
[f(S,A)^2]
\right\}^{1/2}.
\]

\subsection{Occupancy Reweighting: Intuition and Population Update}
\label{sec:prelim-fqi-weighting}

With function approximation, soft FQI applies the Bellman operator
\(\mathcal T_\tau\) to the current \(Q\)-function and fits
\(\mathcal T_\tau Q\) within \(\mathcal F\), typically by least squares under
the offline distribution \(\nu_b\) \citep{munos2008finite}. More generally,
for a probability measure \(\mu\), let
\[
\Pi_{\mathcal F}^{\mu}g
:=
\argmin_{f\in\mathcal F}
\|g-f\|_{2,\mu}^2
\]
denote the \(L^2(\mu)\) projection onto a nonempty closed convex class
\(\mathcal F\) of bounded measurable functions. Standard population soft FQI
therefore iterates
\(\Pi_{\mathcal F}^{\nu_b}\mathcal T_\tau\). Although
\(\mathcal T_\tau\) is contractive in the supremum norm, projection in
\(L^2(\nu_b)\) need not preserve this contraction. Bellman completeness,
\(\mathcal T_\tau(\mathcal F)\subseteq\mathcal F\), avoids this mismatch
because projection does not alter \(\mathcal T_\tau Q\) when
\(Q\in\mathcal F\)
\citep{gordon1995stable,tsitsiklis1996analysis,chen2019information}.

Rather than impose Bellman completeness, we choose the projection norm using
the behavior of \(\mathcal T_\tau\) near \(Q_\tau^\star\). Its derivative at
\(Q_\tau^\star\) is
\(D\mathcal T_\tau(Q_\tau^\star)[H]
=\gamma P_{\pi_\tau^\star}H\), so for \(Q\) near \(Q_\tau^\star\),
\[
\mathcal T_\tau(Q)
\approx
\mathcal T_\tau(Q_\tau^\star)
+
\gamma P_{\pi_\tau^\star}(Q-Q_\tau^\star)
=
\mathcal T_\tau^{\pi_\tau^\star}(Q).
\]
By the discounted balance equation,
\(\gamma P_{\pi_\tau^\star}\) is a \(\sqrt{\gamma}\)-contraction in
\(L^2(\mu_\tau^\star)\)
\citep{vdLaanKallus2025FQE}. This suggests fitting the Bellman target in
\(L^2(\mu_\tau^\star)\), rather than \(L^2(\nu_b)\), when the current
iterate is close to \(Q_\tau^\star\).

The corresponding oracle population update projects the Bellman target onto
\(\mathcal F\) in \(L^2(\mu_\tau^\star)\). Since metric projection is
nonexpansive in \(L^2(\mu_\tau^\star)\), the
\(\sqrt{\gamma}\)-contraction of the linearization is preserved after
projection. This yields only a local contraction because the linear
approximation of \(\mathcal T_\tau\) deteriorates away from
\(Q_\tau^\star\).

If \(\mu_\tau^\star\ll\nu_b\), the oracle projection can be computed under
the offline distribution by weighting with
\(\mathrm d\mu_\tau^\star/\mathrm d\nu_b\). This weight depends on the
unknown soft-optimal policy \(\pi_\tau^\star\), so we instead use the
reference-restart occupancy induced by an available iterate
\(Q^{\mathrm{init}}\) and hold it fixed across subsequent updates:
\begin{equation}
\label{eq:stage-frozen-map}
\mathcal U_{\tau,Q^{\mathrm{init}}}(Q)
:=
\Pi_{\mathcal F}^{\mu_{\tau,Q^{\mathrm{init}}}}
\mathcal T_\tau Q.
\end{equation}
When \(Q^{\mathrm{init}}\) is sufficiently close to \(Q_\tau^\star\), this
occupancy is close enough to the oracle occupancy for the local contraction
to persist. The reference restart retains the fixed component
\((1-\gamma)\nu_{\mathrm{ref}}\), preserving nonvanishing mass under the
reference distribution even as the induced policy becomes highly
concentrated.

The remaining challenge is that the local contraction region shrinks as
temperature decreases. Specifically, the nonlinear remainder grows on the
scale \(\|Q-Q_\tau^\star\|_\infty/\tau\). We therefore use
\emph{temperature annealing} to turn these local contraction results into a
global convergence argument. In the spirit of classical continuation methods,
we begin at a sufficiently large temperature, where the projected population
map satisfies a contraction bound over the entire function class and therefore
permits arbitrary initialization. We then decrease \(\tau\) gradually, using
the output at each temperature to initialize the next and track the path of
soft-optimal solutions \(\{Q_\tau^\star\}\) as the contraction regions shrink.

\section{Occupancy-Reweighted Soft FQI}
\label{sec:occupancy-reweighted-fqi}

Occupancy-reweighted soft FQI replaces the population projection in
\eqref{eq:stage-frozen-map} by weighted least squares on offline transitions
\((S,A,R,S')\), where \((S,A)\sim\nu_b\),
\(S'\mid(S,A)\sim P(\cdot\mid S,A)\), and
\(\E[R\mid S,A]=\bar r(S,A)\).
For the theoretical analysis, we use fresh batches to decouple successive
regression steps, as in finite-sample analyses of fitted value and
\(Q\)-iteration
\citep{munos2008finite,calandriello2014sparse,fan2020theoretical}.
Index temperatures by \(j\) and within-temperature updates by \(k\), and let
\(\mathcal D_j^w\) and \(\mathcal D_{j,k}\) denote mutually independent ratio
and regression batches of sizes \(n_w\) and \(m\), respectively.

\textbf{Positive-temperature annealing.}
Fix \(\tau_0>\tau_{\mathrm{tar}}>0\), a number of cooling intervals \(J\),
and \(K\) regression updates per temperature. Use the geometric schedule
\begin{equation}
\label{eq:geometric-temperature-schedule}
\tau_j
:=
\tau_0
\left(\frac{\tau_{\mathrm{tar}}}{\tau_0}\right)^{j/J},
\qquad
0\le j\le J.
\end{equation}
Starting from \(\widehat Q_{0,0}\in\mathcal F\), at temperature \(\tau_j\)
define
\(\mu_j:=\mu_{\tau_j,\widehat Q_{j,0}}\), estimate
\(w_j:=\mathrm d\mu_j/\mathrm d\nu_b\) from \(\mathcal D_j^w\) by
\(\widehat w_j\), and hold these weights fixed for the next \(K\) regression
updates. Specifically, for \(k=0,\ldots,K-1\) and
\(\mathcal D_{j,k}
=\{(S_{j,k,i},A_{j,k,i},R_{j,k,i},S'_{j,k,i})\}_{i=1}^m\), set
\begin{equation}
\label{eq:empirical-soft-fqi-update}
\widehat Q_{j,k+1}
\in
\argmin_{f\in\mathcal F}
\frac1m\sum_{i=1}^m
\widehat w_j(S_{j,k,i},A_{j,k,i})
\left\{
R_{j,k,i}
+\gamma V_{\tau_j,\widehat Q_{j,k}}(S'_{j,k,i})
-f(S_{j,k,i},A_{j,k,i})
\right\}^2.
\end{equation}
For \(j<J\), warm-start the next temperature by setting
\(\widehat Q_{j+1,0}:=\widehat Q_{j,K}\). The final output is
\(\widehat Q_{J,K}\) and its induced policy
\(\pi_{\tau_{\mathrm{tar}},\widehat Q_{J,K}}\).

\textbf{Optional hard FQI continuation.}
To target unregularized control, set
\(\tau_{\mathrm{tar}}=\tau_{\mathrm{sw}}>0\) and initialize hard FQI with
\(\widehat Q_{J,K}\). At \(\tau=0\), the greedy policy can change
discontinuously with \(Q\), so we refresh the reference-restart occupancy
ratio after every hard update. Fix a deterministic tie-breaking rule, let
\(\pi_{0,Q}\) denote the resulting greedy policy, and define
\(\mu_{0,Q}\) by using \(\pi_{0,Q}\) as the continuation policy in
\eqref{eq:stage-frozen-occupancy}.

Let \(\mathcal D_k^{w,\mathrm h}\) and
\(\mathcal D_k^{\mathrm h}\) denote independent ratio and regression batches
of sizes \(n_{w,\mathrm h}\) and \(m_{\mathrm h}\), respectively, and set
\(\widehat Q_0^{\mathrm h}:=\widehat Q_{J,K}\). At hard iteration \(k\),
define \(\mu_k^{\mathrm h}:=\mu_{0,\widehat Q_k^{\mathrm h}}\), estimate
\(w_k^{\mathrm h}:=\mathrm d\mu_k^{\mathrm h}/\mathrm d\nu_b\) from
\(\mathcal D_k^{w,\mathrm h}\) by \(\widehat w_k^{\mathrm h}\), and update
using
\begin{equation}
\label{eq:empirical-hard-fqi-update}
\widehat Q_{k+1}^{\mathrm h}
\in
\argmin_{f\in\mathcal F}
\frac1{m_{\mathrm h}}\sum_{i=1}^{m_{\mathrm h}}
\widehat w_k^{\mathrm h}(S_{k,i},A_{k,i})
\left\{
R_{k,i}
+\gamma\max_{a'\in\mathcal A}
\widehat Q_k^{\mathrm h}(S'_{k,i},a')
-f(S_{k,i},A_{k,i})
\right\}^2,
\end{equation}
where
\(\mathcal D_k^{\mathrm h}
=\{(S_{k,i},A_{k,i},R_{k,i},S'_{k,i})\}_{i=1}^{m_{\mathrm h}}\).
After \(K_{\mathrm{hard}}\) iterations, return
\(\widehat Q_{K_{\mathrm{hard}}}^{\mathrm h}\) and its greedy policy.

The fresh-batch construction makes each regression batch independent of the
estimated weight and of the data used to construct the current iterate,
simplifying the finite-sample analysis. In practice, one may reuse a common
offline dataset across ratio estimation and regression updates; cross-fitting
provides a practical way to retain sample splitting while using the data more
efficiently \citep{foster2023orthogonal,van2026researcher}. The
positive-temperature phase uses \(J+1\) ratio fits and \((J+1)K\) regression
updates, whereas the optional hard phase uses \(K_{\mathrm{hard}}\) of each,
since weights are fixed within each positive-temperature stage but refreshed
at every hard iteration. For fixed initial and target temperatures, finitely
many temperature stages suffice when the approximation and statistical errors
are small enough for the annealing path to remain within the local contraction
regions. Moreover, when the statistical errors decay polynomially in the
per-batch sample sizes, geometric within-stage convergence requires only
logarithmically many updates to reduce optimization error to the same order.

\textbf{Discounted occupancy-ratio estimation.}
Our analysis does not require a particular ratio estimator, only control of
the estimation error for each fixed-policy reference-restart occupancy.
Unlike in off-policy value estimation, the ratio here determines the weighting
measure for least-squares regression rather than correcting a value estimand.
Available approaches include DICE estimators, minimax estimators, and
balancing-weight methods
\citep{liu2018breaking,nachum2019dualdice,zhang2020gendice,
uehara2020minimax,wang2023projected}. Fitted occupancy-ratio evaluation
(FORE) provides another option based on fitted fixed-point iteration with
density-ratio losses \citep{van2026fitted}. Ratio estimates may be constrained
or clipped as needed and warm-started across temperatures or, during the hard
continuation, across iterations.

\section{Population Convergence Analysis}
\label{sec:population-theory}

We first establish local contraction at a fixed positive temperature with the
reference-restart occupancy held fixed across updates. Temperature annealing
then connects successive contraction regions to yield convergence from
arbitrary initialization.

\subsection{Local Convergence at a Positive Temperature}
\label{sec:local-geometry}

Fix \(\tau>0\). Section~\ref{sec:prelim-fqi-weighting} showed that projection
in \(L^2(\mu_\tau^\star)\) preserves the contraction of the first-order
approximation of \(\mathcal T_\tau\) near \(Q_\tau^\star\). The population
algorithm instead projects under \(\mu_{\tau,Q^{\mathrm{init}}}\), induced by
the initializer and held fixed across updates. We therefore characterize how
close the initializer and subsequent iterates must remain to
\(Q_\tau^\star\) for contraction to persist. We impose the following
conditions.

\begin{enumerate}[label=\textbf{C\arabic*)}, ref={C\arabic*},
leftmargin=1.5em, series=cond]

\item \label{cond::convex}
\textbf{Function-class regularity.}
\(\mathcal F\) is convex and compact in the supremum norm.

\item \label{cond::fitted-realizability}
\textbf{Realizability.}
\(Q_\tau^\star\in\mathcal F\).

\item \label{cond::supnorm}
\textbf{Reference-measure interpolation.}
There exist \(\alpha\in(0,1)\) and \(C_0<\infty\) such that
\[
\|f-f'\|_\infty
\le
C_0
\|f-f'\|_{2,\nu_{\mathrm{ref}}}^{\alpha}
\qquad
\text{for every }f,f'\in\mathcal F.
\]

\end{enumerate}
Condition~\ref{cond::convex} holds, for example, for finite-dimensional
linear classes with bounded features and a compact convex coefficient set,
and for many bounded smooth nonparametric classes; examples are given in
Appendix~\ref{app:local-regularity-examples}.
Condition~\ref{cond::fitted-realizability} is needed for exact convergence to
\(Q_\tau^\star\) at a fixed positive temperature, while the annealing result
below allows misspecification at intermediate temperatures.
Condition~\ref{cond::supnorm} converts
\(L^2(\nu_{\mathrm{ref}})\) control into uniform control. Since the soft policy
varies with \(Q\) on the scale
\(\|Q-Q_\tau^\star\|_\infty/\tau\), this condition controls the nonlinear
remainder in the first-order approximation and hence the size of the local
contraction region.

The interpolation condition holds for several standard regular classes under
appropriate design-measure assumptions. For finite-dimensional linear classes
with bounded features and a nonsingular reference Gram matrix, it holds with
exponent \(1\), and hence with any fixed \(\alpha\in(0,1)\) on a bounded
class after adjusting \(C_0\). For bounded RKHS balls with uniformly bounded
Mercer eigenfunctions and eigenvalues satisfying
\(\lambda_j\lesssim j^{-2r}\), \(r>1/2\), one may take
\(\alpha=1-1/(2r)\) \citep{mendelson2010regularization}. On
\(d\)-dimensional domains, one may take \(\alpha=2s/(2s+d)\) for H\"older
classes of smoothness \(s>0\) \citep{triebel2006theory} and
\(\alpha=1-d/(2s)\) for Sobolev classes of smoothness \(s>d/2\)
\citep{adams2003sobolev}; see
Appendix~\ref{app:local-regularity-examples} for details.

Set \(C_{\mathrm{int}}:=C_0(1-\gamma)^{-\alpha/2}\). For
\(R_{\mathrm{loc}}>0\), define
\begin{align}
\label{eq:positive-frozen-constants}
\rho_{\tau}(R_{\mathrm{loc}})
&:=
\left[
\gamma
\exp\left\{
\frac{4C_{\mathrm{int}}R_{\mathrm{loc}}^\alpha}{\tau}
\right\}
\right]^{1/2},
&
R_\tau^{\max}
&:=
\left[
\frac{\tau}{4C_{\mathrm{int}}}
\log\left(\frac1\gamma\right)
\right]^{1/\alpha}.
\end{align}
Then \(\rho_{\tau}(R_{\mathrm{loc}})<1\) whenever
\(R_{\mathrm{loc}}<R_\tau^{\max}\).

\begin{theorem}[Local population convergence with fixed occupancy]
\label{thm:positive-stage-frozen-local}
Assume Conditions~\ref{cond::convex}--\ref{cond::supnorm}. Fix
\(0<R_{\mathrm{loc}}<R_\tau^{\max}\) and suppose
\(\|Q_0-Q_\tau^\star\|_{2,\mu_{\tau,Q_0}}
\le R_{\mathrm{loc}}\).
Then the iterates
\(Q_{k+1}=\mathcal U_{\tau,Q_0}(Q_k)\) satisfy
\begin{equation}
\label{eq:positive-stage-frozen-rate}
\|Q_k-Q_\tau^\star\|_{2,\mu_{\tau,Q_0}}
\le
\rho_{\tau}(R_{\mathrm{loc}})^k
\|Q_0-Q_\tau^\star\|_{2,\mu_{\tau,Q_0}},
\qquad k\ge0.
\end{equation}
\end{theorem}
Theorem~\ref{thm:positive-stage-frozen-local} gives geometric convergence to
\(Q_\tau^\star\) within a contraction region of radius
\(R_\tau^{\max}\asymp\tau^{1/\alpha}\). Thus the region shrinks
polynomially as \(\tau\downarrow0\). Moreover, for any fixed
\(\theta\in(0,1)\),
\(\rho_\tau(\theta R_\tau^{\max})
=\gamma^{(1-\theta^\alpha)/2}\), so the convergence rate on any fixed
fraction of the contraction region is uniform in \(\tau\). The shrinking
contraction region motivates the temperature-annealing argument below.

\subsection{Global Convergence via Temperature Annealing}
\label{sec:population-annealing}

Temperature annealing removes the local initialization requirement of
Theorem~\ref{thm:positive-stage-frozen-local}. Lemma~
\ref{prop:global-high-temperature} shows that there exists a finite threshold
\(\tau_{\mathrm{init}}\) above which the population update satisfies a
contraction bound over all of \(\mathcal F\), uniformly in the initializer,
with modulus
\(\rho_{\mathrm{init}}:=(1+\sqrt{\gamma})/2<1\), up to approximation error.
Fix \(\tau_0\ge\tau_{\mathrm{init}}\) and define
\(R_{\mathrm{init}}
:=\sup_{Q\in\mathcal F}
\|Q-Q_{\tau_0}^\star\|_{2,\mu_{\tau_0,Q}}<\infty\).
After \(K\) updates, the contribution of arbitrary initialization is at most
\(\rho_{\mathrm{init}}^K R_{\mathrm{init}}\). Moreover,
\(\|Q_\tau^\star-Q_{\tau'}^\star\|_\infty
\le \gamma L_{\mathrm{ref}}|\tau-\tau'|/(1-\gamma)\), so gradual cooling
allows the output at one temperature to initialize the next.

Fix \(\tau_{\mathrm{tar}}\in(0,\tau_0)\), use the geometric schedule
\eqref{eq:geometric-temperature-schedule}, and set
\(Q_j^\star:=Q_{\tau_j}^\star\) for \(0\le j\le J\).
Starting from any \(Q_{0,0}\in\mathcal F\), define
\(\mu_j:=\mu_{\tau_j,Q_{j,0}}\) and
\(\|H\|_j:=\|H\|_{2,\mu_j}\). At temperature \(\tau_j\), iterate
\(Q_{j,k+1}:=\mathcal U_{\tau_j,Q_{j,0}}(Q_{j,k})\) for
\(0\le k<K\), holding \(\mu_j\) fixed, and set
\(Q_{j+1,0}:=Q_{j,K}\) for \(j<J\). At
\(\tau_J=\tau_{\mathrm{tar}}\), continue iterating with \(\mu_J\) fixed.

We allow \(\mathcal F\) to be misspecified at intermediate temperatures. Let
\[
\varepsilon_j
:=
\inf_{f\in\mathcal F}\|f-Q_j^\star\|_\infty,
\qquad
\varepsilon_{\mathrm{path}}
:=
\max_{0\le j\le J}\varepsilon_j.
\]
To account for this approximation error, fix \(R_{\mathrm{loc}}>0\) and define
\begin{equation}
\label{eq:annealing-local-modulus}
\begin{aligned}
\rho_{\tau}(R_{\mathrm{loc}},\varepsilon)
&:=
\left[
\gamma
\exp\left\{
\frac{
4\{C_{\mathrm{int}}(R_{\mathrm{loc}}+\varepsilon)^\alpha+\varepsilon\}
}{\tau}
\right\}
\right]^{1/2},
&
\bar\rho
&:=
\max\left\{
\rho_{\mathrm{init}},
\max_{1\le j\le J}
\rho_{\tau_j}(R_{\mathrm{loc}},\varepsilon_j)
\right\}.
\end{aligned}
\end{equation}
Thus \(\rho_\tau(R_{\mathrm{loc}},0)=\rho_\tau(R_{\mathrm{loc}})\), while
\(\bar\rho\) is a common contraction modulus across temperatures.

Write
\(\operatorname{err}_{\mathrm{pop}}(\varepsilon)
:=2\varepsilon/(1-\bar\rho)\) for the limiting within-temperature error due
to approximation error \(\varepsilon\).
\begin{theorem}[Global population convergence under geometric annealing]
\label{thm:population-annealing}
Assume Conditions~\ref{cond::convex} and~\ref{cond::supnorm}, and choose
\(\tau_0\ge\tau_{\mathrm{init}}\). Suppose
\(\bar\rho<1\) and
\(\operatorname{err}_{\mathrm{pop}}(\varepsilon_{\mathrm{path}})<R_{\mathrm{loc}}\).
Choose \(K,J\) such that
\[
C_{\mathrm{int}}
\Bigl\{
\bar\rho^K\max\{R_{\mathrm{init}},R_{\mathrm{loc}}\}
+ \operatorname{err}_{\mathrm{pop}}(\varepsilon_{\mathrm{path}})
+\varepsilon_{\mathrm{path}}
\Bigr\}^{\alpha}
+\varepsilon_{\mathrm{path}}
+\frac{\gamma\tau_0L_{\mathrm{ref}}}{(1-\gamma)J}
\log\left(\frac{\tau_0}{\tau_{\mathrm{tar}}}\right)
\le R_{\mathrm{loc}}.
\]
Then, from every \(Q_{0,0}\in\mathcal F\), subsequent iterations at the
target temperature with \(\mu_J\) fixed satisfy
\[
\|Q_{J,k}-Q_J^\star\|_J
\le
\bar\rho^kR_{\mathrm{loc}}
+\operatorname{err}_{\mathrm{pop}}(\varepsilon_J),
\qquad
k\ge0.
\]
\end{theorem}
Theorem~\ref{thm:population-annealing} removes the local initialization
requirement: the high-temperature stage provides a global starting point,
while the annealing schedule keeps each subsequent warm start within the next
contraction region. Intermediate temperatures may be misspecified; only the
target approximation error affects the limiting target-stage bound. In
particular, if \(Q_J^\star\in\mathcal F\), then \(\varepsilon_J=0\) and
\[
\|Q_{J,k}-Q_J^\star\|_J
\le
\bar\rho^kR_{\mathrm{loc}}
\to0,
\]
so realizability is required only at the target for exact convergence.

The condition
\(\operatorname{err}_{\mathrm{pop}}(\varepsilon_{\mathrm{path}})
<R_{\mathrm{loc}}\) keeps the limiting within-stage error inside the local
contraction region. The displayed schedule condition instead controls
transfer between stages: its left-hand side bounds the next-stage
initialization error, and its right-hand side is the admissible radius.
Increasing \(K\) reduces the residual optimization error geometrically,
while increasing \(J\) reduces the largest temperature change between
successive stages. Lemma~\ref{prop:uniform-annealing-feasibility} shows that
finite \(K,J\) satisfy this condition when the path approximation error is
sufficiently small.

\subsection{Global Convergence with a Hard FQI Continuation}

To target unregularized control, we stop annealing at a switch temperature
\(\tau_{\mathrm{sw}}>0\) and use the terminal soft iterate to initialize the
hard continuation in Section~\ref{sec:occupancy-reweighted-fqi}. This avoids
following the soft-optimal path through increasingly small contraction regions
as \(\tau\downarrow0\).

Recall \(V_{0,Q}(s)=\max_a Q(s,a)\), and define
\[
(\mathcal T_0Q)(s,a)
:=
\bar r(s,a)
+\gamma\E[V_{0,Q}(S')\mid s,a].
\]
Let \(Q_{\mathrm{hard}}^\star\) denote the bounded fixed point of
\(\mathcal T_0\), and let \(a_{\mathrm{hard}}^\star(s)\) be the maximizing
action selected by the fixed tie-breaking rule. Write
\(\mu_{\mathrm{hard}}:=\mu_{0,Q_{\mathrm{hard}}^\star}\), with state
marginal \(\xi_{\mathrm{hard}}\), and define the hard action gap
\[
G(s)
:=
Q_{\mathrm{hard}}^\star(s,a_{\mathrm{hard}}^\star(s))
-
\max_{a\ne a_{\mathrm{hard}}^\star(s)}
Q_{\mathrm{hard}}^\star(s,a).
\]
The hard phase requires the following additional conditions.

\begin{enumerate}[label=\textbf{H\arabic*)}, ref={H\arabic*},
leftmargin=1.5em, series=hardpolish]

\item \label{cond::hard-realizability}
\textbf{Hard-\(Q\) realizability.}
\(Q_{\mathrm{hard}}^\star\in\mathcal F\).

\item \label{cond::hard-soft-margin}
\textbf{Hard action-gap margin and compatibility.}
There exist \(\beta>0\) and \(C_G<\infty\) such that
\[
\xi_{\mathrm{hard}}\{G\le u\}
\le C_Gu^\beta,
\qquad
0<u\le1,
\qquad
\alpha\left(1+\frac{\beta}{2}\right)>1.
\]

\end{enumerate}
Condition~\ref{cond::hard-soft-margin} controls states near a tie, where a
small error in \(Q\) can change the greedy action. If the hard action gap is
uniformly positive, \(G\ge g_{\min}>0\)
\(\xi_{\mathrm{hard}}\)-almost surely, then the margin condition holds for
every \(\beta>0\). More generally, in continuous-state models, a bounded
density for \(G(S)\) near zero gives \(\beta=1\), in which case the
compatibility condition requires \(\alpha>2/3\).

Lemma~\ref{lem:general-reference-soft-hard-comparison} gives
\[
\|Q_\tau^\star-Q_{\mathrm{hard}}^\star\|_\infty
\le
\frac{\gamma L_{\mathrm{ref}}}{1-\gamma}\tau,
\qquad \tau>0.
\]
Thus a sufficiently accurate soft iterate at a sufficiently small switch
temperature can enter the hard contraction region.

\begin{theorem}[Global convergence to the hard optimum]
\label{thm:population-hard-polishing}
Assume Conditions~\ref{cond::convex}, \ref{cond::supnorm}, and
\ref{cond::hard-realizability}--\ref{cond::hard-soft-margin}. Suppose the
conditions of Theorem~\ref{thm:population-annealing} hold with
\(\tau_{\mathrm{tar}}=\tau_{\mathrm{sw}}>0\). There exists
\(R_{\mathrm{hard}}>0\) such that, if
\begin{equation}
\label{eq:hard-switch-temperature}
C_{\mathrm{int}}
\left\{
\bar\rho^K R_{\mathrm{loc}}
+ \operatorname{err}_{\mathrm{pop}}(\varepsilon_J)
+\varepsilon_J
\right\}^{\alpha}
+\varepsilon_J
+\frac{\gamma L_{\mathrm{ref}}}{1-\gamma}\tau_{\mathrm{sw}}
<
R_{\mathrm{hard}},
\end{equation}
then, from every \(Q_{0,0}\in\mathcal F\), the terminal soft iterate
\(Q_{J,K}\) lies in the hard contraction region. Initializing the hard phase
with \(Q_0^{\mathrm h}=Q_{J,K}\), the hard iterates satisfy
\begin{equation}
\label{eq:hard-polishing-l2-rate}
\|Q_k^{\mathrm h}-Q_{\mathrm{hard}}^\star\|_
{2,\mu_{\mathrm{hard}}}
\le
\left(\frac{1+\sqrt\gamma}{2}\right)^k
\|Q_0^{\mathrm h}-Q_{\mathrm{hard}}^\star\|_
{2,\mu_{\mathrm{hard}}},
\qquad k\ge0.
\end{equation}
\end{theorem}

The switch condition requires the terminal soft error and soft-to-hard bias
to fit within the hard contraction region. As \(K\to\infty\), the
optimization component vanishes, so sufficiently small endpoint approximation
error and switch temperature ensure the condition for some finite \(K\).
Exact realizability at \(\tau_{\mathrm{sw}}\) is therefore not required.
Once the hard phase begins, convergence is geometric with a rate independent
of the switch temperature; Condition~\ref{cond::supnorm} also yields uniform
convergence.

\section{Finite-Sample Guarantees}
\label{sec::finitesample}

We analyze the fresh-batch positive-temperature and hard-continuation
procedures described in Section~\ref{sec:occupancy-reweighted-fqi}. The same
annealing construction and population contraction regions carry over, while
sampling adds regression and occupancy-ratio estimation error to the
within-stage error bound. Fix a failure probability \(\eta\in(0,1/4)\).

Regression error is controlled by a metric-entropy critical radius. Let
\(N(\varepsilon,\mathcal F,L^2(\mathsf P))\) denote the
\(L^2(\mathsf P)\)-covering number of \(\mathcal F\), and define
\[
\mathcal J(\delta,\mathcal F)
:=
\int_0^\delta
\sup_{\mathsf P}
\sqrt{\log N(\varepsilon,\mathcal F,L^2(\mathsf P))}
\,\mathrm d\varepsilon,
\qquad
\delta_\ell
:=
\inf\left\{
\delta>0:
\mathcal J(\delta,\mathcal F)
\le
\sqrt{\ell}\,\delta^2
\right\}
\vee
\sqrt{\frac{\log(e\ell)}{\ell}},
\]
where the supremum is over finitely supported distributions. For standard
classes, \(\delta_\ell\) has order
\(\sqrt{s_0\log(p/s_0)/\ell}\) for \(s_0\)-sparse linear classes,
\(\sqrt{V\log(\ell/V)/\ell}\) for VC-subgraph classes, and
\(\ell^{-s/(2s+d)}\) for \(d\)-variate H\"older or Sobolev classes of
smoothness \(s\)
\citep{van1996weak,nickl2007bracketing,mendelson2010regularization,
wainwright2019high}.

Occupancy-ratio estimation changes the weighting measure used by each
regression. To convert control of the weighted loss into control in the
target occupancy norm, we impose the following condition.

\begin{enumerate}[label=\textbf{C\arabic*)}, ref={C\arabic*},
leftmargin=1.5em, resume=cond]

\item \label{cond:finite-frozen-curvature}
\textbf{Bounded weights and restricted curvature.}
There exist \(B_w<\infty\) and
\(\underline\lambda_{\mathcal F}>0\) such that every target occupancy
\(\mu\) and estimated weight \(\widehat w\) used by the algorithm satisfy
\(\mu\ll\nu_b\) and
\[
0\le\widehat w\le B_w
\quad\nu_b\text{-almost surely},
\qquad
\E_{\nu_b}[\widehat wH^2]
\ge
\underline\lambda_{\mathcal F}\|H\|_{2,\mu}^2
\quad\text{for every }H\in\mathcal F-\mathcal F.
\]

\end{enumerate}

For the exact ratio \(w=\mathrm d\mu/\mathrm d\nu_b\), the two quadratic
forms are equal. A simple sufficient condition is
\(\widehat w\ge c w\) pointwise for some \(c>0\), which gives
\(\underline\lambda_{\mathcal F}=c\). The reference restart also implies
\(w\ge(1-\gamma)e^{-L_{\mathrm{ref}}}\) \(\nu_b\)-almost surely, so
\(\|\widehat w-w\|_\infty
\le\delta(1-\gamma)e^{-L_{\mathrm{ref}}}\), with \(\delta<1\), is
sufficient with \(c=1-\delta\). For linear classes,
Lemma~\ref{prop:ratio-curvature-linear} gives another sufficient
condition based on a nonsingular reference Gram matrix and sufficiently
accurate relative \(L^2\) ratio estimation.
 
\subsection{Positive-Temperature Annealing}

We use the stage occupancies and weights
\((\mu_j,w_j,\widehat w_j)\) defined in
Section~\ref{sec:occupancy-reweighted-fqi}, with \(\mu_j\) now random through
the empirical initializer \(\widehat Q_{j,0}\).

Let \(C_{\mathrm{reg}}\) and \(C_w\) denote the regression- and ratio-error
constants in the one-step bound of
Lemma~\ref{lem:finite-stage-frozen-one-step}. There are \((J+1)K\)
regression batches of size \(m\); define the uniform regression-error scale
\[
\delta_{\mathrm{reg}}
:=
C_{\mathrm{reg}}
\left\{
\delta_m+
\sqrt{\frac{\log\{e(J+1)K/\eta\}}{m}}
\right\}.
\]
For the \(J+1\) ratio batches of size \(n_w\), suppose that, for some
deterministic \(\delta_w>0\),
\begin{equation}
\label{eq:finite-frozen-ratio-guarantee}
\Pr\left\{
\max_{0\le j\le J}
\left\|
\frac{\widehat w_j}{w_j}-1
\right\|_{2,\mu_j}
\le \delta_w
\right\}
\ge 1-\eta.
\end{equation}
Such a guarantee can be obtained, for example, using fitted occupancy-ratio
evaluation \citep{van2026fitted}; see
Lemma~\ref{cor:fore-stagewise-control}.

Set \(\rho_{\mathrm{ann}}:=(1+\bar\rho)/2\).
Define the finite-sample within-stage error bound by
\begin{equation}
\label{eq:finite-annealing-error-bound-main}
\operatorname{err}_{\mathrm{ann}}(\varepsilon)
:=
\frac{
\delta_{\mathrm{reg}}
+2\varepsilon
+C_w(1+C_{\mathrm{int}})\delta_w\varepsilon^\alpha
}
{1-\rho_{\mathrm{ann}}}
+
\left\{
\frac{
C_w(1+C_{\mathrm{int}})\delta_w
}{
1-\rho_{\mathrm{ann}}
}
\right\}^{1/(1-\alpha)}.
\end{equation}
\begin{theorem}[Finite-sample convergence under geometric annealing]
\label{thm:finite-sample-annealing}
Assume Conditions~\ref{cond::convex}, \ref{cond::supnorm}, and
\ref{cond:finite-frozen-curvature}. Let
\(\tau_0\ge\tau_{\mathrm{init}}\), and suppose
\(\bar\rho<1\),
\eqref{eq:finite-frozen-ratio-guarantee} holds,
and
\(\operatorname{err}_{\mathrm{ann}}(\varepsilon_{\mathrm{path}})<R_{\mathrm{loc}}\).
Choose \(J,K\) such that
\[
C_{\mathrm{int}}
\Bigl\{
\rho_{\mathrm{ann}}^K\max\{R_{\mathrm{init}},R_{\mathrm{loc}}\}
+
\operatorname{err}_{\mathrm{ann}}(\varepsilon_{\mathrm{path}})
+\varepsilon_{\mathrm{path}}
\Bigr\}^{\alpha}
+\varepsilon_{\mathrm{path}}
+\frac{\gamma\tau_0L_{\mathrm{ref}}}{(1-\gamma)J}
\log\left(\frac{\tau_0}{\tau_{\mathrm{tar}}}\right)
\le R_{\mathrm{loc}}.
\]
Then, from every
\(\widehat Q_{0,0}\in\mathcal F\), with probability at least \(1-2\eta\),
\begin{equation}
\label{eq:finite-annealing-target-main}
\|\widehat Q_{J,K}-Q_{\tau_{\mathrm{tar}}}^\star
\|_{2,\mu_J}
\le
\rho_{\mathrm{ann}}^K R_{\mathrm{loc}}
+\operatorname{err}_{\mathrm{ann}}(\varepsilon_J).
\end{equation}
\end{theorem}
The condition
\(\operatorname{err}_{\mathrm{ann}}(\varepsilon_{\mathrm{path}})
<R_{\mathrm{loc}}\) ensures that the limiting within-stage error remains
inside the local contraction region, while the displayed schedule condition
controls transfer between successive temperatures. When the approximation,
regression, and ratio-estimation errors are sufficiently small, finite \(J,K\)
satisfy both conditions: increasing \(J\) reduces temperature drift, while
increasing \(K\) reduces optimization error geometrically. With fixed
per-batch sample sizes, the statistical error bounds depend on \(J,K\) only
through logarithmic factors.

Path approximation error therefore determines whether the annealing path can
be traversed, whereas the approximation component of the target-stage error
depends only on \(\varepsilon_J\). If
\(Q_{\tau_{\mathrm{tar}}}^\star\in\mathcal F\),
\(1-\rho_{\mathrm{ann}}\gtrsim1-\gamma\), and the remaining constants are
uniform in \(\gamma\), then \eqref{eq:finite-annealing-error-bound-main} gives
\begin{equation}
\label{eq:finite-annealing-horizon-main}
\operatorname{err}_{\mathrm{ann}}(0)
\lesssim
\frac{\delta_{\mathrm{reg}}}{1-\gamma}
+
\frac{\delta_w^{1/(1-\alpha)}}
{(1-\gamma)^{(1+\alpha/2)/(1-\alpha)}}.
\end{equation}
Thus regression error incurs the standard effective-horizon factor
\((1-\gamma)^{-1}\), while ratio-estimation error enters at the higher order
\(\delta_w^{1/(1-\alpha)}\) but with stronger horizon dependence. For a
feasible schedule, \(K\) can be chosen so that optimization error is no larger
than the statistical error. Notably, the bound has no explicit inverse
dependence on the target temperature \(\tau_{\mathrm{tar}}\).

\subsection{Hard FQI Continuation}

We next analyze the hard continuation of
Section~\ref{sec:occupancy-reweighted-fqi}, initialized after annealing to
\(\tau_{\mathrm{sw}}>0\). We use the hard-phase notation
\((\mu_k^{\mathrm h},w_k^{\mathrm h},\widehat w_k^{\mathrm h})\)
from Section~\ref{sec:occupancy-reweighted-fqi}. Let \(e_{\mathrm{sw}}\)
denote the positive-temperature error bound in
\eqref{eq:finite-annealing-target-main}:
\[
e_{\mathrm{sw}}
:=
\rho_{\mathrm{ann}}^K R_{\mathrm{loc}}
+\operatorname{err}_{\mathrm{ann}}(\varepsilon_J).
\]

For the \(K_{\mathrm{hard}}\) regression batches of size \(m_{\mathrm h}\),
define
\[
\delta_{\mathrm{reg,h}}
:=
C_{\mathrm{reg,h}}
\left\{
\delta_{m_{\mathrm h}}
+
\sqrt{\frac{\log(eK_{\mathrm{hard}}/\eta)}{m_{\mathrm h}}}
\right\},
\qquad
\rho_{\mathrm{hard,fs}}
:=
\frac{3+\sqrt{\gamma}}{4},
\]
where \(C_{\mathrm{reg,h}}\) and \(C_{w,\mathrm h}\) are the regression-
and ratio-error constants, respectively, in the hard one-step bound of
Lemma~\ref{lem:finite-hard-one-step}. Suppose further that, for some
deterministic \(\delta_{w,\mathrm h}>0\),
\begin{equation}
\label{eq:finite-hard-ratio-event}
\Pr\left\{
\max_{0\le k<K_{\mathrm{hard}}}
\left\|
\frac{\widehat w_k^{\mathrm h}}{w_k^{\mathrm h}}-1
\right\|_{2,\mu_k^{\mathrm h}}
\le
\delta_{w,\mathrm h}
\right\}
\ge1-\eta.
\end{equation}
Define the hard-phase statistical error bound
\[
\operatorname{err}_{\mathrm{hard}}
:=
\frac{\delta_{\mathrm{reg,h}}}
{1-\rho_{\mathrm{hard,fs}}}
+
\left\{
\frac{
C_{w,\mathrm h}(1+C_{\mathrm{int}})\delta_{w,\mathrm h}
}{
1-\rho_{\mathrm{hard,fs}}
}
\right\}^{1/(1-\alpha)}.
\]

\begin{theorem}[Finite-sample convergence of the hard FQI continuation]
\label{thm:finite-sample-hard-continuation}
Suppose the conditions of
Theorem~\ref{thm:finite-sample-annealing} hold with
\(\tau_{\mathrm{tar}}=\tau_{\mathrm{sw}}>0\). In addition, assume
Conditions~\ref{cond::hard-realizability}--\ref{cond::hard-soft-margin} and
the ratio guarantee \eqref{eq:finite-hard-ratio-event}. There exists
\(R_{\mathrm{hard,fs}}>0\) such that, if
\begin{equation}
\label{eq:finite-hard-main-switch}
C_{\mathrm{int}}
(e_{\mathrm{sw}}+\varepsilon_J)^\alpha
+\varepsilon_J
+\frac{\gamma L_{\mathrm{ref}}}{1-\gamma}\tau_{\mathrm{sw}}
\le R_{\mathrm{hard,fs}},
\qquad
\operatorname{err}_{\mathrm{hard}}
\le R_{\mathrm{hard,fs}},
\end{equation}
then, from every \(\widehat Q_{0,0}\in\mathcal F\), with probability at
least \(1-4\eta\),
\begin{equation}
\label{eq:finite-hard-main-rate}
\|\widehat Q_{K_{\mathrm{hard}}}^{\mathrm h}
-Q_{\mathrm{hard}}^\star\|_
{2,\mu_{\mathrm{hard}}}
\le
\rho_{\mathrm{hard,fs}}^{K_{\mathrm{hard}}}
R_{\mathrm{hard,fs}}
+\operatorname{err}_{\mathrm{hard}}.
\end{equation}
\end{theorem}
The first condition in \eqref{eq:finite-hard-main-switch} places the final
soft iterate inside the hard contraction region, accounting for the
positive-temperature error \(e_{\mathrm{sw}}\), endpoint approximation
error, and \(O(\tau_{\mathrm{sw}})\) soft-to-hard bias. The second requires
the hard-phase statistical error bound to remain inside that region. Once
these conditions hold, additional hard updates reduce the remaining
optimization error geometrically toward \(\operatorname{err}_{\mathrm{hard}}\).

\section{Numerical Experiments}
\label{sec:numerical-experiments}

We use two two-action ergodic Garnet MDPs to illustrate the two mechanisms
studied above: instability from projection under the offline state-action
distribution and failure of direct low-temperature iteration. The weighting
construction has six states and the annealing construction has ten. Both use
$\gamma=0.95$, a uniform reference policy $\pi_{\mathrm{ref}}$, the behavior
state marginal $\nu_0$ as the restart distribution, and the geometric
temperature schedule
\[
0.8\to0.267773\to0.089628\to0.03.
\]
The fitted class contains the soft-optimal $Q$-functions at all four
temperatures and the hard optimum, but is Bellman incomplete away from these
targets. At each positive-temperature stage, we form
$\pi_{\tau_j,\widehat Q_{j,0}}$, estimate its reference-restart occupancy as in
\eqref{eq:stage-frozen-occupancy}, and hold the resulting regression weights
fixed for $K=400$ Bellman updates. The direct comparator instead performs
$1{,}600$ updates at $\tau=0.03$ using a single frozen occupancy ratio. The
hard-FQI continuation starts from the annealed soft endpoint and performs 400
hard updates, refreshing the reference-restart ratio of the current greedy
policy after each update. We report
\[
\operatorname{Err}_{Q,\tau}(\widehat Q)
:=
(1-\gamma)
\frac{\|\widehat Q-Q_\tau^\star\|_{2,\mu_\tau^\star}}
{\|\bar r\|_\infty},
\]
with the analogous definition at the hard optimum. Full details are given in
Appendix~\ref{app:paired-mechanism}.

\paragraph{Population behavior.}
In the weighting construction, projection under the offline state-action
distribution has dominant eigenvalue $-1.190$ and spectral radius $1.190$, and
the resulting iteration enters a period-two cycle for each of the three fixed
initialization radii. Projection under the reference-restart
occupancy instead has spectral radius $0.95$ and converges. In the annealing
construction, direct iteration at $\tau=0.03$ fails for 190 of the 4,096 fixed
initializations, whereas the geometric temperature schedule
converges from all 4,096, with maximum normalized error below
$9.7\times10^{-9}$. The refreshed hard-FQI continuation converges to the hard
optimum in both constructions.

\paragraph{Finite-sample behavior.}
We compare unweighted FQI, raw two-fold cross-fitted FORE, and the tempered
weights
\[
w^{1/5}/\E[w^{1/5}],
\]
denoted $T_w=5$, following \citet{sinha2022experience}. We define
target-neighborhood success by $\operatorname{Err}_{Q,\tau}\leq0.05$. At
$n=20{,}000$, annealed unweighted FQI in the weighting construction has median
error $0.1674$ and $0\%$ success, whereas both raw FORE and $T_w=5$ have median
error $0.0021$ and $100\%$ success. In the annealing construction, raw FORE
improves from median error $0.0934$ and $6\%$ success under direct iteration to
$0.0063$ and $100\%$ success under annealing. Unweighted FQI and $T_w=5$ attain
$100\%$ success under direct iteration in this construction and are therefore
already in a stable direct-iteration regime, while annealing closes the
initialization gap for raw FORE.

The hard-FQI continuation yields the same weighting contrast. In the weighting
construction, unweighted hard FQI has median error $0.1717$, policy-mismatch
mass $0.1924$, and normalized regret $0.3838$. Refreshed raw FORE and $T_w=5$
both have median error $0.0022$, zero median policy mismatch and regret, and
$100\%$ success. In the annealing construction, the corresponding median errors
are $0.0049$, $0.0220$, and $0.0068$ for unweighted FQI, raw FORE, and $T_w=5$,
respectively; success rates are $86\%$ for raw FORE and $99.7\%$ for $T_w=5$.

\begin{figure}[!t]
\centering
\includegraphics[width=.99\linewidth]
{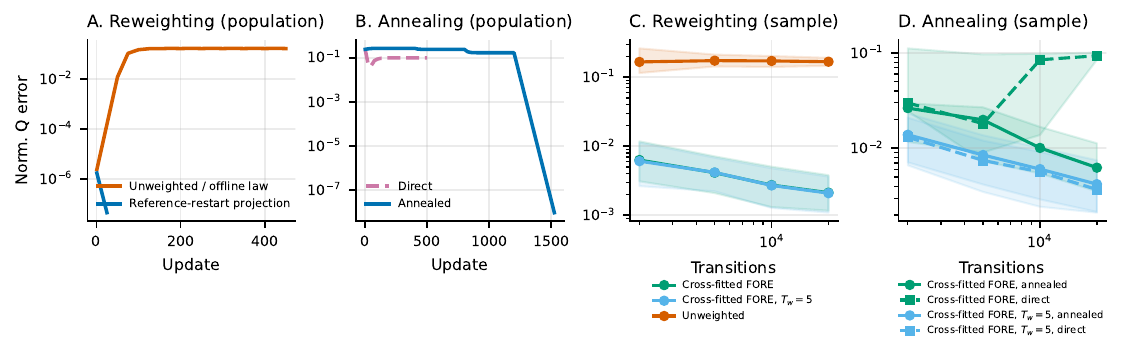}
\caption{\textbf{Convergence dynamics of occupancy-reweighted soft FQI.}
(A) Projection under the offline state-action distribution cycles in the
weighting construction, whereas projection under the reference-restart
occupancy converges.
(B) Direct low-temperature iteration reaches a non-target period-two cycle,
whereas geometric temperature annealing reaches the target.
(C)--(D) Median terminal normalized $Q$-error and interquartile range across
independent datasets, comparing unweighted FQI, raw cross-fitted FORE, and
tempered $T_w=5$ weighting. Results use 150 datasets for $n<20{,}000$ and 300
for $n=20{,}000$.}
\label{fig:soft-fqi-main}
\end{figure}

\subsection{Continuous-State Experiment with Learned Occupancy Ratios}
\label{sec:cartpole-soft-fqi-control}

We next examine the same mechanism when the occupancy ratios must be learned in
a continuous state space. The environment is a stochastic continuing-reset
CartPole-inspired process with four continuous state coordinates and two
actions. Each replicate consists of $25{,}000$ transitions generated from 512
behavior chains after a 2,000-step burn-in. Within each replicate, all methods
use the same offline dataset, reward, 64-dimensional fixed random-feature map,
zero $Q$-initialization, ridge penalty $10^{-5}$, and $\gamma=0.98$.

The direct path performs 80 Bellman updates at $\tau=0.128$, whereas the
annealed path uses temperatures $(2,0.8,0.32,0.128)$ for $(15,15,15,35)$
updates. For each path, we compare unweighted regression with $T_w=5$ FORE
weighting. The direct weighted method estimates its occupancy ratio once, while
the annealed method re-estimates the ratio at the start of each temperature
stage. Each soft method is then followed by five hard-FQI updates, with weighted
methods refreshing the ratio before each hard update; an unweighted hard-only
comparator performs 85 updates from $Q_0$. Thus the methods are matched in the
number of $Q$-projections, while weighted methods incur additional
occupancy-ratio estimation. We report discounted return after 80 soft updates
over 30 independently generated training datasets.

At $K=80$, weighted annealed soft FQI improves raw discounted return over
matched unweighted annealed soft FQI by
$\CartPoleControlPrimaryGain$ points (95\% CI
$[\CartPoleControlPrimaryCILower,\CartPoleControlPrimaryCIUpper]$), with a
positive paired difference in $\CartPoleControlPrimaryWins$ replicates. The
corresponding improvement in KL-regularized return is $4.70$ points (95\% CI
$[3.64,5.83]$). For the complete annealed procedure, including stagewise
occupancy-ratio refresh, the annealing-by-weighting contrast is $6.88$ points
(95\% CI $[4.91,9.05]$). By comparison, direct weighting with a single
target-temperature ratio changes raw return by $-1.85$ points (95\% CI
$[-3.48,-0.28]$). After five additional hard-FQI
updates, weighted annealing exceeds matched unweighted annealing by $5.43$
points (95\% CI $[4.33,6.59]$) and the unweighted hard-only comparator by
$5.89$ points (95\% CI $[4.80,7.02]$). Across the four annealed
occupancy-ratio estimates, the median minimum effective-sample-size fraction is
$\CartPoleControlMedianESS$, and the median maximum tempered weight is
$\CartPoleControlMedianMaxWeight$. 

\begin{figure}[!t]
\centering
\includegraphics[width=.94\linewidth]
{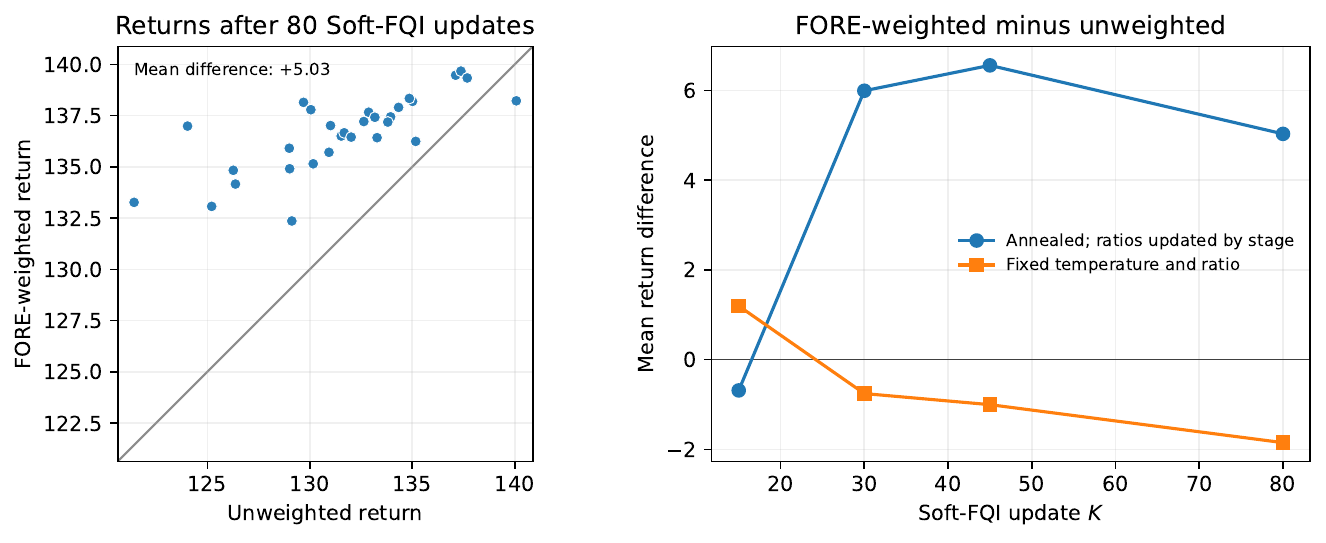}
\caption{\textbf{Continuous-state experiment with learned occupancy ratios.}
Left: weighted versus unweighted raw discounted return after $K=80$ soft
Bellman updates; the diagonal marks equality. Right: mean paired return
differences during training. Within each of the 30 datasets, the methods share
the offline data, feature map, initialization, and $Q$-projection budget.}
\label{fig:cartpole-soft-fqi-control}
\end{figure}

\paragraph{Sensitivity to weight tempering and hard-FQI reweighting.}
We evaluate weight tempering on 100 additional training datasets; see
Appendix~\ref{app:cartpole-soft-fqi-sensitivity}. At $K=80$, the $T_w=5$
specification again improves raw return by
$\CartPoleSensitivityKEightyTfiveGain$ points (95\% CI
$[\CartPoleSensitivityKEightyTfiveCILower,
\CartPoleSensitivityKEightyTfiveCIUpper]$)
and KL-regularized return by
$\CartPoleSensitivityKEightyTfiveSoftGain$ points (95\% CI
$[\CartPoleSensitivityKEightyTfiveSoftCILower,
\CartPoleSensitivityKEightyTfiveSoftCIUpper]$).
For tempering exponents $0,0.1,0.2,0.5,1$, the paired raw-return differences
are $0,3.37,5.14,6.88,-16.17$, respectively. This nonmonotonic pattern
highlights the finite-sample instability of untempered estimated weights and
the benefit of moderate tempering. Starting from the same
exponent-$0.2$, $K=200$ estimate, refreshing the occupancy ratio during hard
FQI rather than switching to unweighted hard updates improves raw return by
$\CartPoleSensitivityHardOneGain$ points after one hard update and
$\CartPoleSensitivityHardHundredGain$ points after 100.

\FloatBarrier
\section{Discussion}

Occupancy-reweighted soft FQI aligns fitted updates with a discounted-occupancy
norm in which soft control is locally contractive around the soft-optimal
solution. This provides an alternative to Bellman completeness: rather than
requiring Bellman updates to remain inside the function class, the method
changes the regression weights so that projection preserves the relevant
local contraction. Temperature annealing then turns this local result into
convergence from arbitrary initialization by moving gradually from a
high-temperature problem to the desired target temperature. For
unregularized control, the resulting soft iterate can instead initialize a
hard FQI continuation.

The main practical requirements are adequate coverage for occupancy-ratio
estimation and sufficient approximation quality along the annealing path.
The method also introduces choices such as the initial temperature, cooling
schedule, and number of regression updates per temperature. When estimated
occupancy ratios are highly variable, weight tempering can improve stability
by trading exact occupancy alignment for lower-variance regression
\citep{sinha2022experience}. Developing data-adaptive choices for the
temperature schedule and weighting scheme is an important direction for
future work.

Although our analysis focuses on soft FQI, the same principle may be useful
for actor--critic methods such as soft actor--critic (SAC)
\citep{haarnoja2018soft,sinha2022experience}. Soft FQI uses the exact
soft-greedy policy induced by the current \(Q\)-function, whereas SAC replaces
this policy by a learned actor. Extending the present analysis to SAC would
therefore require accounting for actor approximation and optimization error
in addition to occupancy-ratio and critic-estimation error. More broadly, the
results suggest that choosing regression weights to match the contraction
properties of the control update may be useful beyond fitted \(Q\)-iteration.

\FloatBarrier
\setlength{\parskip}{5.5pt}
\setlength{\abovedisplayskip}{7pt plus 2pt minus 5pt}
\setlength{\belowdisplayskip}{7pt plus 2pt minus 5pt}
\setlength{\abovedisplayshortskip}{0pt plus 3pt}
\setlength{\belowdisplayshortskip}{4pt plus 3pt minus 3pt}
\setlength{\textfloatsep}{20pt plus 2pt minus 4pt}
\bibliographystyle{plainnat}
\bibliography{ref}

\clearpage
\appendix

\setcounter{equation}{0}
\setcounter{theorem}{0}
\setcounter{lemma}{0}
\setcounter{figure}{0}
\setcounter{table}{0}
\setcounter{algorithm}{0}
\renewcommand{\theequation}{S\arabic{equation}}
\renewcommand{\thetheorem}{S\arabic{theorem}}
\renewcommand{\thelemma}{S\arabic{lemma}}
\renewcommand{\thefigure}{S\arabic{figure}}
\renewcommand{\thetable}{S\arabic{table}}
\renewcommand{\thealgorithm}{S\arabic{algorithm}}
\providecommand*{\theHequation}{}
\providecommand*{\theHtheorem}{}
\providecommand*{\theHlemma}{}
\providecommand*{\theHfigure}{}
\providecommand*{\theHtable}{}
\providecommand*{\theHalgorithm}{}
\renewcommand*{\theHequation}{S\arabic{equation}}
\renewcommand*{\theHtheorem}{S\arabic{theorem}}
\renewcommand*{\theHlemma}{S\arabic{lemma}}
\renewcommand*{\theHfigure}{S\arabic{figure}}
\renewcommand*{\theHtable}{S\arabic{table}}
\renewcommand*{\theHalgorithm}{S\arabic{algorithm}}

The appendix is organized by purpose. Appendix~A extends the related-work
discussion, and Appendix~B reports the controlled finite-state and
continuous-state experiments.
Appendix~C develops the population operator bounds and proves the
positive-temperature contraction and annealing results. Appendix~D proves
local convergence of hard FQI with refreshed occupancy weights and verifies
initialization from a fixed positive temperature. Appendix~E treats
discounted-occupancy ratio estimation. Appendix~F provides the
empirical-process tools used by the finite-sample analysis. Appendices~G and H
prove the finite-sample positive-temperature and hard-FQI guarantees,
respectively.

\section{Extended Related Work}
\label{app:related-work}

\textbf{Fitted value iteration and FQI.}
Classical analyses of fitted value iteration and FQI control projected Bellman
iteration through stability of the projected operator or propagation of
one-step approximation and estimation errors
\citep{gordon1995stable,tsitsiklis1996analysis,munos2005error,
munos2008finite,farahmand2010error,scherrer2014approximate}.
Under function approximation, convergence generally requires additional
structure linking the Bellman operator, fitted class, and sampling
distribution, such as Bellman completeness, small inherent Bellman error,
concentrability, or structured representations
\citep{chen2019information,amortila2020variant,wang2021exponential,
wang2021statistical,chang2022learning}. Difficulties persist even under linear
approximation in control: \citet{meyn2024projected} establish boundedness and
existence of a projected Bellman solution under a specialized training policy,
\citet{liu2025linear} obtain convergence to a bounded set, and
\citet{mehta2026optimistic} exhibit realizable examples with multiple
projected Bellman solutions. Our approach instead changes the projection norm:
occupancy reweighting yields local contraction around the soft-optimal
solution without Bellman completeness, including for nonlinear fitted classes,
while temperature annealing extends this result to convergence from arbitrary
initialization.

\textbf{Minimax and adversarial \(Q\)-learning.}
Minimax methods avoid Bellman completeness by testing Bellman residuals through
saddle-point objectives over auxiliary critic or test-function classes
\citep{dai2018sbeed,uehara2020minimax,uehara2021finite,
jin2021bellman,xie2021batch,uehara2023offline}. They can obtain guarantees
under \(Q\)-function realizability, including under partial coverage for
entropy-regularized control \citep{uehara2023offline}, but require sufficient
richness or identification through the auxiliary adversarial problem.
Occupancy-reweighted soft FQI instead retains standard least-squares Bellman
regressions and introduces no auxiliary critic into the \(Q\)-update.
Stability is obtained by changing the projection norm.

\textbf{Distribution correction and occupancy weighting.}
Stationary and emphatic weighting have long been used to stabilize off-policy
policy evaluation, while discounted occupancy-ratio methods, such as DualDICE
and GenDICE, estimate related distribution corrections
\citep{mahmood2015emphatic,sutton2016emphatic,yu2015convergence,
hallak2017consistent,geladaBellemare2019CovariateShift,
nachum2019dualdice,zhang2020gendice,uehara2020minimax}.
Most closely related is occupancy-weighted FQE, which reweights each fitted
Bellman regression by the occupancy ratio of a fixed evaluation policy and
thereby obtains global contraction in that policy's occupancy norm
\citep{vdLaanKallus2025FQE}. A related control precedent is
\citet{sinha2022experience}, who use estimates of the current-policy occupancy
ratio to reweight critic updates in an actor--critic algorithm, motivated by
contraction in fitted policy evaluation.

These policy-evaluation arguments do not directly establish convergence for
control, where the policy and its occupancy vary with \(Q\). We extend the
occupancy-weighting principle to soft control by showing that the local
behavior around the soft-optimal solution inherits the contraction of policy
evaluation. KL regularization controls the size of the resulting local
contraction region
\citep{ziebart2008maximum,ziebart2010causal,neu2017unified,
geist2019theory,haarnoja2017reinforcement,haarnoja2018soft}, while temperature
annealing extends local convergence to convergence from arbitrary
initialization. This role of temperature differs from the bias--variance
tuning studied by \citet{uehara2023offline}.

\textbf{Trust-region policy optimization.}
Our local analysis is related in spirit to trust-region policy optimization
(TRPO) \citep{schulman2015trust}. TRPO constrains successive policy changes so
that a surrogate objective based on the current-policy visitation distribution
remains accurate under distribution shift. Its reference policy therefore
changes across iterations, whereas \(\pi_{\mathrm{ref}}\) is fixed throughout
our analysis. Our setting instead concerns fitted \(Q\)-iteration: occupancy
reweighting selects a regression norm that supports local contraction, while
temperature annealing connects successive contraction regions without an
explicit trust-region constraint.

\section{Controlled Experiments}
\label{app:paired-mechanism}

\subsection{Finite-State Constructions}

We construct two ergodic Garnet MDPs \citep{bhatnagar2009natural} with two
actions, branching factor one, discount $\gamma=0.95$, and 10\% uniform
exploration in the behavior policy. The transition kernels include uniform
teleportation to ensure ergodicity, and rewards are sampled from
$[-r_{\max},r_{\max}]$. The weighting construction has six states,
teleportation probability $0.05$, and $r_{\max}=0.14$; the annealing
construction has ten states, teleportation probability $0.03$, and
$r_{\max}=0.06$. Both use behavior-policy logit scale $1.5$.

The reference policy is uniform. For any policy $\pi$, the regression
distribution is its reference-restart occupancy
\[
\mu_\pi
=
(1-\gamma)(\nu_0\otimes\pi_{\mathrm{ref}})
+\gamma\mu_\pi P_\pi,
\]
where $\nu_0$ is the behavior state marginal. The fitted feature span contains
an intercept, the optimal $Q$-functions at
\[
\tau\in\{0.8,0.267773,0.089628,0.03\},
\]
the hard-optimal $Q$-function, and additional nuisance directions. The maximum
approximation error along this temperature path is below
$5.2\times10^{-15}$, whereas designated off-path Bellman images have
normalized projection errors $0.160$ and $0.516$ in the weighting and
annealing constructions, respectively. Thus the fitted classes contain the
target paths but are not Bellman complete.

The annealed recursion performs 400 Bellman updates at each of the four
temperatures. The direct comparator performs the same total of $1{,}600$
updates at $\tau=0.03$. At the beginning of temperature stage $j$, let
$\bar\pi_j=\pi_{\tau_j,Q_{j,0}}$. On each FORE training fold, the occupancy
iteration is
\[
\widehat\mu_j^{(\ell+1)}(s',a')
=
(1-\gamma)\widehat\nu_S(s')\pi_{\mathrm{ref}}(a'\mid s')
+
\gamma\widehat m_j^{(\ell)}(s')\bar\pi_j(a'\mid s'),
\qquad
\widehat w_j^{(\ell+1)}
=
\frac{\widehat\mu_j^{(\ell+1)}}{\widehat\nu}.
\]
The resulting ratio is held fixed within each temperature stage. FORE is
two-fold cross-fitted, with each estimated ratio used for Bellman regression
on the held-out fold. Raw weights are normalized to have mean one. The
tempered method uses
\[
w_{T_w=5}
=
\frac{w_{\mathrm{raw}}^{1/5}}
{\E_{\mathrm{fit}}[w_{\mathrm{raw}}^{1/5}]}.
\]

Hard FQI starts from the annealed soft endpoint. At each hard Bellman update,
we form the current greedy policy, re-estimate its reference-restart occupancy
ratio, and use the resulting weights in the next regression.

\subsection{Population Results}

For the weighting construction, the Jacobian of the population Bellman
projection under the offline state-action distribution at
$Q_{0.03}^\star$ has dominant eigenvalue $-1.190$ and spectral radius $1.190$.
Initializations along its unstable eigenvector at sup-norm radii
$10^{-3}$, $10^{-5}$, and $10^{-7}$ enter a period-two cycle. Projection under
the reference-restart occupancy instead has spectral radius $0.95$ and
converges.

For the annealing construction, direct iteration at $\tau=0.03$ fails for 190
of 4,096 initializations. The four-stage temperature schedule converges from
all 4,096, with maximum normalized error below $9.7\times10^{-9}$. Starting
from the annealed endpoint, hard FQI with refreshed reference-restart
weighting converges to the hard optimum in both constructions.

\subsection{Finite-Sample Results}

We generate 150 independent transition datasets at
$n\in\{2{,}000,5{,}000,10{,}000\}$ and 300 at $n=20{,}000$. Within each
dataset, all methods use the same transitions and initialization. In the
weighting construction, we initialize along the unstable eigenvector of the
offline population map at sup-norm distance $10^{-5}$ from
$Q_{0.03}^\star$. In the annealing construction, we use a fixed initialization
for which direct low-temperature population iteration cycles.

Target-neighborhood success is defined by normalized $Q$-error at most $0.05$.

\begin{table}[H]
\centering
\caption{Positive-temperature endpoint at $n=20{,}000$ across 300 independent
datasets. Errors are median [25th, 75th percentiles], and success denotes
normalized $Q$-error at most $0.05$.}
\label{tab:soft-fqi-finite-sample}
\scriptsize
\resizebox{\linewidth}{!}{%
\begin{tabular}{lllrrr}
\toprule
Setting & Method & Schedule & Normalized $Q$ error & Success (\%) & Failure (\%) \\ 
\midrule
Annealing & Cross-fitted FORE & Annealed & 0.0063 [0.0035, 0.0113] & 100.0 & 0.0 \\
Annealing & Cross-fitted FORE & Direct & 0.0934 [0.0846, 0.1008] & 6.0 & 0.0 \\
Annealing & Cross-fitted FORE, $T_w=5$ & Annealed & 0.0042 [0.0022, 0.0075] & 100.0 & 0.0 \\
Annealing & Cross-fitted FORE, $T_w=5$ & Direct & 0.0037 [0.0021, 0.0066] & 100.0 & 0.0 \\
Annealing & Unweighted & Annealed & 0.0043 [0.0024, 0.0074] & 100.0 & 0.0 \\
Annealing & Unweighted & Direct & 0.0043 [0.0024, 0.0074] & 100.0 & 0.0 \\
Weighting & Cross-fitted FORE & Annealed & 0.0021 [0.0012, 0.0038] & 100.0 & 0.0 \\
Weighting & Cross-fitted FORE & Direct & 0.0021 [0.0012, 0.0038] & 100.0 & 0.0 \\
Weighting & Cross-fitted FORE, $T_w=5$ & Annealed & 0.0021 [0.0011, 0.0036] & 100.0 & 0.0 \\
Weighting & Cross-fitted FORE, $T_w=5$ & Direct & 0.0021 [0.0011, 0.0036] & 100.0 & 0.0 \\
Weighting & Unweighted & Annealed & 0.1674 [0.1459, 0.1904] & 0.0 & 0.0 \\
Weighting & Unweighted & Direct & 0.1688 [0.1459, 0.1918] & 0.0 & 0.0 \\
\bottomrule
\end{tabular}
}
\end{table}

\begin{figure}[H]
\centering
\includegraphics[width=.99\linewidth]
{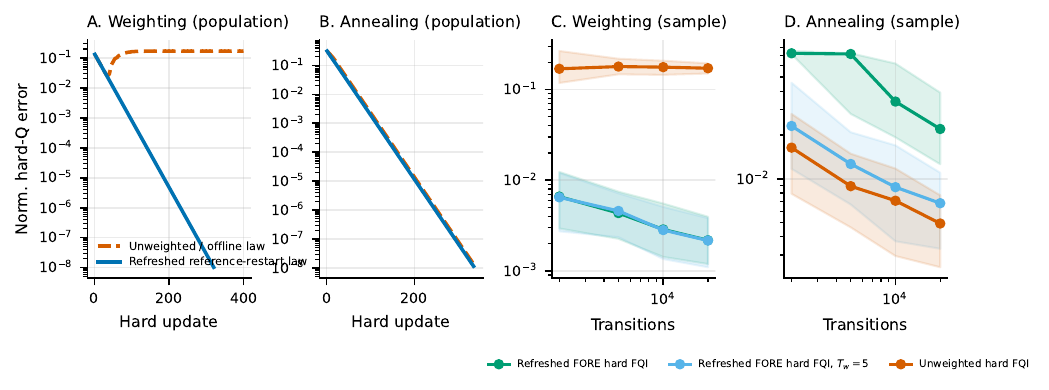}
\caption{\textbf{Hard FQI with refreshed reference-restart weights.}
(A)--(B) Population hard-$Q$ error after the annealed soft endpoint under
offline and reference-restart projection.
(C)--(D) Median hard-$Q$ error and interquartile range across independent
transition datasets for unweighted FQI, raw FORE, and $T_w=5$.}
\label{fig:soft-fqi-hardmax}
\end{figure}

\begin{table}[H]
\centering
\caption{Hard-FQI endpoint after 400 updates from the annealed soft endpoint
at $n=20{,}000$, across 300 independent datasets. Policy-mismatch mass is
measured under the hard-optimal reference-restart state marginal, and regret
is normalized by $\|\bar r\|_\infty/(1-\gamma)$. Errors are median
[25th, 75th percentiles].}
\label{tab:soft-fqi-hardmax}
\scriptsize
\resizebox{\linewidth}{!}{%
\begin{tabular}{lllrrrrr}
\toprule
Setting & Method & Schedule & Normalized $Q$ error & Policy mismatch & Normalized regret & Success (\%) & Failure (\%) \\ 
\midrule
Annealing & Cross-fitted refreshed FORE & Annealed & 0.0220 [0.0126, 0.0391] & 0.0000 & 0.0000 & 86.0 & 0.0 \\
Annealing & Cross-fitted refreshed FORE, $T_w=5$ & Annealed & 0.0068 [0.0033, 0.0110] & 0.0000 & 0.0000 & 99.7 & 0.0 \\
Annealing & Unweighted & Annealed & 0.0049 [0.0025, 0.0077] & 0.0000 & 0.0000 & 100.0 & 0.0 \\
Weighting & Cross-fitted refreshed FORE & Annealed & 0.0022 [0.0012, 0.0040] & 0.0000 & 0.0000 & 100.0 & 0.0 \\
Weighting & Cross-fitted refreshed FORE, $T_w=5$ & Annealed & 0.0022 [0.0011, 0.0038] & 0.0000 & 0.0000 & 100.0 & 0.0 \\
Weighting & Unweighted & Annealed & 0.1717 [0.1502, 0.1945] & 0.1924 & 0.3838 & 0.0 & 0.0 \\
\bottomrule
\end{tabular}
}
\end{table}

\begin{table}[H]
\centering
\caption{FORE and weight diagnostics at $n=20{,}000$. Ratio $L^2$ error is
computed against the exact occupancy ratio. Effective-sample-size fractions
and weight tails summarize the variability of the estimated regression
weights.}
\label{tab:soft-fqi-fore-performance}
\scriptsize
\resizebox{\linewidth}{!}{%
\begin{tabular}{llllrrrrrr}
\toprule
Endpoint & Setting & Method & Schedule & rESS & Median max $w$ & 95th pct. max $w$ & Applied-$w$ $L^2$ vs. oracle & Runtime (s) & Failure (\%) \\ 
\midrule
Fixed Positive & Annealing & Cross-fitted FORE & Annealed & 0.190 & 18.4 & 759.9 & 0.302 & 1.71 & 0.0 \\
Fixed Positive & Annealing & Cross-fitted FORE & Direct & 0.608 & 14.5 & 790.3 & 0.119 & 1.75 & 0.0 \\
Fixed Positive & Annealing & Cross-fitted FORE, $T_w=5$ & Annealed & 0.923 & 2.1 & 4.5 & 1.769 & 1.73 & 0.0 \\
Fixed Positive & Annealing & Cross-fitted FORE, $T_w=5$ & Direct & 0.968 & 1.8 & 4.1 & 0.622 & 1.81 & 0.0 \\
Fixed Positive & Weighting & Cross-fitted FORE & Annealed & 0.220 & 11.9 & 13.2 & 0.126 & 1.77 & 0.0 \\
Fixed Positive & Weighting & Cross-fitted FORE & Direct & 0.214 & 12.1 & 13.5 & 0.124 & 1.71 & 0.0 \\
Fixed Positive & Weighting & Cross-fitted FORE, $T_w=5$ & Annealed & 0.903 & 2.0 & 2.0 & 1.618 & 1.81 & 0.0 \\
Fixed Positive & Weighting & Cross-fitted FORE, $T_w=5$ & Direct & 0.903 & 2.0 & 2.0 & 1.649 & 1.81 & 0.0 \\
Hardmax & Annealing & Cross-fitted refreshed FORE & Annealed To Hard & 0.013 & 109.2 & 2416.9 & 1.019 & 1.38 & 0.0 \\
Hardmax & Annealing & Cross-fitted refreshed FORE, $T_w=5$ & Annealed To Hard & 0.886 & 3.8 & 7.8 & 8.210 & 1.25 & 0.0 \\
Hardmax & Weighting & Cross-fitted refreshed FORE & Annealed To Hard & 0.207 & 12.3 & 13.7 & 0.125 & 1.16 & 0.0 \\
Hardmax & Weighting & Cross-fitted refreshed FORE, $T_w=5$ & Annealed To Hard & 0.889 & 2.0 & 2.1 & 1.673 & 1.20 & 0.0 \\
\bottomrule
\end{tabular}
}
\end{table}

In the weighting construction, occupancy reweighting removes the instability
induced by projection under the offline state-action distribution. In the
annealing construction, raw FORE success increases from $6\%$ under direct
low-temperature iteration to $100\%$ under annealing. Tempering substantially
reduces weight variability at the hard endpoint: the median
effective-sample-size fraction increases from $0.013$ for raw FORE to $0.886$
for $T_w=5$, while the 95th percentile of the maximum weight decreases from
$2{,}416.9$ to $7.8$. The corresponding success rate increases from $86.0\%$
to $99.7\%$.

\subsection{Continuous-State Experiment with Learned Occupancy Ratios}
\label{app:cartpole-soft-fqi-control}

The continuous-state experiment uses the stochastic continuing-reset
CartPole-inspired process from
Section~\ref{sec:cartpole-soft-fqi-control}. Each of the 30 datasets contains
$25{,}000$ transitions from 512 behavior trajectories after a 2,000-step
burn-in. The state is
\[
s=(x,\dot x,\theta,\dot\theta).
\]
Transitions follow the standard CartPole dynamics with additive Gaussian
noise of standard deviation $0.001$. Failed trajectories are immediately
restarted from $\operatorname{Unif}([-0.05,0.05]^4)$. The reward is
\[
r(s)
=
\left(2-\frac{|\theta|}{12\pi/180}\right)
\left(2-\frac{|x|}{2.4}\right)-1.
\]
The behavior action-one probability is logistic with coefficient vector
$(0.8,1.3,12,2)$, and $\gamma=0.98$.

All methods use a fixed 64-dimensional random Fourier feature class for $Q$,
zero initialization, and ridge-regularized Bellman regression. The reference
policy is uniform. FORE uses a separate 128-dimensional random-feature model.
Estimated weights are bounded to $[10^{-4},50]$, transformed using $T_w=5$,
and normalized to have mean one. FORE and the $Q$-regressions are fit on the
same offline dataset.

The direct path performs 80 Bellman updates at $\tau=0.128$. The annealed path
uses temperatures
\[
(2,0.8,0.32,0.128)
\]
for $(15,15,15,35)$ updates. For each path, we compare unweighted Bellman
regression with $T_w=5$ FORE weighting. The direct weighted method estimates
its occupancy ratio once, whereas the annealed method re-estimates the ratio
at the beginning of each temperature stage. Each soft method is followed by
five hard-FQI updates, with the weighted methods refreshing the occupancy
ratio before each hard update. The hard-only comparator performs 85
unweighted hard Bellman updates from $Q_0$.

The $K=80$ comparison uses 30 independent training datasets. Each fitted
policy is evaluated using 2,000 paired Monte Carlo trajectories of horizon
700. We report raw discounted return and the uniform-reference regularized
return
\[
\sum_{t=0}^{699}\gamma^t
\left\{
r_t-\tau\log\bigl(2\pi_{\tau,Q}(A_t\mid S_t)\bigr)
\right\}.
\]
Confidence intervals are obtained by a paired hierarchical bootstrap over
training datasets and evaluation trajectories.

\subsubsection{Sensitivity to Weight Tempering and Hard-FQI Reweighting}
\label{app:cartpole-soft-fqi-sensitivity}

We further compare tempering exponents
\[
p\in\{0,0.1,0.2,0.5,1\},
\qquad
w_p=\frac{w^p}{\E[w^p]},
\]
using 100 independent training datasets. Each positive exponent uses a
policy-specific occupancy-ratio estimate at the beginning of each temperature
stage. We evaluate the annealed paths after $K=80$ and $K=200$ Bellman
updates, extending the final temperature stage to the latter endpoint.

\begin{figure*}[t]
\centering
\includegraphics[width=\textwidth]
{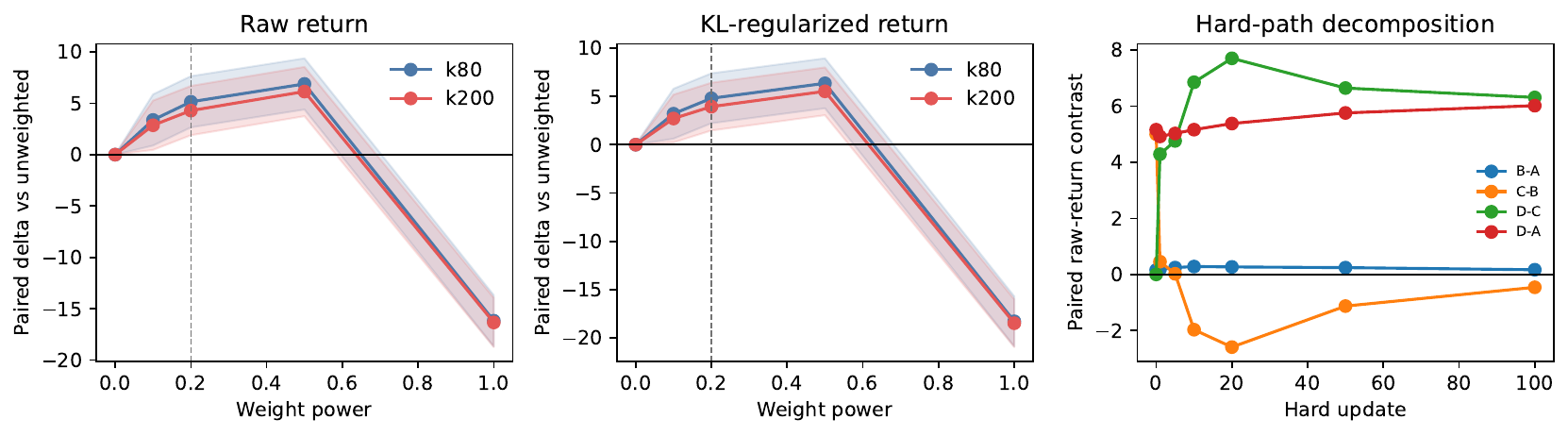}
\caption{\textbf{Sensitivity to weight tempering and hard-FQI reweighting.}
Left and middle: paired gains over unweighted regression at $K=80$ and
$K=200$ for raw and uniform-reference regularized return. Shading gives
simultaneous 95\% bootstrap bands over the positive tempering exponents; the
dashed line marks $T_w=5$, corresponding to $p=0.2$.
Right: paired raw-return contrasts for hard FQI starting from the hard-only,
unweighted-annealed, and $p=0.2$ annealed endpoints, with or without refreshed
occupancy weighting during the hard updates.}
\label{fig:cartpole-soft-fqi-sensitivity}
\end{figure*}

At $K=80$, paired raw-return gains over unweighted regression are $3.37$,
$5.14$, and $6.88$ for $p=0.1$, $0.2$, and $0.5$, respectively, and $-16.17$
for $p=1$. At $K=200$, the corresponding gains are $2.86$, $4.29$, $6.14$,
and $-16.34$. The regularized-return comparisons have the same ordering.
For $T_w=5$, the $K=80$ raw-return gain is
$\CartPoleSensitivityKEightyTfiveGain$ (95\% CI
$[\CartPoleSensitivityKEightyTfiveCILower,
\CartPoleSensitivityKEightyTfiveCIUpper]$).

Weight variability increases sharply with the tempering exponent. At $K=200$,
the median effective-sample-size fractions are $0.784$, $0.536$, and $0.014$
for $p=0.2$, $0.5$, and $1$, respectively, with corresponding median maximum
weights $2.49$, $6.88$, and $147$. The empirical Jacobian spectral radius is
below one for all $p=0.2$ fits, but exceeds one for some $p=0.5$ fits and most
$p=1$ fits.

Finally, starting from the same $p=0.2$ annealed estimate at $K=200$,
refreshing the occupancy ratio during hard FQI improves raw return relative to
unweighted hard updates by $4.29$ points after one update (95\% CI
$[4.10,4.48]$) and $6.31$ points after 100 updates (95\% CI
$[5.77,6.85]$). After 100 hard updates, the fully reweighted path exceeds the
hard-only comparator by $6.02$ points (95\% CI $[5.37,6.67]$).

\FloatBarrier
\section{Population Convergence: Operator Bounds and Proofs}
\label{app:local-geometry-derivative-bounds}

This appendix proves the population results in
Section~\ref{sec:population-theory}. Regularity examples and
discounted-occupancy identities precede the proofs of local convergence with
fixed within-stage occupancy weights and global convergence along the
geometric temperature schedule. Appendix~\ref{app:hard-polishing} treats the
hard Bellman operator, whose analysis additionally requires an action-gap
condition.
\subsection{Examples for the Local Regularity Conditions}
\label{app:local-regularity-examples}

The following examples verify Conditions~\ref{cond::convex}
and~\ref{cond::supnorm}.

\paragraph{Function-class regularity.}
Condition~\ref{cond::convex} holds for continuous finite-dimensional models
with a compact convex parameter set. On a compact domain, it also holds for
bounded H\"older balls, Sobolev balls compactly embedded in the continuous
functions, and bounded RKHS balls with a continuous kernel.

\paragraph{Difference-class interpolation.}
Condition~\ref{cond::supnorm} holds for several standard smooth function
classes under the design-measure assumptions of the corresponding
interpolation inequality. For a class uniformly bounded in a reproducing
kernel Hilbert space (RKHS) norm, suppose the eigenfunctions of the kernel
integral operator on \(L^2(\nu_{\mathrm{ref}})\) are uniformly bounded and its
eigenvalues satisfy \(\lambda_j\lesssim j^{-2r}\) with \(r>1/2\). A standard
spectral truncation argument gives
\(\alpha=(2r-1)/(2r)\in(0,1)\); see, for example,
\citet[Lemma~5.1]{mendelson2010regularization}. Related bounds hold for
bounded signed convex hulls of uniformly bounded bases under the hypotheses of
\citet[Lemma~2]{van2014uniform}. For bounded \(d\)-variate H\"older balls of
smoothness \(s>0\), the Gagliardo--Nirenberg interpolation inequality gives
\(\alpha=2s/(2s+d)\) \citep{triebel2006theory}. For bounded Sobolev balls of
smoothness \(s>d/2\), the Gagliardo--Nirenberg--Sobolev inequality gives
\(\alpha=1-d/(2s)\) \citep{adams2003sobolev}. In both cases, the
\(L^2(\nu_{\mathrm{ref}})\) formulation follows when the design density is
bounded below on its support.

\paragraph{Finite-dimensional linear classes.}
Suppose
\(\mathcal F\subseteq
\{(s,a)\mapsto\phi(s,a)^\top\theta:\theta\in\mathbb R^p\}\),
\(\|\phi(s,a)\|_2\le B_\phi\), and
\[
\E_{\nu_{\mathrm{ref}}}[\phi\phi^\top]\succeq\lambda_0I_p.
\]
Then, for \(H=\phi^\top\theta\),
\[
\|H\|_\infty
\le
B_\phi\|\theta\|_2
\le
\frac{B_\phi}{\sqrt{\lambda_0}}
\|H\|_{2,\nu_{\mathrm{ref}}},
\]
so, on a class with bounded \(L^2(\nu_{\mathrm{ref}})\)-diameter, the same inequality
implies Condition~\ref{cond::supnorm} for any fixed
\(\alpha\in(0,1)\), after adjusting \(C_0\).
\subsection{Operator Identities}

The reference-restart occupancy is a natural measure for policy evaluation.
For a fixed policy \(\pi\), let
\(\mu_\pi\) denote the reference-restart occupancy obtained by
replacing \(\pi_{\tau,Q}\) with \(\pi\) in
\eqref{eq:stage-frozen-occupancy}.

\begin{lemma}[Reference-restart minorization and interpolation]
\label{lem:reference-restart-interpolation}
For every policy \(\pi\),
\[
\mu_\pi\ge(1-\gamma)\nu_{\mathrm{ref}}.
\]
Under Condition~\ref{cond::supnorm}, this implies
\begin{equation}
\label{eq:reference-restart-interpolation}
\|H\|_\infty
\le
C_{\mathrm{int}}\|H\|_{2,\mu_\pi}^{\alpha},
\qquad
H\in\mathcal F-\mathcal F,
\end{equation}
where \(C_{\mathrm{int}}=C_0(1-\gamma)^{-\alpha/2}\).
\end{lemma}

\begin{proof}
The discounted balance equation is
\[
\mu_\pi=(1-\gamma)\nu_{\mathrm{ref}}+\gamma\mu_\pi P_\pi.
\]
Since the second measure on the right is nonnegative,
\(\mu_\pi\ge(1-\gamma)\nu_{\mathrm{ref}}\). Consequently,
\(\|H\|_{2,\nu_{\mathrm{ref}}}
\le(1-\gamma)^{-1/2}\|H\|_{2,\mu_\pi}\).
Condition~\ref{cond::supnorm} therefore yields
\eqref{eq:reference-restart-interpolation}.
\end{proof}

\begin{lemma}[Policy-evaluation contraction in discounted occupancy]
\label{prop:discounted-policy-evaluation-contraction}
For every \(H\in L^2(\mu_\pi)\),
\[
\|\gamma P_\pi H\|_{2,\mu_\pi}
\le
\sqrt\gamma\,\|H\|_{2,\mu_\pi}.
\]
Consequently, the KL-regularized Bellman evaluation operator satisfies
\[
\|\mathcal T_\tau^\pi f-\mathcal T_\tau^\pi g\|_{2,\mu_\pi}
\le
\sqrt\gamma\,\|f-g\|_{2,\mu_\pi}
\]
for all bounded \(f,g\).
\end{lemma}

\begin{proof}
Conditional Jensen's inequality and the discounted balance equation give
\[
\begin{aligned}
\|\gamma P_\pi H\|_{2,\mu_\pi}^2
&=\gamma^2\mu_\pi\{(P_\pi H)^2\}\\
&\le \gamma^2(\mu_\pi P_\pi)H^2\\
&=\gamma\left\{
\mu_\pi H^2-(1-\gamma)\nu_{\mathrm{ref}}H^2
\right\}
\le \gamma\mu_\pi H^2.
\end{aligned}
\]
Taking square roots proves the first claim. Because the reward and
KL-penalty terms cancel when two evaluation updates are subtracted,
\(\mathcal T_\tau^\pi f-\mathcal T_\tau^\pi g
=\gamma P_\pi(f-g)\), which proves the second claim.
\end{proof}

\begin{lemma}[First derivative of the soft Bellman operator]
\label{lem:first-derivative}
Fix \(\tau>0\). For bounded \(Q\) and \(H\), view
\(\mathcal T_\tau:L^\infty(\mathcal S\times\mathcal A)
\to L^\infty(\mathcal S\times\mathcal A)\). Its Fr\'echet derivative at
\(Q\) is
\[
D\mathcal T_\tau(Q)[H] \;=\; \gamma\,P_{\pi_{\tau,Q}}H.
\]
If, in addition, \(H\in L^2(\mu_\tau^\star)\), then
$\|D\mathcal T_\tau(Q_\tau^\star)[H]\|_{2,\mu_\tau^\star}
\;\le\;
\sqrt\gamma\,\|H\|_{2,\mu_\tau^\star}$.
\end{lemma}

\begin{proof}[Proof of Lemma~\ref{lem:first-derivative}]
Fix bounded \(Q,H\). For each state \(s\), the gradient of
\[
q\mapsto
\tau\log\sum_{b\in\mathcal A}
\pi_{\mathrm{ref}}(b\mid s)e^{q_b/\tau}
\]
is the reference-weighted softmax. Its directional derivative along \(H\) is
\[
\sum_{b\in\mathcal A}\pi_{\tau,Q}(b\mid s)H(s,b).
\]
The second directional derivative along a bounded increment \(K\) equals
\(\tau^{-1}\Var_{\pi_{\tau,Q+tK}}\{K(s,A)\}\). Taylor's theorem therefore
gives, uniformly in \(s\),
\[
|V_{\tau,Q+K}(s)-V_{\tau,Q}(s)
-\E_{\pi_{\tau,Q}(\cdot\mid s)}K(s,A)|
\le
\frac{\|K\|_\infty^2}{2\tau},
\]
because \(\Var\{K(s,A)\}\le\|K\|_\infty^2\). Conditional expectation
preserves this uniform quadratic remainder. Hence \(\mathcal T_\tau\) is
Fr\'echet differentiable in the supremum norm, with
\begin{align*}
D\mathcal T_\tau(Q)[H](s,a)
&=
\gamma\,\E_{S'\sim P(\cdot\mid s,a)}
\Big[\sum_b
\pi_{\tau,Q}(b\mid S')\,H(S',b)\Big]
\\[0.3em]
&=
\gamma\,P_{\pi_{\tau,Q}}H(s,a).
\end{align*}
This establishes the derivative formula. At \(Q=Q_\tau^\star\),
\(\pi_{\tau,Q}=\pi_\tau^\star\), and
\(\mu_\tau^\star\) is its reference-restart occupancy.
Lemma~\ref{prop:discounted-policy-evaluation-contraction}
therefore gives the stated norm bound.
\end{proof}

\subsection{Positive-Temperature Contraction and Annealing}
\label{app:stage-frozen-population-annealing}

The reference restart gives a common interpolation constant at every
temperature. Holding the projection
measure fixed within a stage then makes the projected Bellman map contractive
when the Bellman policy is sufficiently close to the policy induced by the
initial iterate.

Recall \(\mu_{\tau,Q^{\mathrm{init}}}\) and
\(\mathcal U_{\tau,Q^{\mathrm{init}}}\) from
\eqref{eq:stage-frozen-occupancy}--\eqref{eq:stage-frozen-map}.
Lemma~\ref{lem:reference-restart-interpolation}, applied to
\(\pi_{\tau,Q^{\mathrm{init}}}\), gives
\begin{equation}
\label{eq:stage-frozen-interpolation}
\|H\|_\infty
\le C_{\mathrm{int}}
\|H\|_{2,\mu_{\tau,Q^{\mathrm{init}}}}^\alpha,
\qquad H\in\mathcal F-\mathcal F.
\end{equation}

For a fixed temperature, define
\[
\varepsilon_\tau
:=
\inf_{f\in\mathcal F}\|f-Q_\tau^\star\|_\infty.
\]
Compactness in Condition~\ref{cond::convex} ensures that the infimum is
attained. For \(Q^{\mathrm{init}}\in\mathcal F\), write
\[
\|H\|_{\tau,Q^{\mathrm{init}}}
:=
\|H\|_{2,\mu_{\tau,Q^{\mathrm{init}}}}.
\]
For \(R_{\mathrm{loc}},\varepsilon\ge0\), define
\[
d(R_{\mathrm{loc}},\varepsilon)
:=
C_{\mathrm{int}}(R_{\mathrm{loc}}+\varepsilon)^\alpha+\varepsilon.
\]

\begin{lemma}[One stage with fixed reference-restart occupancy]
\label{prop:stage-frozen-affine-contraction}
Assume Conditions~\ref{cond::convex} and~\ref{cond::supnorm}. Fix
\(\tau>0\), \(Q^{\mathrm{init}}\in\mathcal F\), and
\(R_{\mathrm{loc}}>0\). If
\[
\|Q^{\mathrm{init}}-Q_\tau^\star\|_{\tau,Q^{\mathrm{init}}}
\le R_{\mathrm{loc}},
\qquad
\|Q-Q_\tau^\star\|_{\tau,Q^{\mathrm{init}}}
\le R_{\mathrm{loc}},
\]
then
\begin{equation}
\label{eq:stage-frozen-affine-contraction}
\|\mathcal U_{\tau,Q^{\mathrm{init}}}(Q)-Q_\tau^\star
\|_{\tau,Q^{\mathrm{init}}}
\le
\rho_{\tau}
(R_{\mathrm{loc}},\varepsilon_\tau)
\|Q-Q_\tau^\star\|_{\tau,Q^{\mathrm{init}}}
+2\varepsilon_\tau.
\end{equation}
In particular, if \(Q_\tau^\star\in\mathcal F\) and
\(\rho_{\tau}(R_{\mathrm{loc}},0)<1\), then
\(\mathcal U_{\tau,Q^{\mathrm{init}}}\) is a contraction toward
\(Q_\tau^\star\) throughout this radius-\(R_{\mathrm{loc}}\) neighborhood.
\end{lemma}

\begin{proof}
Let \(Q_\tau^\circ\in\mathcal F\) attain
\(\varepsilon:=\varepsilon_\tau\), and abbreviate
\(\mu:=\mu_{\tau,Q^{\mathrm{init}}}\). For
\(Q_t:=Q_\tau^\star+t(Q-Q_\tau^\star)\),
\eqref{eq:stage-frozen-interpolation} gives
\[
\|Q^{\mathrm{init}}-Q_\tau^\circ\|_\infty
\le C_{\mathrm{int}}(R_{\mathrm{loc}}+\varepsilon)^\alpha,
\qquad
\|Q-Q_\tau^\circ\|_\infty
\le C_{\mathrm{int}}(R_{\mathrm{loc}}+\varepsilon)^\alpha.
\]
For \(0\le t\le1\),
\[
Q_t-Q^{\mathrm{init}}
=(1-t)(Q_\tau^\star-Q_\tau^\circ)
+t(Q-Q_\tau^\circ)
+Q_\tau^\circ-Q^{\mathrm{init}}.
\]
The triangle inequality and
\(\|Q_\tau^\star-Q_\tau^\circ\|_\infty=\varepsilon\)
therefore give
\[
\|Q_t-Q^{\mathrm{init}}\|_\infty
\le 2C_{\mathrm{int}}(R_{\mathrm{loc}}+\varepsilon)^\alpha+\varepsilon.
\]
The statewise range of \(Q_t-Q^{\mathrm{init}}\) is therefore at most
\(4d(R_{\mathrm{loc}},\varepsilon)\). The softmax formula implies
\begin{equation}
\label{eq:stage-frozen-policy-domination}
\pi_{\tau,Q_t}(a\mid s)
\le
\exp\{4d(R_{\mathrm{loc}},\varepsilon)/\tau\}
\pi_{\tau,Q^{\mathrm{init}}}(a\mid s).
\end{equation}

Set \(H:=Q-Q_\tau^\star\). Conditional Jensen's inequality,
\eqref{eq:stage-frozen-policy-domination}, and discounted balance under
the continuation policy \(\pi_{\tau,Q^{\mathrm{init}}}\) give
\begin{equation}
\label{eq:stage-frozen-integrand-contraction}
\begin{aligned}
\|\gamma P_{\pi_{\tau,Q_t}}H\|_{2,\mu}^2
&\le
\gamma^2\mu P_{\pi_{\tau,Q_t}}H^2\\
&\le
\gamma^2\exp\{4d(R_{\mathrm{loc}},\varepsilon)/\tau\}
\mu P_{\pi_{\tau,Q^{\mathrm{init}}}}H^2\\
&=
\gamma\exp\{4d(R_{\mathrm{loc}},\varepsilon)/\tau\}
\{\mu H^2-(1-\gamma)\nu_{\mathrm{ref}}H^2\}\\
&\le
\gamma\exp\{4d(R_{\mathrm{loc}},\varepsilon)/\tau\}
\|H\|_{2,\mu}^2.
\end{aligned}
\end{equation}
Lemma~\ref{lem:first-derivative} and the fundamental theorem of calculus give
\[
\mathcal T_\tau Q-Q_\tau^\star
=
\int_0^1\gamma P_{\pi_{\tau,Q_t}}H\,\mathrm dt.
\]
Minkowski's inequality and
\eqref{eq:stage-frozen-integrand-contraction} yield
\begin{equation}
\label{eq:stage-frozen-bellman-contraction}
\|\mathcal T_\tau Q-Q_\tau^\star\|_{2,\mu}
\le
\rho_{\tau}(R_{\mathrm{loc}},\varepsilon)
\|Q-Q_\tau^\star\|_{2,\mu}.
\end{equation}

Let \(U:=\mathcal U_{\tau,Q^{\mathrm{init}}}(Q)\). Metric
projection in
\(L^2(\mu)\), with \(Q_\tau^\circ\) as a feasible
comparator, gives
\[
\|U-Q_\tau^\circ\|_{2,\mu}
\le
\|\mathcal T_\tau Q-Q_\tau^\circ\|_{2,\mu}.
\]
The triangle inequality, this projection bound, and
\eqref{eq:stage-frozen-bellman-contraction} give
\[
\begin{aligned}
\|U-Q_\tau^\star\|_{2,\mu}
&\le
\|U-Q_\tau^\circ\|_{2,\mu}+\varepsilon\\
&\le
\|\mathcal T_\tau Q-Q_\tau^\circ\|_{2,\mu}+\varepsilon\\
&\le
\rho_{\tau}(R_{\mathrm{loc}},\varepsilon)
\|Q-Q_\tau^\star\|_{2,\mu}+2\varepsilon.
\end{aligned}
\]
This is \eqref{eq:stage-frozen-affine-contraction}. If \(\varepsilon=0\),
the additive term vanishes, and the strict contraction makes the
radius-\(R_{\mathrm{loc}}\) neighborhood invariant.
\end{proof}

\begin{proof}[Proof of Theorem~\ref{thm:positive-stage-frozen-local}]
Under Condition~\ref{cond::fitted-realizability},
\(\varepsilon_\tau=0\). Equations
\eqref{eq:annealing-local-modulus} and
\eqref{eq:positive-frozen-constants} therefore give
\[
\rho_{\tau}(R_{\mathrm{loc}},0)=\rho_{\tau}(R_{\mathrm{loc}}).
\]
The definition of \(R_\tau^{\max}\) in
Section~\ref{sec:local-geometry} implies
\(\rho_{\tau}(R_{\mathrm{loc}})<1\). The hypothesis places \(Q_0\) in the
radius-\(R_{\mathrm{loc}}\) neighborhood, both as the fixed iterate defining
the occupancy and as the first Bellman iterate.
Lemma~\ref{prop:stage-frozen-affine-contraction} gives
\[
\|Q_{k+1}-Q_\tau^\star\|_{\tau,Q_0}
\le
\rho_{\tau}(R_{\mathrm{loc}})
\|Q_k-Q_\tau^\star\|_{\tau,Q_0}
\]
whenever \(Q_k\) belongs to the same neighborhood. The strict contraction
makes this neighborhood invariant, and induction gives
\eqref{eq:positive-stage-frozen-rate} for every \(k\ge0\).
\end{proof}

The first annealing stage is globally contractive at sufficiently high
temperature. Let
\[
B_{\mathcal F}:=\sup_{Q\in\mathcal F}\|Q\|_\infty,
\qquad
Q_{\max}:=\frac{\|\bar r\|_\infty}{1-\gamma},
\qquad
D_{\mathcal F}
:=2\max\{2B_{\mathcal F},B_{\mathcal F}+Q_{\max}\}.
\]
Because
\(\min_aQ(s,a)\le V_{\tau,Q}(s)\le\max_aQ(s,a)\),
the fixed-point equation gives
\(\|Q_\tau^\star\|_\infty\le Q_{\max}\) for every \(\tau>0\). Define
\[
\tau_{\mathrm{init}}
:=
\frac{D_{\mathcal F}}
{\log\{(\rho_{\mathrm{init}})^2/\gamma\}}.
\]
Recall from Section~\ref{sec:population-annealing} that
\(\rho_{\mathrm{init}}=(1+\sqrt\gamma)/2\). The denominator is positive
because \(\rho_{\mathrm{init}}>\sqrt\gamma\).

\begin{lemma}[Global high-temperature initialization]
\label{prop:global-high-temperature}
Assume Condition~\ref{cond::convex}. If
\(\tau\ge\tau_{\mathrm{init}}\), then, for every
\(Q^{\mathrm{init}},Q\in\mathcal F\),
\begin{equation}
\label{eq:stage-frozen-global-contraction}
\|\mathcal U_{\tau,Q^{\mathrm{init}}}(Q)-Q_\tau^\star
\|_{\tau,Q^{\mathrm{init}}}
\le
\rho_{\mathrm{init}}
\|Q-Q_\tau^\star\|_{\tau,Q^{\mathrm{init}}}
+2\varepsilon_\tau.
\end{equation}
No realizability condition is required at this temperature.
\end{lemma}

\begin{proof}
For \(Q_t:=Q_\tau^\star+t(Q-Q_\tau^\star)\), boundedness gives
\[
\operatorname{range}_a\{Q_t(s,a)-Q^{\mathrm{init}}(s,a)\}
\le D_{\mathcal F}.
\]
Thus
\begin{equation}
\label{eq:stage-frozen-global-policy-domination}
\pi_{\tau,Q_t}(a\mid s)
\le
e^{D_{\mathcal F}/\tau}
\pi_{\tau,Q^{\mathrm{init}}}(a\mid s).
\end{equation}
Lemma~\ref{lem:first-derivative} and the fundamental theorem of calculus give
\[
\mathcal T_\tau Q-Q_\tau^\star
=
\int_0^1\gamma P_{\pi_{\tau,Q_t}}
(Q-Q_\tau^\star)\,\mathrm dt.
\]
Conditional Jensen's inequality, discounted balance under the continuation
policy \(\pi_{\tau,Q^{\mathrm{init}}}\), and
\eqref{eq:stage-frozen-global-policy-domination} give, for
\(H:=Q-Q_\tau^\star\),
\[
\|\gamma P_{\pi_{\tau,Q_t}}H\|_{
\tau,Q^{\mathrm{init}}}^2
\le
\gamma e^{D_{\mathcal F}/\tau}
\|H\|_{\tau,Q^{\mathrm{init}}}^2.
\]
Minkowski's inequality therefore gives
\[
\|\mathcal T_\tau Q-Q_\tau^\star
\|_{\tau,Q^{\mathrm{init}}}
\le
\left\{\gamma e^{D_{\mathcal F}/\tau}\right\}^{1/2}
\|Q-Q_\tau^\star\|_{\tau,Q^{\mathrm{init}}}.
\]
The definition of \(\tau_{\mathrm{init}}\) bounds the multiplier
by \(\rho_{\mathrm{init}}\). If
\(Q_\tau^\circ\) attains
\(\varepsilon_\tau\), metric projection gives
\[
\|\mathcal U_{\tau,Q^{\mathrm{init}}}(Q)
-Q_\tau^\circ\|_{\tau,Q^{\mathrm{init}}}
\le
\|\mathcal T_\tau Q-Q_\tau^\circ
\|_{\tau,Q^{\mathrm{init}}}.
\]
Because
\(\|Q_\tau^\circ-Q_\tau^\star\|_
{\tau,Q^{\mathrm{init}}}
\le\varepsilon_\tau\), the triangle inequality gives
\[
\begin{aligned}
\|\mathcal U_{\tau,Q^{\mathrm{init}}}(Q)-Q_\tau^\star
\|_{\tau,Q^{\mathrm{init}}}
&\le
\|\mathcal T_\tau Q-Q_\tau^\star
\|_{\tau,Q^{\mathrm{init}}}
+2\varepsilon_\tau\\
&\le
\rho_{\mathrm{init}}
\|Q-Q_\tau^\star\|_{\tau,Q^{\mathrm{init}}}
+2\varepsilon_\tau,
\end{aligned}
\]
which is \eqref{eq:stage-frozen-global-contraction}.
\end{proof}

Warm-start transfer requires control of the drift between adjacent
temperature stages.

\begin{lemma}[Lipschitz continuity of the soft-optimal path]
\label{lem:temperature-path-lipschitz}
For every \(\tau,\tau'>0\),
\begin{equation}
\label{eq:temperature-path-lipschitz}
\|Q_\tau^\star-Q_{\tau'}^\star\|_\infty
\le
\frac{\gamma L_{\mathrm{ref}}}{1-\gamma}|\tau-\tau'|.
\end{equation}
\end{lemma}

\begin{proof}
The variational representation of the reference-weighted log-partition
function gives
\[
V_{\tau,Q}(s)
=
\sup_{\pi}
\left\{
\E_\pi[Q(s,A)]
-\tau\KL\!\bigl(\pi(\cdot\mid s)\big\|\pi_{\mathrm{ref}}(\cdot\mid s)\bigr)
\right\}.
\]
Every policy satisfies
\[
0\le\KL(\pi\|\pi_{\mathrm{ref}})
=\sum_a\pi(a\mid s)
\log\frac{\pi(a\mid s)}{\pi_{\mathrm{ref}}(a\mid s)}
\le
\sum_a\pi(a\mid s)
\log\frac{1}{\pi_{\mathrm{ref}}(a\mid s)}
\le L_{\mathrm{ref}}.
\]
Hence
\(\|\mathcal T_\tau Q-\mathcal T_{\tau'}Q\|_\infty
\le\gamma L_{\mathrm{ref}}|\tau-\tau'|\).
Both Bellman operators are \(\gamma\)-contractions in supremum norm.
Using their fixed-point equations and adding and subtracting
\(\mathcal T_\tau Q_{\tau'}^\star\) gives
\[
\begin{aligned}
\|Q_\tau^\star-Q_{\tau'}^\star\|_\infty
&\le
\|\mathcal T_\tau Q_\tau^\star
-\mathcal T_\tau Q_{\tau'}^\star\|_\infty
+\|\mathcal T_\tau Q_{\tau'}^\star
-\mathcal T_{\tau'}Q_{\tau'}^\star\|_\infty\\
&\le
\gamma\|Q_\tau^\star-Q_{\tau'}^\star\|_\infty
+\gamma L_{\mathrm{ref}}|\tau-\tau'|.
\end{aligned}
\]
Moving the first term on the right to the left proves
\eqref{eq:temperature-path-lipschitz}.
\end{proof}

The next lemma shows that sufficiently small path approximation error
permits the geometric schedule in
Section~\ref{sec:population-annealing} to remain inside the local contraction
regions.

\begin{lemma}[Feasibility of the annealing conditions]
\label{prop:uniform-annealing-feasibility}
Assume Conditions~\ref{cond::convex} and~\ref{cond::supnorm}. Choose
\(\tau_0\ge\tau_{\mathrm{init}}\). There are constants
\(R_{\mathrm{ann}}>0\), \(\varepsilon_{\mathrm{ann}}>0\), and
\(\rho_{\mathrm{feas}}\in(0,1)\), depending on the fixed interval
\([\tau_{\mathrm{tar}},\tau_0]\), such that
\(\bar\rho\le\rho_{\mathrm{feas}}\) and
\(\operatorname{err}_{\mathrm{pop}}(\varepsilon_{\mathrm{path}})<R_{\mathrm{ann}}\), with
\(R_{\mathrm{loc}}=R_{\mathrm{ann}}\), whenever
\(\varepsilon_{\mathrm{path}}\le\varepsilon_{\mathrm{ann}}\). For every such
\(\varepsilon_{\mathrm{path}}\), finite shared \(K\) and \(J\) satisfy the
schedule condition in Theorem~\ref{thm:population-annealing}.
\end{lemma}

\begin{proof}
By \eqref{eq:annealing-local-modulus},
\[
\rho_{\tau}(R_{\mathrm{loc}},\varepsilon)
=\sqrt\gamma\,
\exp\{2d(R_{\mathrm{loc}},\varepsilon)/\tau\}.
\]
Because \(\tau\ge\tau_{\mathrm{tar}}>0\) on the fixed interval and
\(d(R_{\mathrm{loc}},\varepsilon)\to0\) as
\((R_{\mathrm{loc}},\varepsilon)\to(0,0)\), this expression converges
uniformly to \(\sqrt\gamma<1\). Choose
\(R_{\mathrm{ann}}>0\) small enough
that
\[
\sup_{\tau\in[\tau_{\mathrm{tar}},\tau_0]}
\rho_{\tau}(R_{\mathrm{ann}},0)<1.
\]
Continuity then gives \(\varepsilon_1>0\) and
\(\rho_{\mathrm{feas}}<1\) such that
\(\bar\rho\le\rho_{\mathrm{feas}}\)
whenever \(\varepsilon_{\mathrm{path}}\le\varepsilon_1\). Under this
bound,
\[
\begin{aligned}
\operatorname{err}_{\mathrm{pop}}(\varepsilon_{\mathrm{path}})
&\le\frac{2\varepsilon_{\mathrm{path}}}{1-\rho_{\mathrm{feas}}},\\
C_{\mathrm{int}}
\{\operatorname{err}_{\mathrm{pop}}(\varepsilon_{\mathrm{path}})
+\varepsilon_{\mathrm{path}}\}^\alpha
+\varepsilon_{\mathrm{path}}
&\le
C_{\mathrm{int}}
\left\{
\frac{2\varepsilon_{\mathrm{path}}}{1-\rho_{\mathrm{feas}}}
+\varepsilon_{\mathrm{path}}
\right\}^{\alpha}
+\varepsilon_{\mathrm{path}}.
\end{aligned}
\]
Both right-hand sides converge to zero with
\(\varepsilon_{\mathrm{path}}\). Reducing \(\varepsilon_1\) to a
positive \(\varepsilon_{\mathrm{ann}}\) therefore gives
\(\operatorname{err}_{\mathrm{pop}}(\varepsilon_{\mathrm{path}})<R_{\mathrm{ann}}\)
and
\[
C_{\mathrm{int}}
\left\{
\frac{2\varepsilon_{\mathrm{path}}}{1-\rho_{\mathrm{feas}}}
+\varepsilon_{\mathrm{path}}
\right\}^{\alpha}
+\varepsilon_{\mathrm{path}}
<R_{\mathrm{ann}}.
\]
Write
\[
b
:=
\frac{2\varepsilon_{\mathrm{path}}}{1-\rho_{\mathrm{feas}}}
+\varepsilon_{\mathrm{path}},
\qquad
s
:=
R_{\mathrm{ann}}-C_{\mathrm{int}}b^\alpha
-\varepsilon_{\mathrm{path}}>0,
\]
and let
\(R_+:=\max\{R_{\mathrm{init}},R_{\mathrm{ann}}\}\).
Since \(\rho_{\mathrm{feas}}<1\), choose \(K\) such that
\[
C_{\mathrm{int}}
\{\rho_{\mathrm{feas}}^K R_+ + b\}^{\alpha}
-C_{\mathrm{int}}b^\alpha
\le \frac{s}{2}.
\]
Then choose \(J\) such that
\[
\frac{\gamma\tau_0L_{\mathrm{ref}}}{(1-\gamma)J}
\log\left(\frac{\tau_0}{\tau_{\mathrm{tar}}}\right)
\le \frac{s}{2}.
\]
The bounds
\(\bar\rho\le\rho_{\mathrm{feas}}\) and
\(\operatorname{err}_{\mathrm{pop}}(\varepsilon_{\mathrm{path}})
\le b-\varepsilon_{\mathrm{path}}\) now show that the left-hand side of
the schedule condition in Theorem~\ref{thm:population-annealing} is at most
\(C_{\mathrm{int}}b^\alpha+\varepsilon_{\mathrm{path}}+s
=R_{\mathrm{ann}}\).
\end{proof}

\begin{proof}[Proof of Theorem~\ref{thm:population-annealing}]
Lemmas~\ref{prop:global-high-temperature} and
\ref{prop:stage-frozen-affine-contraction}, together with the definition of
\(\bar\rho\), give
\[
\|Q_{j,k+1}-Q_j^\star\|_j
\le
\bar\rho\|Q_{j,k}-Q_j^\star\|_j
+2\varepsilon_j,
\qquad 0\le j<J,
\]
globally at stage zero and inside the radius-\(R_{\mathrm{loc}}\) ball at
every later stage.
By the definition of \(\operatorname{err}_{\mathrm{pop}}\),
\((1-\bar\rho)\operatorname{err}_{\mathrm{pop}}(\varepsilon_{\mathrm{path}})
=2\varepsilon_{\mathrm{path}}\), unrolling
this recursion gives
\begin{equation}
\label{eq:stage-frozen-within-stage}
\|Q_{j,k}-Q_j^\star\|_j
\le
\bar\rho^k\|Q_{j,0}-Q_j^\star\|_j
+(1-\bar\rho^k)\operatorname{err}_{\mathrm{pop}}(\varepsilon_{\mathrm{path}}),
\qquad 0\le j<J.
\end{equation}
At stage zero, the first norm is at most \(R_{\mathrm{init}}\). At a later
stage whose initializer has error at most \(R_{\mathrm{loc}}\), the bound
remains below \(R_{\mathrm{loc}}\) because
\(\operatorname{err}_{\mathrm{pop}}(\varepsilon_{\mathrm{path}})<R_{\mathrm{loc}}\).

It remains to verify the warm starts. Let
\(Q_j^\circ\) be a best supremum-norm approximant to
\(Q_j^\star\). Equation~\eqref{eq:stage-frozen-within-stage} implies
\[
\|Q_{j,K}-Q_j^\circ\|_j
\le
x_j+(1-\bar\rho^K)
\operatorname{err}_{\mathrm{pop}}(\varepsilon_{\mathrm{path}})
+\varepsilon_{\mathrm{path}},
\]
where \(x_0=\bar\rho^KR_{\mathrm{init}}\) and
\(x_j=\bar\rho^KR_{\mathrm{loc}}\) for \(j\ge1\).
The difference
\(Q_{j,K}-Q_j^\circ\) belongs to
\(\mathcal F-\mathcal F\). Since
\[
Q_{j+1,0}-Q_{j+1}^\star
=
(Q_{j,K}-Q_j^\circ)
+(Q_j^\circ-Q_j^\star)
+(Q_j^\star-Q_{j+1}^\star),
\]
and \(\|H\|_{j+1}\le\|H\|_\infty\) for every bounded \(H\),
\eqref{eq:stage-frozen-interpolation} and
Lemma~\ref{lem:temperature-path-lipschitz} imply
\[
\|Q_{j+1,0}-Q_{j+1}^\star\|_{j+1}
\le
C_{\mathrm{int}}
\left\{
x_j+(1-\bar\rho^K)
\operatorname{err}_{\mathrm{pop}}(\varepsilon_{\mathrm{path}})
+\varepsilon_{\mathrm{path}}
\right\}^\alpha
+\varepsilon_{\mathrm{path}}
+\frac{\gamma L_{\mathrm{ref}}}{1-\gamma}
(\tau_j-\tau_{j+1}).
\]
For \(c=(\tau_{\mathrm{tar}}/\tau_0)^{1/J}\),
the geometric schedule gives
\[
\tau_j-\tau_{j+1}
=\tau_j(1-c)
\le\frac{\tau_0}{J}
\log\left(\frac{\tau_0}{\tau_{\mathrm{tar}}}\right),
\]
where the last inequality uses \(1-e^{-x}\le x\) for \(x\ge0\).
Since
\(x_j\le\bar\rho^K\max\{R_{\mathrm{init}},R_{\mathrm{loc}}\}\) and
\((1-\bar\rho^K)\operatorname{err}_{\mathrm{pop}}
(\varepsilon_{\mathrm{path}})
\le\operatorname{err}_{\mathrm{pop}}(\varepsilon_{\mathrm{path}})\),
the schedule condition in Theorem~\ref{thm:population-annealing} makes the
preceding upper bound at most \(R_{\mathrm{loc}}\). This completes the
warm-start induction. At the target,
Lemma~\ref{prop:stage-frozen-affine-contraction} and
the definition of \(\operatorname{err}_{\mathrm{pop}}\), namely
\(2\varepsilon_J=(1-\bar\rho)\operatorname{err}_{\mathrm{pop}}(\varepsilon_J)\) imply
\[
\|Q_{J,k+1}-Q_J^\star\|_J
\le
\bar\rho\|Q_{J,k}-Q_J^\star\|_J
+(1-\bar\rho)\operatorname{err}_{\mathrm{pop}}(\varepsilon_J).
\]
Unrolling the recursion and using
\((1-\bar\rho^k)\operatorname{err}_{\mathrm{pop}}(\varepsilon_J)
\le\operatorname{err}_{\mathrm{pop}}(\varepsilon_J)\) proves the target-stage
bound in Theorem~\ref{thm:population-annealing}. If \(\varepsilon_J=0\), the additive
term in Lemma~\ref{prop:stage-frozen-affine-contraction} vanishes, and
the sharper exact-convergence bound follows.
\end{proof}

\section{Hard FQI with Refreshed Occupancy Weights}
\label{app:hard-polishing}

This section establishes the local population guarantee for hard FQI after
the positive-temperature initialization proved in
Appendix~\ref{app:stage-frozen-population-annealing}. In the hard phase, the
exact occupancy ratio is recomputed before each Bellman regression.

Throughout the proof, write \(Q^\star:=Q_{\mathrm{hard}}^\star\),
\(\pi^\star:=\pi_{0,Q^\star}\), and
\[
P_Q:=P_{\pi_{0,Q}},
\qquad
\mu_Q
:=(1-\gamma)\nu_{\mathrm{ref}}+\gamma\mu_QP_Q,
\qquad
\xi_Q:=\text{the state marginal of }\mu_Q.
\]
Define
\[
P_\star:=P_{\pi^\star},\qquad
\mu_\star:=\mu_{\mathrm{hard}},\qquad
\xi_\star
:=\xi_{\mathrm{hard}}.
\]
For a bounded state function \(g\), write
\[
(Pg)(s,a):=\E\{g(S')\mid S=s,A=a\}.
\]
For the proof, define the current-occupancy hard update by
\[
\mathcal U_0(Q)
:=
\Pi_{\mathcal F}^{\mu_Q}\mathcal T_0Q.
\]

\subsection{Reference-restart minorization and occupancy perturbation}

Lemma~\ref{lem:reference-restart-interpolation}, applied to the greedy policy
\(\pi_{0,Q}\), gives
\begin{equation}
\label{eq:hard-exploratory-interpolation}
\|H\|_\infty
\le
C_{\mathrm{int}}\|H\|_{2,\mu_Q}^{\alpha},
\qquad H\in\mathcal F-\mathcal F,
\end{equation}
uniformly in \(Q\).

For \(Q\in\mathcal F\), let \(a_Q(s)\) be the maximizer of \(Q(s,\cdot)\)
selected by the fixed tie-breaking rule, and define
\[
D_Q:=\{s:a_Q(s)\ne a_{\mathrm{hard}}^\star(s)\},
\qquad
\delta_Q:=\|Q-Q^\star\|_\infty.
\]
For probability measures \(\mu\) and \(\nu\), write
\[
\|\mu-\nu\|_{\mathrm{TV}}
:=
\sup_B|\mu(B)-\nu(B)|,
\]
equivalently one half of their \(L^1\) distance under any common dominating
measure.

\begin{lemma}[Greedy mismatch and occupancy perturbation]
\label{lem:hard-occupancy-perturbation}
Under Condition~\ref{cond::hard-soft-margin}, there is
\(C_{\mathrm{occ}}<\infty\)
such that, for every \(Q\in\mathcal F\) with \(\delta_Q\le1/2\),
\begin{align}
\label{eq:hard-greedy-mismatch-set}
D_Q
&\subseteq\{G\le2\delta_Q\},\\
\label{eq:hard-current-mismatch-probability}
\xi_Q(D_Q)+\xi_\star(D_Q)
&\le C_{\mathrm{occ}}\delta_Q^\beta,\\
\label{eq:hard-occupancy-tv}
\|\mu_Q-\mu_\star\|_{\mathrm{TV}}
&\le C_{\mathrm{occ}}\delta_Q^\beta.
\end{align}
\end{lemma}

\begin{proof}
If \(s\in D_Q\), optimality of \(a_Q(s)\) for \(Q\) gives
\[
Q^\star(s,a_{\mathrm{hard}}^\star(s))
-Q^\star(s,a_Q(s))
\le
\{Q-Q^\star\}(s,a_Q(s))
-\{Q-Q^\star\}(s,a_{\mathrm{hard}}^\star(s))
\le2\delta_Q.
\]
The left side is at least \(G(s)\), proving
\eqref{eq:hard-greedy-mismatch-set}. Condition~\ref{cond::hard-soft-margin}
and \(\delta_Q\le1/2\) give
\[
\xi_\star(D_Q)
\le C_G(2\delta_Q)^\beta.
\]

The restart law is the same for \(Q\) and \(Q^\star\). Subtracting their
balance equations gives the state-action resolvent identity
\[
\mu_Q-\mu_\star
=
\gamma\mu_\star(P_Q-P_\star)
(I-\gamma P_Q)^{-1}.
\]
The kernels \(P_Q\) and \(P_\star\) differ only when the successor state
belongs to \(D_Q\). Thus, with
\(P(D)(s,a):=P\{S'\in D\mid S=s,A=a\}\),
\[
\|\mu_\star(P_Q-P_\star)\|_{\mathrm{TV}}
\le \mu_\star P(D_Q).
\]
Total-variation contraction of Markov kernels and
\((I-\gamma P_Q)^{-1}=\sum_{t\ge0}\gamma^tP_Q^t\) then give
\[
\|\mu_Q-\mu_\star\|_{\mathrm{TV}}
\le
\frac{\gamma}{1-\gamma}\mu_\star P(D_Q).
\]
Taking state marginals in the balance equation for
\(\mu_\star\) yields
\(\xi_\star=(1-\gamma)\nu_0
+\gamma\mu_\star P\). Hence
\[
\|\mu_Q-\mu_\star\|_{\mathrm{TV}}
\le
\frac{\xi_\star(D_Q)}{1-\gamma}
\le
\frac{C_G2^\beta}{1-\gamma}\delta_Q^\beta.
\]
This proves \eqref{eq:hard-occupancy-tv}. Marginalization cannot increase
total variation, so
\[
\xi_Q(D_Q)
\le
\xi_\star(D_Q)
+\|\mu_Q-\mu_\star\|_{\mathrm{TV}}
\le C\delta_Q^\beta,
\]
which proves \eqref{eq:hard-current-mismatch-probability}.
\end{proof}

The next lemma converts the total-variation bound into a comparison of the
two projection norms. This is the step at which the interpolation exponent
and the margin exponent interact.

\begin{lemma}[Hard reference-restart norm comparison]
\label{lem:hard-exploratory-norm-comparison}
Let
\[
e:=\|Q-Q^\star\|_{2,\mu_\star},
\qquad
p_{\mathrm{norm}}:=\frac{\alpha\beta}{2(1-\alpha)}.
\]
There are \(c_{\mathrm{norm}},C_{\mathrm{norm}}>0\) such that, whenever
\(e\le c_{\mathrm{norm}}\), every \(H\in\mathcal F-\mathcal F\) satisfies
\begin{align}
\label{eq:hard-norm-transfer-forward}
\|H\|_{2,\mu_Q}
&\le
\sqrt2\|H\|_{2,\mu_\star}+C_{\mathrm{norm}}e^{p_{\mathrm{norm}}},\\
\label{eq:hard-norm-transfer-reverse}
\|H\|_{2,\mu_\star}
&\le
\sqrt2\|H\|_{2,\mu_Q}+C_{\mathrm{norm}}e^{p_{\mathrm{norm}}}.
\end{align}
\end{lemma}

\begin{proof}
Lemma~\ref{lem:reference-restart-interpolation} gives
\(\delta_Q\le C_{\mathrm{int}}e^\alpha\). Choose
\(c_{\mathrm{norm}}\) small enough that
\(C_{\mathrm{int}}c_{\mathrm{norm}}^\alpha\le1/2\).
Lemma~\ref{lem:hard-occupancy-perturbation} then
implies
\[
\|\mu_Q-\mu_\star\|_{\mathrm{TV}}
\le C e^{\alpha\beta}.
\]
For
\(x:=\|H\|_{2,\mu_\star}\) and
\(y:=\|H\|_{2,\mu_Q}\), total variation and
\eqref{eq:hard-exploratory-interpolation}, first under
\(\mu_\star\) and then under \(\mu_Q\), give
\[
y^2\le x^2+Ce^{\alpha\beta}x^{2\alpha},
\qquad
x^2\le y^2+Ce^{\alpha\beta}y^{2\alpha}.
\]
Set \(b:=(2C)^{1/\{2(1-\alpha)\}}e^{p_{\mathrm{norm}}}\). If \(x\ge b\), the
additional term in the first inequality is at most \(x^2/2\), so
\(y\le\sqrt{3/2}\,x\). If \(x<b\), that term is at most \(b^2/2\), so
\(y\le\sqrt{x^2+b^2/2}\le\sqrt{3/2}\,b\). Increasing the coefficient of
\(e^{p_{\mathrm{norm}}}\) gives \eqref{eq:hard-norm-transfer-forward}.
Applying these two cases to the second inequality, with \(x\) and \(y\) interchanged,
gives \eqref{eq:hard-norm-transfer-reverse}.
\end{proof}

\subsection{Local contraction of the hard update}

Define the oracle hard update
\[
\mathcal U_0^{\mathrm{or}}(Q)
:=
\Pi_{\mathcal F}^{\mu_\star}\mathcal T_0Q.
\]

\begin{lemma}[Oracle hard contraction with a margin remainder]
\label{lem:hard-oracle-contraction}
Let
\[
p_{\mathrm{rem}}
:=\alpha\left(1+\frac{\beta}{2}\right).
\]
There are \(c_{\mathrm{or}},C_{\mathrm{or}}>0\) such that, whenever
\(e=\|Q-Q^\star\|_{2,\mu_\star}\le c_{\mathrm{or}}\),
\begin{equation}
\label{eq:hard-oracle-contraction}
\|\mathcal U_0^{\mathrm{or}}(Q)-Q^\star\|_{2,\mu_\star}
\le
\sqrt\gamma\,e+C_{\mathrm{or}}e^{p_{\mathrm{rem}}}.
\end{equation}
\end{lemma}

\begin{proof}
Set \(H:=Q-Q^\star\) and \(\delta:=\|H\|_\infty\). For each state, write
\[
\max_a Q(s,a)-\max_a Q^\star(s,a)
=H(s,a_{\mathrm{hard}}^\star(s))+\operatorname{rem}_Q(s).
\]
The maximizing property of \(a_{\mathrm{hard}}^\star(s)\) for \(Q^\star\)
gives
\[
0\le \operatorname{rem}_Q(s)\le2\delta\,\mathbf1\{G(s)\le2\delta\}.
\]
Equation~\eqref{eq:hard-exploratory-interpolation} gives
\(\delta\le C_{\mathrm{int}}e^\alpha\). Reduce \(c_{\mathrm{or}}\), if
necessary, so that \(2C_{\mathrm{int}}c_{\mathrm{or}}^\alpha\le1\); then
Condition~\ref{cond::hard-soft-margin} applies at \(u=2\delta\).
The discounted balance equation for \(\mu_\star\), conditional Jensen's
inequality, and Condition~\ref{cond::hard-soft-margin} give
\begin{align}
\label{eq:hard-bellman-remainder}
\|\gamma P \operatorname{rem}_Q\|_{2,\mu_\star}^2
&\le
\gamma^2\E_{\mu_\star}\!\left[P\{\operatorname{rem}_Q^2\}\right]
\le
\gamma\E_{\xi_\star}[\operatorname{rem}_Q^2]\\
&\le
4\gamma\delta^2
\xi_\star\{G\le2\delta\}
\le C\delta^{2+\beta}.
\end{align}
Consequently, the exact Bellman decomposition satisfies
\begin{equation}
\label{eq:hard-bellman-decomposition}
\mathcal T_0Q-\mathcal T_0Q^\star
=\gamma P_{\pi^\star}H+\gamma P \operatorname{rem}_Q.
\end{equation}

Conditional Jensen's inequality and the matched balance equation for
\(\mu_\star\) give
\[
\|\gamma P_{\pi^\star}H\|_{2,\mu_\star}^2
\le
\gamma\|H\|_{2,\mu_\star}^2.
\]
Condition~\ref{cond::hard-realizability} implies
\(\Pi_{\mathcal F}^{\mu_\star}\mathcal T_0Q^\star=Q^\star\).
Metric projection is nonexpansive in \(L^2(\mu_\star)\).
Using the exact decomposition
\eqref{eq:hard-bellman-decomposition}, the bound
\(\|\gamma P_{\pi^\star}H\|_{2,\mu_\star}\le\sqrt\gamma\,e\),
and the remainder bound
\eqref{eq:hard-bellman-remainder} gives
\[
\begin{aligned}
\|\mathcal U_0^{\mathrm{or}}(Q)-Q^\star\|_{2,\mu_\star}
&\le
\|\mathcal T_0Q-\mathcal T_0Q^\star\|_{2,\mu_\star}\\
&\le
\sqrt\gamma\,e+\|\gamma P \operatorname{rem}_Q\|_{2,\mu_\star}\\
&\le
\sqrt\gamma\,e+C\delta^{1+\beta/2}.
\end{aligned}
\]
Finally, the bound \(\delta\le C_{\mathrm{int}}e^\alpha\) converts the last term to
\(Ce^{\alpha(1+\beta/2)}=Ce^{p_{\mathrm{rem}}}\), proving
\eqref{eq:hard-oracle-contraction}.
\end{proof}

\begin{lemma}[Current reference-restart projection stability at zero temperature]
\label{prop:hard-current-projection-stability}
Let
\[
p_{\mathrm{proj}}:=\frac{\alpha(1+\beta)}{2-\alpha}.
\]
There are \(c_{\mathrm{pr}},C_{\mathrm{pr}}>0\) such that, whenever
\(e=\|Q-Q^\star\|_{2,\mu_\star}\le c_{\mathrm{pr}}\),
\begin{equation}
\label{eq:hard-current-projection-stability}
\|\mathcal U_0(Q)-\mathcal U_0^{\mathrm{or}}(Q)\|
_{2,\mu_\star}
\le
C_{\mathrm{pr}}\{e^{p_{\mathrm{proj}}}+e^{p_{\mathrm{norm}}}\}.
\end{equation}
\end{lemma}

\begin{proof}
Abbreviate
\[
Y:=\mathcal T_0Q,
\qquad
f_\star:=\Pi_{\mathcal F}^{\mu_\star}Y,
\qquad
f_Q:=\Pi_{\mathcal F}^{\mu_Q}Y,
\qquad
\Delta:=f_Q-f_\star.
\]
Using \(f_Q\) as a comparator for the
\(\mu_\star\)-projection and \(f_\star\) as a comparator for the
\(\mu_Q\)-projection gives
\[
\langle Y-f_\star,\Delta\rangle_{\mu_\star}\le0,
\qquad
\langle Y-f_Q,-\Delta\rangle_{\mu_Q}\le0.
\]
Since \(Y-f_Q=Y-f_\star-\Delta\), the second inequality is
\(\|\Delta\|_{2,\mu_Q}^2
\le\langle Y-f_\star,\Delta\rangle_{\mu_Q}\).
Subtracting the nonpositive first inner product therefore gives
\begin{equation}
\label{eq:hard-projection-vi}
\|\Delta\|_{2,\mu_Q}^2
\le
(\mu_Q-\mu_\star)\{(Y-f_\star)\Delta\}.
\end{equation}
The hard Bellman operator is a \(\gamma\)-contraction in supremum norm.
Lemma~\ref{lem:hard-oracle-contraction} and
\eqref{eq:hard-exploratory-interpolation} therefore give, for sufficiently
small \(e\),
\begin{equation}
\label{eq:hard-projection-residual-localization}
\begin{aligned}
\|Y-Q^\star\|_\infty
&\le\gamma\|Q-Q^\star\|_\infty
\le\gamma C_{\mathrm{int}}e^\alpha,\\
\|f_\star-Q^\star\|_\infty
&\le
C_{\mathrm{int}}
\|f_\star-Q^\star\|_{2,\mu_\star}^{\alpha}\\
&\le
C_{\mathrm{int}}
\{\sqrt\gamma\,e+C_{\mathrm{or}}e^{p_{\mathrm{rem}}}\}^{\alpha}
\le Ce^\alpha.
\end{aligned}
\end{equation}
Thus \(\|Y-f_\star\|_\infty\le Ce^\alpha\) by the triangle inequality.
Lemma~\ref{lem:hard-occupancy-perturbation} gives
\(\|\mu_Q-\mu_\star\|_{\mathrm{TV}}\le Ce^{\alpha\beta}\).
Set \(y:=\|\Delta\|_{2,\mu_Q}\). Applying
\eqref{eq:hard-projection-residual-localization},
\eqref{eq:hard-occupancy-tv}, and
\eqref{eq:hard-exploratory-interpolation} to
\eqref{eq:hard-projection-vi} yields
\[
y^2
\le
Ce^{\alpha(1+\beta)}y^\alpha.
\]
Thus \(y\le Ce^{p_{\mathrm{proj}}}\). Finally,
Lemma~\ref{lem:hard-exploratory-norm-comparison} applied to
\(\Delta\in\mathcal F-\mathcal F\) gives
\[
\|\Delta\|_{2,\mu_\star}
\le\sqrt2y+C_{\mathrm{norm}}e^{p_{\mathrm{norm}}},
\]
which proves \eqref{eq:hard-current-projection-stability}.
\end{proof}

\begin{lemma}[Local contraction of hard FQI with refreshed occupancy weights]
\label{prop:hard-local-contraction}
Assume Conditions~\ref{cond::convex}, \ref{cond::supnorm}, and
\ref{cond::hard-realizability}--\ref{cond::hard-soft-margin}. There exists
\(R_{\mathrm{hard}}\in(0,1]\) such that, whenever
\(\|Q_0-Q^\star\|_{2,\mu_\star}\le R_{\mathrm{hard}}\), the iterates
\(Q_{k+1}=\mathcal U_0(Q_k)\) remain in this neighborhood and satisfy
\begin{equation}
\label{eq:hard-local-contraction-rate}
\|Q_k-Q^\star\|_{2,\mu_\star}
\le
\rho_{\mathrm{hard}}^k
\|Q_0-Q^\star\|_{2,\mu_\star},
\qquad
\rho_{\mathrm{hard}}:=\frac{1+\sqrt\gamma}{2}.
\end{equation}
Moreover,
\[
\|Q_k-Q^\star\|_\infty
\le
C_{\mathrm{int}}\rho_{\mathrm{hard}}^{\alpha k}
\|Q_0-Q^\star\|_{2,\mu_\star}^{\alpha}.
\]
\end{lemma}

\begin{proof}
The compatibility condition
\(\alpha(1+\beta/2)>1\) is equivalent to
\(\alpha\beta>2(1-\alpha)\). It implies
\[
p_{\mathrm{rem}}>1,\qquad
p_{\mathrm{proj}}-p_{\mathrm{rem}}
=\frac{\alpha(p_{\mathrm{rem}}-1)}{2-\alpha}>0,\qquad
p_{\mathrm{norm}}-p_{\mathrm{rem}}
=\alpha(p_{\mathrm{norm}}-1)>0.
\]
Lemma~\ref{lem:hard-oracle-contraction} and
Lemma~\ref{prop:hard-current-projection-stability}, combined by the
triangle inequality, give
\begin{equation}
\label{eq:hard-full-local-recursion}
\|\mathcal U_0(Q)-Q^\star\|_{2,\mu_\star}
\le
\sqrt\gamma\,e+Ce^{p_{\mathrm{rem}}},
\qquad 0<e\le1.
\end{equation}
Indeed, the triangle inequality first bounds the left side by
\[
\|\mathcal U_0^{\mathrm{or}}(Q)-Q^\star\|_{2,\mu_\star}
+
\|\mathcal U_0(Q)-\mathcal U_0^{\mathrm{or}}(Q)
\|_{2,\mu_\star},
\]
and \(p_{\mathrm{proj}}\ge p_{\mathrm{rem}}\) and
\(p_{\mathrm{norm}}\ge p_{\mathrm{rem}}\) imply
\(e^{p_{\mathrm{proj}}}+e^{p_{\mathrm{norm}}}
\le2e^{p_{\mathrm{rem}}}\) for \(0<e\le1\).
Choose
\(R_{\mathrm{hard}}\in(0,1]\) small enough that the higher-order term in
\eqref{eq:hard-full-local-recursion} is at most
\((\rho_{\mathrm{hard}}-\sqrt\gamma)e\) whenever
\(0<e\le R_{\mathrm{hard}}\), and so that
\(2C_{\mathrm{int}}R_{\mathrm{hard}}^\alpha\le1\). The resulting ball is
invariant, and iteration proves \eqref{eq:hard-local-contraction-rate}.

The reference-restart minorization at \(Q^\star\) and
Condition~\ref{cond::supnorm} give
\[
\|Q_k-Q^\star\|_\infty
\le
C_{\mathrm{int}}
\|Q_k-Q^\star\|_{2,\mu_\star}^{\alpha},
\]
and substituting \eqref{eq:hard-local-contraction-rate} proves
\[
\|Q_k-Q^\star\|_\infty
\le
C_{\mathrm{int}}\rho_{\mathrm{hard}}^{\alpha k}
\|Q_0-Q^\star\|_{2,\mu_\star}^{\alpha}.
\]
\end{proof}

\subsection{Initialization at a fixed positive temperature}

Define
\[
\ell_{\mathrm{ref}}(s)
:=
\log\frac{1}{\pi_{\mathrm{ref}}
(a_{\mathrm{hard}}^\star(s)\mid s)},
\]
and let \(H_{\mathrm{ref}}\) be the unique bounded solution of
\begin{equation}
\label{eq:reference-drift-equation}
H_{\mathrm{ref}}(s,a)
=
\gamma\E\!\left[
\ell_{\mathrm{ref}}(S')
+H_{\mathrm{ref}}(S',a_{\mathrm{hard}}^\star(S'))
\,\middle|\,s,a
\right].
\end{equation}

\begin{lemma}[General-reference soft-to-hard comparison]
\label{lem:general-reference-soft-hard-comparison}
Let \(H_{\mathrm{ref}}\) be defined by
\eqref{eq:reference-drift-equation}. For every \(\tau>0\),
\begin{equation}
\label{eq:general-reference-soft-hard-sandwich}
Q^\star-\tau H_{\mathrm{ref}}
\le Q_\tau^\star\le Q^\star,
\qquad
\|Q_\tau^\star-Q^\star\|_\infty
\le\tau\|H_{\mathrm{ref}}\|_\infty
\le
\frac{\gamma L_{\mathrm{ref}}}{1-\gamma}\tau.
\end{equation}
\end{lemma}

\begin{proof}
Abbreviate \(a^\star(s):=a_{\mathrm{hard}}^\star(s)\),
\(\ell(s):=\ell_{\mathrm{ref}}(s)\), and \(H:=H_{\mathrm{ref}}\).
The bounded solution of \eqref{eq:reference-drift-equation} has the
representation
\[
H(s,a)
=
\gamma\E_{s,a}\!\left[
\sum_{t=0}^\infty\gamma^t\ell(S_{t+1})
\right],
\]
where the process follows \(\pi^\star\) after its first transition.
Consequently, \(0\le H\le\gamma L_{\mathrm{ref}}/(1-\gamma)\).

Set \(\widetilde Q_\tau:=Q^\star-\tau H\), and, for
\(a\in\mathcal A\), write
\[
g_a(s):=Q^\star(s,a^\star(s))-Q^\star(s,a),
\qquad
H^\star(s):=H(s,a^\star(s)).
\]
A direct evaluation of the reference-weighted log-partition function gives
\begin{align}
\label{eq:general-reference-comparison-value}
V_{\tau,\widetilde Q_\tau}(s)
&=
\max_aQ^\star(s,a)
-\tau\{H^\star(s)+\ell(s)\}
+b_\tau(s),\\
b_\tau(s)
&:=
\tau\log\!\left[
1+
\sum_{a\ne a^\star(s)}
\frac{\pi_{\mathrm{ref}}(a\mid s)}
{\pi_{\mathrm{ref}}(a^\star(s)\mid s)}
\exp\!\left\{
-\frac{g_a(s)}{\tau}
-H(s,a)+H^\star(s)
\right\}
\right]\ge0.
\notag
\end{align}
Combining \eqref{eq:general-reference-comparison-value} with the Bellman
equation for \(Q^\star\) and
\eqref{eq:reference-drift-equation} yields the exact residual identity
\[
\mathcal T_\tau\widetilde Q_\tau-\widetilde Q_\tau
=\gamma P b_\tau\ge0.
\]
Thus \(\widetilde Q_\tau\) is a subsolution. Moreover,
\(\mathcal T_\tau Q^\star\le\mathcal T_0Q^\star=Q^\star\), so
\(Q^\star\) is a supersolution. Monotonicity and supremum-norm contraction
of \(\mathcal T_\tau\) imply, for every \(n\ge1\),
\[
\widetilde Q_\tau
\le \mathcal T_\tau^{\circ n}\widetilde Q_\tau,
\qquad
\mathcal T_\tau^{\circ n}Q^\star\le Q^\star.
\]
Both sequences converge in supremum norm to the unique fixed point
\(Q_\tau^\star\). Taking limits proves the pointwise sandwich in
\eqref{eq:general-reference-soft-hard-sandwich}. Taking the
supremum norm and using
\(\|H\|_\infty\le\gamma L_{\mathrm{ref}}/(1-\gamma)\) proves the remaining
bounds.
\end{proof}

\begin{proof}[Proof of Theorem~\ref{thm:population-hard-polishing}]
Run Theorem~\ref{thm:population-annealing} with the fixed target
\(\tau_{\mathrm{sw}}\).
At the final positive-temperature stage, the frozen
occupancy satisfies the reference minorization used in
\eqref{eq:stage-frozen-interpolation}. Let
\(Q_{\mathrm{sw}}^\circ\in\mathcal F\) be a best
supremum-norm approximant to \(Q_{\tau_{\mathrm{sw}}}^\star\). The
target-stage bound in Theorem~\ref{thm:population-annealing} gives
\[
\|Q_{J,K}-Q_{\mathrm{sw}}^\circ\|_
{2,\mu_J}
\le
\bar\rho^KR_{\mathrm{loc}}
+(1-\bar\rho^K)\operatorname{err}_{\mathrm{pop}}(\varepsilon_J)
+\varepsilon_J.
\]
Applying \eqref{eq:stage-frozen-interpolation} to
\(Q_{J,K}-Q_{\mathrm{sw}}^\circ\in\mathcal F-\mathcal F\), and then using
\(\|Q_{\mathrm{sw}}^\circ-Q_{\tau_{\mathrm{sw}}}^\star\|_\infty
=\varepsilon_J\), yields
\[
\|Q_{J,K}-Q_{\tau_{\mathrm{sw}}}^\star\|_\infty
\le
C_{\mathrm{int}}
\left\{
\bar\rho^KR_{\mathrm{loc}}
+(1-\bar\rho^K)\operatorname{err}_{\mathrm{pop}}(\varepsilon_J)
+\varepsilon_J
\right\}^{\alpha}
+\varepsilon_J.
\]
Because \((1-\bar\rho^K)\operatorname{err}_{\mathrm{pop}}(\varepsilon_J)
\le\operatorname{err}_{\mathrm{pop}}(\varepsilon_J)\), the right-hand side
is bounded by the soft-iteration term in
\eqref{eq:hard-switch-temperature}.
Set \(Q_0^{\mathrm h}:=Q_{J,K}\). Lemma~
\ref{lem:general-reference-soft-hard-comparison} and
\eqref{eq:hard-switch-temperature} then give
\[
\|Q_0^{\mathrm h}-Q^\star\|_\infty
<R_{\mathrm{hard}}.
\]
Therefore
\(\|Q_0^{\mathrm h}-Q^\star\|_{2,\mu_\star}
<R_{\mathrm{hard}}\). Lemma~\ref{prop:hard-local-contraction}, with
\(Q_0=Q_0^{\mathrm h}\), gives \eqref{eq:hard-polishing-l2-rate}.

There is one hard Bellman regression and one refreshed occupancy-ratio fit
per hard iteration. Moreover,
\[
1-\sqrt\gamma
=
\frac{1-\gamma}{1+\sqrt\gamma},
\qquad
1-\rho_{\mathrm{hard}}=\frac{1-\sqrt\gamma}{2}.
\]
Thus the hard contraction gap is of order \(1-\gamma\).
Equation~\eqref{eq:hard-polishing-l2-rate} and
\eqref{eq:hard-exploratory-interpolation} give
\[
\|Q_k^{\mathrm h}-Q^\star\|_\infty
\le
C_{\mathrm{int}}\rho_{\mathrm{hard}}^{\alpha k}
\|Q_0^{\mathrm h}-Q^\star\|_
{2,\mu_\star}^{\alpha}.
\]
This upper bound is at most \(\varepsilon\) whenever
\[
k\ge
\frac{
\log\{C_{\mathrm{int}}
\|Q_0^{\mathrm h}-Q^\star\|_
{2,\mu_\star}^\alpha/\varepsilon\}
}{
\alpha\log(1/\rho_{\mathrm{hard}})
}.
\]
Because \(\log(1/\rho_{\mathrm{hard}})\ge1-\rho_{\mathrm{hard}}\), this is
\[
O\!\left(
\frac{\log(1/\varepsilon)+\log\{1/(1-\gamma)\}}{1-\gamma}
\right)
\]
when \(C_0\) and the initial \(L^2(\mu_\star)\) error are uniform in
\(\gamma\). The positive-temperature annealing phase and its switch
temperature are fixed independently of the target terminal accuracy.
\end{proof}

\section{Discounted-Occupancy Ratio Estimation}
\label{app:ratio-estimation}

This appendix gives fixed-policy rates for estimating discounted-occupancy
ratios and verifies the simultaneous ratio-error control used by both
finite-sample phases.

\subsection{Fixed-Policy Identification and Estimation}

\paragraph{Identification and estimator choices.}
For a fixed policy \(\pi\) and state-action restart distribution \(\zeta\),
let \(\mu_{\zeta,\pi}\) be the solution of
\[
\mu_{\zeta,\pi}
=(1-\gamma)\zeta+\gamma\mu_{\zeta,\pi}P_\pi.
\]
The usual discounted occupancy takes
\(\zeta=\nu_0\otimes\pi\); both phases of our method take
\(\zeta=\nu_{\mathrm{ref}}\). When
\(\mu_{\zeta,\pi}\ll\nu_b\), the
discounted balance equation identifies
\(w_{\zeta,\pi}=\mathrm d\mu_{\zeta,\pi}/\mathrm d\nu_b\) through the
following identity,
valid for every bounded measurable \(g\):
\begin{equation}
\label{eqn:app-discounted-moment}
\E_{\nu_b}\!\left[w_{\zeta,\pi}\{g-\gamma P_\pi g\}\right]
=
(1-\gamma)\E_{\zeta}g,
\end{equation}
where \((P_\pi g)(s,a)=\E\{g(S',A')\mid S=s,A=a\}\), with
\(S'\sim P(\cdot\mid s,a)\) and \(A'\sim\pi(\cdot\mid S')\). Taking
\(g\equiv1\) gives
\(\E_{\nu_b}w_{\zeta,\pi}=1\), and the discounted recursion uniquely
characterizes \(w_{\zeta,\pi}\) up to \(\nu_b\)-null sets. For the usual
occupancy, the initial moment on the right can be estimated from offline
initial states by integrating over or sampling
\(A\sim\pi(\cdot\mid S)\); for the reference-restart phases it is an expectation under
\(\nu_{\mathrm{ref}}\), which is directly sampled by pairing offline states
with actions drawn from the known reference policy.

DICE and minimax estimators enforce \eqref{eqn:app-discounted-moment} over a
critic class while fitting the ratio
\citep{nachum2019dualdice,zhang2020gendice,uehara2020minimax,
uehara2021finite,wang2023projected}. Their bounds combine ratio approximation,
critic identification, and statistical error. FORE instead applies a fitted
fixed-point recursion to a density-ratio loss \citep{van2026fitted}. It does
not require completeness of an auxiliary critic class, but imposes conditions
on coverage and the ratio class.

\begin{lemma}[Density lower bound for a reference-restart occupancy]
\label{lem:positive-occupancy-density}
Fix a policy \(\pi\), and let
\(\mu_\pi
=(1-\gamma)\nu_{\mathrm{ref}}
+\gamma\mu_\pi P_\pi\). For every measurable set \(E\),
\[
\mu_\pi(E)
\ge
(1-\gamma)e^{-L_{\mathrm{ref}}}\nu_b(E).
\]
Consequently, if \(\mu_\pi\ll\nu_b\), then
\[
\frac{\mathrm d\mu_\pi}{\mathrm d\nu_b}
\ge
(1-\gamma)e^{-L_{\mathrm{ref}}}
>0
\quad \nu_b\text{-almost surely}.
\]
\end{lemma}

\begin{lemma}[Restricted curvature from ratio accuracy]
\label{prop:ratio-curvature-linear}
Let \(\mu\ll\nu_b\), write \(w:=\mathrm d\mu/\mathrm d\nu_b\), and suppose
\(\mathcal F\subseteq\{(s,a)\mapsto\phi(s,a)^\top\theta:
\theta\in\mathbb R^p\}\), \(\|\phi(s,a)\|_2\le B_\phi\), and
\[
\Sigma_\mu:=\E_\mu[\phi\phi^\top]\succeq\lambda_\mu I_p
\qquad\text{for some \(\lambda_\mu>0\)}.
\]
Let \(\widehat w\ge0\), and set
\(\varepsilon_w(\mu)
:=\|\widehat w/w-1\|_{2,\mu}\), where the quotient is evaluated on the
support of \(\mu\). Then
\[
\E_{\nu_b}[\widehat wH^2]
\ge
\max\left\{0,1-
\frac{B_\phi}{\sqrt{\lambda_\mu}}
\varepsilon_w(\mu)
\right\}
\|H\|_{2,\mu}^2
\qquad\text{for every }H\in\mathcal F-\mathcal F.
\]
In particular, if
\(\varepsilon_w(\mu)\le\sqrt{\lambda_\mu}/(2B_\phi)\), then
Condition~\ref{cond:finite-frozen-curvature} holds with
\(\underline\lambda_{\mathcal F}=1/2\), provided the estimated weight is
also bounded.
\end{lemma}

Suppose in addition that
\(\Sigma_{\mathrm{ref}}:=\E_{\nu_{\mathrm{ref}}}[\phi\phi^\top]
\succeq\lambda_0I_p\). The reference-restart minorization gives
\[
\Sigma_\mu\succeq
(1-\gamma)\Sigma_{\mathrm{ref}}
\succeq(1-\gamma)\lambda_0I_p.
\]
Thus, for every positive-temperature or hard reference-restart occupancy,
\[
\varepsilon_w(\mu)
\le
\frac{\sqrt{(1-\gamma)\lambda_0}}{2B_\phi}
\]
verifies the restricted-curvature part of
Condition~\ref{cond:finite-frozen-curvature}. Clipping verifies its
bounded-weight requirement. No pointwise lower bound on the
estimated-to-true ratio is needed.

\paragraph{A fixed-policy FORE rate.}
Fix a restart distribution \(\zeta\) and a policy \(\pi\), and suppress
\(\zeta\) in the notation below. Let \(\widehat w_\pi\) be the FORE estimate of
\(w_\pi=\mathrm d\mu_\pi/\mathrm d\nu_b\) after \(K_w\) fitted-ratio
iterations on \(n_w\) transitions. Write
\[
\varepsilon_w(\pi)
:=
\left\|\frac{\widehat w_\pi}{w_\pi}-1\right\|_{2,\mu_\pi}.
\]
Let \(\mathcal W_{\mathrm F}\) be the fitted normalized ratio class. For
nonnegative \(f\) and \(g\), define
\[
D_{\nu}^{\mathrm{gen}}(f\|g)
:=\E_{\nu}[f\log(f/g)-f+g],
\qquad
\varepsilon_{\mathrm{KL}}(\pi)
:=
\inf_{w\in\mathcal W_{\mathrm F}}
D_{\nu_b}^{\mathrm{gen}}(w\|w_\pi),
\]
and let \(\mathfrak r_{n_w,\mathrm{fit}}(\pi)\) denote the critical radius
for the log-ratio and multiplier classes in
\citet[Theorem~4.2]{van2026fitted}.

\begin{theorem}[Fixed-policy FORE rate]
\label{thm:fixed-policy-fore-rate}
Assume that the fitted log-ratio class is closed and convex and that its
centered elements are uniformly bounded. Assume also the density lower bound,
source and one-step coverage, and critical-radius conditions of
\citet[Theorem~4.2 and Lemma~D.8]{van2026fitted}. Then there is a constant
\(C_{\mathrm{FORE}}<\infty\), depending only on the coverage and bounded-log
constants, such that, for every \(\eta\in(0,1)\), with probability at least
\(1-\eta\),
\[
\begin{aligned}
\varepsilon_w(\pi)
\le C_{\mathrm{FORE}}\Bigg[&
\left(\frac{1+\gamma}{2}\right)^{K_w/2}
\left\{
D_{\nu_b}^{\mathrm{gen}}(1\|w_\pi)
\right\}^{1/2}
+
\left\{
\frac{\varepsilon_{\mathrm{KL}}(\pi)}{1-\gamma}
\right\}^{1/2}\\
&+
\frac{\log(en_w)}{1-\gamma}
\left\{
\mathfrak r_{n_w,\mathrm{fit}}(\pi)
+
\sqrt{\frac{\log(1/\eta)}{n_w}}
\right\}
\Bigg].
\end{aligned}
\]
\end{theorem}

The three terms are the finite-iteration, ratio-class approximation, and
statistical errors, respectively. Correct specification means that
\(\log w_\pi\) lies in the fitted log-ratio class up to an
additive constant, in which case
\(\varepsilon_{\mathrm{KL}}(\pi)=0\). This condition is distinct from
\(Q\)-function realizability. For the reference-restart policies used here,
the density lower bound in
Lemma~\ref{lem:positive-occupancy-density} is uniform in temperature.

If the coverage and complexity constants are uniform over a deterministic
policy family \(\mathfrak P\), define
\[
\overline D
:=
\sup_{\pi\in\mathfrak P}
D_{\nu_b}^{\mathrm{gen}}(1\|w_\pi),
\quad
\overline\varepsilon_{\mathrm{KL}}
:=
\sup_{\pi\in\mathfrak P}\varepsilon_{\mathrm{KL}}(\pi),
\quad
\overline{\mathfrak r}_{n_w}
:=
\sup_{\pi\in\mathfrak P}
\mathfrak r_{n_w,\mathrm{fit}}(\pi).
\]
For a common fixed-policy constant \(C_{\mathrm{FORE}}<\infty\), set
\begin{align*}
\delta_{\mathrm{FORE}}
(n_w,\eta,M,K_w;\mathfrak P)
:={}&
C_{\mathrm{FORE}}\Bigg[
\left(\frac{1+\gamma}{2}\right)^{K_w/2}
\overline D^{1/2}
+
\left\{
\frac{\overline\varepsilon_{\mathrm{KL}}}{1-\gamma}
\right\}^{1/2}\\
&\hspace{4.5em}
+
\frac{\log(en_w)}{1-\gamma}
\left\{
\overline{\mathfrak r}_{n_w}
+
\sqrt{\frac{\log(M/\eta)}{n_w}}
\right\}
\Bigg].
\end{align*}

\begin{lemma}[Simultaneous FORE bounds for adaptively selected policies]
\label{cor:fore-stagewise-control}
For \(0\le t<M\), let \(\mathcal H_t\) denote the information available
before the \(t\)th ratio fit. Let the restart distribution \(\zeta_t\) and
policy \(\pi_t\in\mathfrak P\) be measurable with respect to
\(\mathcal H_t\). Conditional on \(\mathcal H_t\), suppose the conditions of
Theorem~\ref{thm:fixed-policy-fore-rate} hold with the common constants used
in \(\delta_{\mathrm{FORE}}\) for the \(K_w\)-step estimate
\(\widehat w_{\zeta_t,\pi_t}\), based on a ratio sample of size \(n_w\).
Then, with probability at least \(1-\eta\),
\[
\max_{0\le t<M}
\left\|
\frac{\widehat w_{\zeta_t,\pi_t}}{w_{\zeta_t,\pi_t}}-1
\right\|_{2,\mu_{\zeta_t,\pi_t}}
\le
\delta_{\mathrm{FORE}}
(n_w,\eta,M,K_w;\mathfrak P).
\]
\end{lemma}

For positive-temperature annealing, apply the lemma with
\(M=J+1\) and
\((\zeta_j,\pi_j)
=(\nu_{\mathrm{ref}},\pi_{\tau_j,\widehat Q_{j,0}})\). For the hard phase, use
\(M=K_{\mathrm{hard}}\) and
\((\zeta_k,\pi_k)
=(\nu_{\mathrm{ref}},\pi_{0,\widehat Q_k^{\mathrm h}})\).
Thus the lemma verifies the ratio-error events in both finite-sample
theorems. If an auxiliary sample of total size \(N_w\) is divided among the
\(M\) ratio fits, take \(n_w=\lfloor N_w/M\rfloor\). Under correct
specification, choosing
\(K_w=O\{(1-\gamma)^{-1}\log n_w\}\) makes the optimization term no larger
than the statistical term.

\subsection{Proofs}

\begin{proof}[Proof of Lemma~\ref{lem:positive-occupancy-density}]
Disintegrate \(\nu_b=\nu_0\otimes\pi_b\). Then
\[
\pi_{\mathrm{ref}}(a\mid s)
\ge e^{-L_{\mathrm{ref}}}\pi_b(a\mid s),
\]
where the inequality uses \(\pi_b(a\mid s)\le1\). The discounted balance
equation therefore gives, for every measurable \(E\),
\[
\mu_\pi(E)
\ge(1-\gamma)\nu_{\mathrm{ref}}(E)
\ge(1-\gamma)e^{-L_{\mathrm{ref}}}\nu_b(E).
\]
Radon--Nikodym differentiation under \(\mu_\pi\ll\nu_b\) gives the stated
density bound.
\end{proof}

\begin{proof}[Proof of Lemma~\ref{prop:ratio-curvature-linear}]
Write \(u:=\widehat w/w-1\) on the support of \(\mu\). Since
\(\widehat w\ge0\), for
\(H=\phi^\top\theta\in\mathcal F-\mathcal F\),
\begin{align*}
\E_{\nu_b}[\widehat wH^2]
&\ge
\E_{\nu_b}[\mathbf 1_{\{w>0\}}\widehat wH^2]\\
&=
\E_\mu[(1+u)H^2]
\\
&=
\|H\|_{2,\mu}^2+\E_\mu[uH^2].
\end{align*}
Cauchy--Schwarz, boundedness of \(\phi\), and
\(\Sigma_\mu\succeq\lambda_\mu I_p\) give
\[
|\E_\mu[uH^2]|
\le
\varepsilon_w(\mu)\{\E_\mu[H^4]\}^{1/2}
\le
\varepsilon_w(\mu)\|H\|_\infty\|H\|_{2,\mu}
\le
\frac{B_\phi}{\sqrt{\lambda_\mu}}
\varepsilon_w(\mu)\|H\|_{2,\mu}^2.
\]
Combining this inequality with the preceding decomposition of
\(\E_{\nu_b}[\widehat wH^2]\), and using
\(\E_{\nu_b}[\widehat wH^2]\ge0\), gives
\[
\E_{\nu_b}[\widehat wH^2]
\ge
\max\left\{0,1-
\frac{B_\phi}{\sqrt{\lambda_\mu}}\varepsilon_w(\mu)
\right\}\|H\|_{2,\mu}^2,
\]
which is the first conclusion of
Lemma~\ref{prop:ratio-curvature-linear}.
Substituting
\(\varepsilon_w(\mu)\le\sqrt{\lambda_\mu}/(2B_\phi)\) gives the
curvature constant \(1/2\), proving the second claim.
\end{proof}

\begin{proof}[Proof of Theorem~\ref{thm:fixed-policy-fore-rate}]
Set \(X=(S,A)\), and use \(P_\pi\) as the transition kernel on the
state--action space. For a normalized ratio \(v\), define
\[
\mathsf B_{\zeta,\pi}v
:=
\frac{\mathrm d\{(1-\gamma)\zeta
+\gamma(v\nu_b)P_\pi\}}{\mathrm d\nu_b}.
\]
By \eqref{eqn:app-discounted-moment},
\(\mathsf B_{\zeta,\pi}w_\pi=w_\pi\). For any two normalized ratios
\(u\) and \(v\), the corresponding updates have the same restart component
\((1-\gamma)\zeta\). Moreover, generalized KL for normalized ratios equals
relative entropy between the associated probability measures. Joint
convexity and data processing under \(P_\pi\) therefore imply
\[
D_{\nu_b}^{\mathrm{gen}}
(\mathsf B_{\zeta,\pi}u\|\mathsf B_{\zeta,\pi}v)
\le
\gamma D_{\nu_b}^{\mathrm{gen}}(u\|v).
\]
This is the population contraction used in
\citet[Theorem~4.2]{van2026fitted}, with the state variable replaced by
\(X\) and the discount factor set to \(\gamma\).

It remains to modify the source term in the empirical update. Replace the
initial-distribution average in the cited proof by an average under the fixed
restart distribution \(\zeta\), corresponding to the right side of
\eqref{eqn:app-discounted-moment}. The same empirical-process bound follows
from the assumed source-coverage and critical-radius conditions. If the
expectation under \(\zeta\) is evaluated exactly, this empirical term is
absent. No additional conditioning term occurs because \(\zeta\) and
\(\pi\) are fixed.

Applying Theorem~4.2 of the cited paper with these substitutions yields
\[
\begin{aligned}
D_{\nu_b}^{\mathrm{gen}}(\widehat w_\pi\|w_\pi)
\le C\Bigg[&
\left(\frac{1+\gamma}{2}\right)^{K_w}
D_{\nu_b}^{\mathrm{gen}}(1\|w_\pi)
+\frac{\varepsilon_{\mathrm{KL}}(\pi)}{1-\gamma}\\
&+
\frac{\log^2(en_w)}{(1-\gamma)^2}
\left\{
\mathfrak r_{n_w,\mathrm{fit}}^2(\pi)
+\frac{\log(1/\eta)}{n_w}
\right\}
\Bigg].
\end{aligned}
\]
Under the density lower bound and bounded-centered-log condition, apply
\citet[Lemma~D.8]{van2026fitted} to obtain
\[
\varepsilon_w^2(\pi)
=
\E_{\nu_b}
\left[\frac{(\widehat w_\pi-w_\pi)^2}{w_\pi}\right]
\le
C D_{\nu_b}^{\mathrm{gen}}(\widehat w_\pi\|w_\pi).
\]
Taking square roots and using subadditivity of the square root gives the
stated bound after absorbing the fixed coverage and bounded-log quantities
into \(C_{\mathrm{FORE}}\).
\end{proof}

\begin{proof}[Proof of Lemma~\ref{cor:fore-stagewise-control}]
Fix \(0\le t<M\). Conditional on \(\mathcal H_t\), the restart distribution
\(\zeta_t\) and policy \(\pi_t\) are fixed. Apply
Theorem~\ref{thm:fixed-policy-fore-rate} conditionally with confidence level
\(\eta/M\), and use the common constants and deterministic envelopes over
\(\mathfrak P\), to obtain
\[
\Pr\left\{
\left.
\left\|
\frac{\widehat w_{\zeta_t,\pi_t}}{w_{\zeta_t,\pi_t}}-1
\right\|_{2,\mu_{\zeta_t,\pi_t}}
>
\delta_{\mathrm{FORE}}(n_w,\eta,M,K_w;\mathfrak P)
\,\right|\,\mathcal H_t
\right\}
\le \frac{\eta}{M}.
\]
Taking expectations and summing these failure probabilities over
\(0\le t<M\) proves the simultaneous bound. No independence across the
\(M\) fitted ratios is required.
\end{proof}

\section{Empirical-Process Tools for Weighted Bellman Regression}
\label{app:weighted-regression-finite-sample-tools}

This appendix collects the empirical-process bounds used for a weighted
least-squares Bellman update. At each update, the iterate and estimated
occupancy ratio are measurable with respect to the pre-update history, while
the regression batch is conditionally independent of that history. The same
argument applies to cross-fitting when the iterate and ratio used on a fold
are estimated without that fold.
\subsection{Empirical-Process Preliminaries}

\subsubsection{Local Maximal Inequality}

Let $O_1,\ldots,O_n \in \mathcal{O}$ be independent random variables. For any
function $f:\mathcal{O} \to \mathbb{R}$, define
\[
\|f\|_{2,n}
:=
\left\{\frac{1}{n}\sum_{i=1}^n\E[f(O_i)^2]\right\}^{1/2}.
\]

For a star-shaped class of functions $\mathcal G$ and a radius
$\delta \in (0,\infty)$, define the localized Rademacher complexity
\[
\mathcal{R}_n(\mathcal G, \delta)
:=
\E\left[
\sup_{\substack{f \in \mathcal G \\ \|f\|_{2,n} \le \delta}}
\frac{1}{n} \sum_{i=1}^n \epsilon_i f(O_i)
\right],
\]
where $\epsilon_i$ are i.i.d.\ Rademacher random variables.

The following local maximal inequality is Lemma~11 of
\citet{foster2023orthogonal}; see also Lemma~11 of
\citet{van2025nonparametric}.

\Needspace{0.52\textheight}
\begin{lemma}[Local maximal inequality]\label{lemma:loc_max_ineq}
Let $\mathcal G$ be a star-shaped class of functions satisfying
$\sup_{f\in\mathcal G} \|f\|_{\infty} \le M$.
Let $\delta = \delta_n \in (0,1)$ satisfy the critical radius condition
$\mathcal{R}_n(\mathcal G,\delta) \le \delta^2$,
and suppose that, as $n\to\infty$,
\[
\frac{1}{\sqrt{n}}\sqrt{\log\log(e/\delta_n)} = o(\delta_n).
\]
Then there exists a universal constant $C>0$ such that, for all $\eta \in (0,1)$,
with probability at least $1 - \eta$, every $f \in \mathcal G$ satisfies
\begin{align*}
&\left|
\frac{1}{n}\sum_{i=1}^n
\bigl(f(O_i) - \E[f(O_i)]\bigr)
\right|
\;\le\;
C\Bigl(
    \delta^2
    + \delta\,\|f\|_{2,n}
\Bigr)
\\
&\qquad+\;
C\Bigl(
    \frac{\sqrt{\log(e/\eta)}\,\|f\|_{2,n}}{\sqrt{n}}
    + \frac{M\,\log(e/\eta)}{n}
\Bigr).
\end{align*}
\end{lemma}
\begin{proof}[Proof of Lemma~\ref{lemma:loc_max_ineq}]
Lemma~11 of \citet{van2025nonparametric} gives a universal constant \(C\)
such that, for every \(u\ge1\), with probability at least \(1-e^{-u^2}\),
every \(f\in\mathcal G\) satisfies
\[
\frac1n\sum_{i=1}^n\{f(O_i)-\E f(O_i)\}
\le
C\left\{
\delta^2+\delta\|f\|_{2,n}
+\frac{u\|f\|_{2,n}}{\sqrt n}
+\frac{Mu^2}{n}
\right\}.
\]
The class \(-\mathcal G\) is star-shaped and has the same envelope and
localized Rademacher complexity as \(\mathcal G\). Apply the one-sided
inequality to both classes with
\(u=\sqrt{\log(2e/\eta)}\), and take a union bound. Absorbing the fixed
factor inside the logarithm into \(C\) yields the stated two-sided bound.
\end{proof}

For the critical radius used below, the side condition in
Lemma~\ref{lemma:loc_max_ineq} is automatic. If
\(\delta_n\ge c\sqrt{\log(en)/n}\) for a fixed \(c>0\) and \(\delta_n<1\),
then \(1/\delta_n\le c^{-1}\sqrt{n/\log(en)}\), and hence
\[
\frac{n^{-1/2}\sqrt{\log\log(e/\delta_n)}}{\delta_n}
\le
C_c\sqrt{\frac{\log\log n}{\log(en)}}\to0.
\]
Thus the lemma applies for all sufficiently large \(n\). For the finitely many
smaller sample sizes, boundedness yields the same form of bound after enlarging
the constant by a factor depending only on \(c\).

The following lemma bounds the localized Rademacher complexity in terms of the
uniform entropy integral and is a direct consequence of Theorem~2.1 of
\citet{van2011local}.

We use the covering number and uniform entropy integral defined in
Section~\ref{sec::finitesample}, with \(\mathcal F\) replaced by a generic
function class \(\mathcal G\).

\begin{lemma}[Localized Rademacher complexity from uniform entropy]
\label{lemma:local_rademacher_entropy}
Let \(\mathcal G\) be a star-shaped class of functions such that
\(\sup_{f\in\mathcal G}\|f\|_\infty \le M\). Then, for every \(\delta>0\),
\[
\mathcal R_n(\mathcal G,\delta)
\;\lesssim\;
\frac{1}{\sqrt n}\,\mathcal J(\delta,\mathcal G)
\left(1+\frac{\mathcal J(\delta,\mathcal G)}{\delta^2 \sqrt n}\right),
\]
where the implicit constant depends only on \(M\).
\end{lemma}
\begin{proof}[Proof of Lemma~\ref{lemma:local_rademacher_entropy}]
Apply Theorem~2.1 of \citet{van2011local} to the
\(L^2(\mathsf P)\)-localized class, after rescaling its envelope to one, and
use Rademacher symmetrization. Expressed in terms of the absolute entropy
integral, the resulting expectation bound is
\[
\mathcal R_n(\mathcal G,\delta)
\le
\frac{C_M}{\sqrt n}\mathcal J(\delta,\mathcal G)
\left\{1+
\frac{\mathcal J(\delta,\mathcal G)}{\delta^2\sqrt n}
\right\},
\]
where \(C_M\) depends only on the envelope bound \(M\). This is the claimed
inequality.
\end{proof}

\begin{lemma}[Empirical-process bound for one weighted least-squares update]
\label{lemma::empiricalprocess}
Assume Condition~\ref{cond::convex}, and let
\(\delta_n\in(0,1)\) satisfy
\[
\mathcal J(\delta_n,\mathcal F)\le \sqrt n\,\delta_n^2,
\qquad
\delta_n\ge \sqrt{\frac{\log(en)}{n}}.
\]
Let $\mathbb P_0$ denote the law of $(S,A,R,S')$ under $\nu_b$ and dynamics $P$,
and let $\mathbb P_n$ be the empirical measure.
Let $B_w\ge1$ and let $\omega$ be any fixed state-action function satisfying
\(\|\omega\|_\infty\le B_w\).
Fix \(\tau\ge0\), with
\(V_{0,Q}(s):=\max_{a\in\mathcal A}Q(s,a)\), and define
\[
\mathcal V_{\tau,\mathcal F}
:=
\{V_{\tau,Q}:Q\in\mathcal F\}.
\]

Then there exists a constant \(C>0\), depending only on
\(B_R,B_{\mathcal F}\), and \(L_{\mathrm{ref}}\), such that, for all
$\eta \in (0,1)$, with probability at least $1 - \eta$,
simultaneously for all \(Q_1,Q_2\in\mathcal F\) and
\(V_1\in\mathcal V_{\tau,\mathcal F}\),
\begin{align*}
&\Bigl|
  (\mathbb P_n - \mathbb P_0)\big[
    \omega\,
    \bigl(Q_1-Q_2\bigr)(S,A)
\\[-0.4em]
&\qquad\qquad\quad
    \times
    \bigl\{
      R
      + \gamma V_1(S')
      - Q_2(S,A)
    \bigr\}
  \big]
\Bigr|
\\[0.4em]
&\;\le\;
C\!\left(
  B_w(\delta_n)^2
  +
  \delta_n\,
  \Bigl\|
    \omega\bigl(Q_1-Q_2\bigr)
  \Bigr\|_{2,\mathbb P_0}
\right)
\\[0.4em]
&\qquad+\;
C\!\left(
  \frac{
    \sqrt{\log(e/\eta)}\,
    \Bigl\|
      \omega\bigl(Q_1-Q_2\bigr)
    \Bigr\|_{2,\mathbb P_0}
  }{\sqrt{n}}
\right)
\\
&\qquad+\;
C\,B_w\frac{\log(e/\eta)}{n}.
\end{align*}
\end{lemma}
\begin{proof}[Proof of Lemma~\ref{lemma::empiricalprocess}]
Fix a bounded weight function $\omega$ with
$\|\omega\|_\infty \le B_w$.
Define the function class
\begin{align*}
\mathcal{G}_\omega
:=
\Bigl\{
g_{Q_1,Q_2,V_1} :\;
&(s,a,r,s') \mapsto
\omega(s,a)\,
\bigl(Q_1(s,a) - Q_2(s,a)\bigr)
\\[-0.4em]
&\times
\bigl\{
  r
  + \gamma V_1(s')
  - Q_2(s,a)
\bigr\}
:\;
Q_1,Q_2 \in \mathcal{F},\; V_1 \in \mathcal V_{\tau,\mathcal F}
\Bigr\}.
\end{align*}
By the bounded-reward setup and Condition~\ref{cond::convex}, there is
\(M_{\mathrm{reg}}:=\max\{B_R,B_{\mathcal F}\}<\infty\) such that every
\(Q \in \mathcal F\), every \(V_1\in\mathcal V_{\tau,\mathcal F}\), and \(R\) are
uniformly bounded by \(M_{\mathrm{reg}}\). Indeed,
\(\min_aQ(s,a)\le V_{\tau,Q}(s)\le\max_aQ(s,a)\) for
\(\tau>0\). For \(\tau=0\), the definition
\(V_{0,Q}(s)=\max_aQ(s,a)\) gives the identical pair of inequalities.
Thus, for any
$g_{Q_1,Q_2,V_1} \in \mathcal{G}_\omega$,
\begin{align*}
|g_{Q_1,Q_2,V_1}(s,a,r,s')|
&\le
|\omega(s,a)|\,
\bigl|Q_1(s,a)-Q_2(s,a)\bigr|
\\
&\qquad{}\times
\bigl(|r| + \gamma|V_1(s')| + |Q_2(s,a)|\bigr)
\\[0.25em]
&\le
C_{\mathrm{env}}(M_{\mathrm{reg}})\,
|\omega(s,a)|\,\bigl|Q_1(s,a)-Q_2(s,a)\bigr|,
\end{align*}
for some constant
\(C_{\mathrm{env}}(M_{\mathrm{reg}})<\infty\). In particular,
$\mathcal{G}_\omega$ has envelope proportional to \(B_w\), and
\begin{equation}
\label{eq:norm-g-vs-wdiff}
\|g_{Q_1,Q_2,V_1}\|_{2,\mathbb P_0}
\;\le\;
C_{\mathrm{env}}(M_{\mathrm{reg}})\,
\|\omega\,(Q_1-Q_2)\|_{2,\mathbb P_0}.
\end{equation}

\paragraph{Entropy control.}
The two critical-radius inequalities give
\(\mathcal J(\delta_n,\mathcal F)\le \sqrt n\,\delta_n^2\) and
\(\delta_n\ge \sqrt{\log(en)/n}\). Fix a finitely supported distribution
\(\mathsf Q_S\) on \(\mathcal S\), and let
\(\overline{\mathsf Q}:=\mathsf Q_S\otimes\pi_{\mathrm{ref}}\), a
distribution on \(\mathcal S\times\mathcal A\). The log-sum-exp map is
\(1\)-Lipschitz in the actionwise maximum norm. Moreover,
\(\pi_{\mathrm{ref}}(a\mid s)\ge e^{-L_{\mathrm{ref}}}\).
Thus, pointwise in \(s\),
\[
|V_{\tau,Q}(s)-V_{\tau,\widetilde Q}(s)|^2
\le
\max_a|Q(s,a)-\widetilde Q(s,a)|^2
\le
e^{L_{\mathrm{ref}}}
\sum_a\pi_{\mathrm{ref}}(a\mid s)
|Q(s,a)-\widetilde Q(s,a)|^2.
\]
Integrating gives
\begin{equation}
\label{eq:empirical-value-entropy-transfer}
\|V_{\tau,Q}-V_{\tau,\widetilde Q}\|_{2,\mathsf Q_S}
\le
e^{L_{\mathrm{ref}}/2}
\|Q-\widetilde Q\|_{2,\overline{\mathsf Q}}.
\end{equation}
Thus the uniform entropy of \(\mathcal V_{\tau,\mathcal F}\) is controlled by
that of \(\mathcal F\), with a radius rescaling uniform over
\(\tau\ge0\). To make the product-class step explicit, let
\(\mathsf P_X\) and \(\mathsf P_{S'}\) be the \((S,A)\)- and
\(S'\)-marginals of a finitely supported law \(\mathsf P\) on
\((S,A,R,S')\). Boundedness and \(\|\omega/B_w\|_\infty\le1\) give
\[
\begin{aligned}
&\|g_{Q_1,Q_2,V}-g_{\widetilde Q_1,\widetilde Q_2,\widetilde V}
\|_{2,\mathsf P}\\
&\qquad\le C
\left\{
\|Q_1-\widetilde Q_1\|_{2,\mathsf P_X}
+\|Q_2-\widetilde Q_2\|_{2,\mathsf P_X}
+\|V-\widetilde V\|_{2,\mathsf P_{S'}}
\right\}.
\end{aligned}
\]
Consequently,
\[
N(C\varepsilon,\mathcal G_{\omega/B_w},L^2(\mathsf P))
\le
N(\varepsilon,\mathcal F,L^2(\mathsf P_X))^2
N(\varepsilon,\mathcal V_{\tau,\mathcal F},L^2(\mathsf P_{S'})).
\]
Equation~\eqref{eq:empirical-value-entropy-transfer} and the bounded-product covering inequality
\citep[Theorem~2.6.18]{van1996weak} therefore give a
structural constant
\(C_{\mathrm{ent}}=C_{\mathrm{ent}}(M_{\mathrm{reg}},L_{\mathrm{ref}})\)
such that
\[
\mathcal J(\delta,\mathcal G_{\omega/B_w})
\le
C_{\mathrm{ent}}\mathcal J(C_{\mathrm{ent}}\delta,\mathcal F),
\qquad \delta>0.
\]

Let
\(\mathcal G_{\omega/B_w}^\circ
:=\{tg:g\in\mathcal G_{\omega/B_w},\ t\in[0,1]\}\), and let \(M_G\) be its
structural envelope. Discretizing \(t\) on a grid of mesh
\(\varepsilon/M_G\) gives, for every discrete \(\mathsf P\),
\[
N(2\varepsilon,\mathcal G_{\omega/B_w}^\circ,L^2(\mathsf P))
\le
\left(1+\left\lceil\frac{M_G}{\varepsilon}\right\rceil\right)
N(\varepsilon,\mathcal G_{\omega/B_w},L^2(\mathsf P)).
\]
Integrating this covering bound gives
\begin{equation}
\label{eq:empirical-star-hull-entropy}
\mathcal J(\delta,\mathcal G_{\omega/B_w}^\circ)
\le
C\left\{
\mathcal J(C\delta,\mathcal F)
+\delta\sqrt{\log(e/\delta)}
\right\}.
\end{equation}
For \(c\ge1\), monotonicity of the covering entropy gives
\[
\mathcal J(c\delta_n,\mathcal F)
\le c\mathcal J(\delta_n,\mathcal F)
\le c\sqrt n\,\delta_n^2.
\]
Moreover,
\(\delta_n\ge\sqrt{\log(en)/n}\) implies
\[
\delta_n\sqrt{\log(e/\delta_n)}
\le C\sqrt n\,\delta_n^2.
\]
Consequently, for
\(\widetilde\delta_n=C_{\mathrm{rad}}\delta_n\),
\eqref{eq:empirical-star-hull-entropy} gives
\[
\mathcal J(\widetilde\delta_n,
\mathcal G_{\omega/B_w}^\circ)
\le
\frac{C}{C_{\mathrm{rad}}}
\sqrt n\,\widetilde\delta_n^2.
\]
Choose the structural constant \(C_{\mathrm{rad}}\ge1\) large enough that
Lemma~\ref{lemma:local_rademacher_entropy} yields
\[
\mathcal R_n(\mathcal G_{\omega/B_w}^\circ,\widetilde\delta_n)
\le
\widetilde\delta_n^2.
\]
If \(\widetilde\delta_n\ge1\), boundedness and
\(\delta_n\ge C_{\mathrm{rad}}^{-1}\) give the conclusion of
Lemma~\ref{lemma::empiricalprocess} after enlarging the structural constant.
It therefore remains to consider
\(\widetilde\delta_n<1\).

\paragraph{Local empirical-process bound.}
Apply Lemma~\ref{lemma:loc_max_ineq} to
\(\mathcal G_{\omega/B_w}^\circ\) at radius \(\widetilde\delta_n\). Every member
of \(\mathcal G_{\omega/B_w}\) belongs to this star hull. Multiplying the resulting
bound by \(B_w\) and using \eqref{eq:norm-g-vs-wdiff} gives
\[
\begin{aligned}
|(\mathbb P_n-\mathbb P_0)g_{Q_1,Q_2,V_1}|
&\le
C\left\{
B_w\widetilde\delta_n^2
+\widetilde\delta_n
\|\omega(Q_1-Q_2)\|_{2,\mathbb P_0}
\right\}\\
&\quad+
C\left\{
\frac{\sqrt{\log(e/\eta)}}{\sqrt n}
\|\omega(Q_1-Q_2)\|_{2,\mathbb P_0}
+B_w\frac{\log(e/\eta)}{n}
\right\}.
\end{aligned}
\]
Substituting
\(\widetilde\delta_n=C_{\mathrm{rad}}\delta_n\) and absorbing
\(C_{\mathrm{rad}}\) into a constant depending only on
\(M_{\mathrm{reg}}\) and \(L_{\mathrm{ref}}\) proves
Lemma~\ref{lemma::empiricalprocess}.
\end{proof}

Lemma~\ref{lem:positive-occupancy-density} ensures that the exact ratio is
strictly positive for every reference-restart occupancy used below.

\begin{lemma}[Weighted Bellman regression with an estimated ratio]
\label{lem:weighted-bellman-regression}
Fix \(\tau\ge0\), \(Q\in\mathcal F\), and a probability measure
\(\mu\ll\nu_b\), and assume its density
\(w:=\mathrm d\mu/\mathrm d\nu_b\) is strictly positive
\(\nu_b\)-almost surely. Suppose
\(\widehat w\) and \(\mu\) satisfy
Condition~\ref{cond:finite-frozen-curvature}, and define
\[
U:=\Pi_{\mathcal F}^{\mu}\mathcal T_\tau Q,
\qquad
\varepsilon_w
:=
\left\|\frac{\widehat w}{w}-1\right\|_{2,\mu}.
\]
Conditional on \(Q\) and \(\widehat w\), let
\(\mathcal D=\{(S_i,A_i,R_i,S_i')\}_{i=1}^m\) be an independent regression
batch and define
\[
\widehat Q^+
\in
\argmin_{f\in\mathcal F}
\frac1m\sum_{i=1}^m
\widehat w(S_i,A_i)
\left\{
R_i+\gamma V_{\tau,Q}(S_i')-f(S_i,A_i)
\right\}^2.
\]
For every \(\eta_{\mathrm{reg}}\in(0,1)\), with conditional probability at
least \(1-\eta_{\mathrm{reg}}\),
\begin{equation}
\label{eq:weighted-bellman-regression}
\|\widehat Q^+-U\|_{2,\mu}
\le
C_{\mathrm{reg}}
\left\{
\delta_m+\sqrt{\frac{\log(e/\eta_{\mathrm{reg}})}{m}}
\right\}
+
C_w\varepsilon_w\|\mathcal T_\tau Q-U\|_\infty,
\end{equation}
where \(C_{\mathrm{reg}}\) and \(C_w\) depend only on the boundedness,
complexity, reference-policy, and restricted-curvature constants in
Lemma~\ref{lemma::empiricalprocess} and
Condition~\ref{cond:finite-frozen-curvature}. In particular, they are
uniform over \(\tau\ge0\) and have no additional dependence on \(\gamma\).
\end{lemma}

\begin{proof}
Condition on \(Q\) and \(\widehat w\), and set
\[
H:=\widehat Q^+-U,
\qquad
M:=\frac{\widehat Q^++U}{2},
\qquad
Y:=R+\gamma V_{\tau,Q}(S').
\]
Let \(\mathbb P_m:=m^{-1}\sum_{i=1}^m
\delta_{(S_i,A_i,R_i,S_i')}\) be the empirical measure of the regression
batch, and write
\(\|h\|_{\widehat w}^2:=\E_{\nu_b}[\widehat w h^2]\).
Because \(U,\widehat Q^+\in\mathcal F\) and \(\mathcal F\) is convex,
\(M\in\mathcal F\); moreover, \(U\) is a feasible comparator in the
empirical regression. The identity
\[
\{Y-\widehat Q^+\}^2-\{Y-U\}^2=-2H\{Y-M\}
\]
and empirical optimality give
\(\mathbb P_m[\widehat wH(Y-M)]\ge0\). The variational inequality for the
\(L^2(\mu)\) projection gives
\(\E_\mu[H\{\mathcal T_\tau Q-U\}]\le0\). Hence, writing
\(u:=\widehat w/w-1\), conditional mean-zero of
\(Y-\mathcal T_\tau Q\) gives
\[
\begin{aligned}
\mathbb P_0[\widehat wH(Y-M)]
&=
\E_\mu[(1+u)H\{\mathcal T_\tau Q-U-H/2\}]\\
&\le
\E_\mu[uH\{\mathcal T_\tau Q-U\}]
-\frac12\|H\|_{\widehat w}^2.
\end{aligned}
\]
Combining this inequality with
\(\mathbb P_m[\widehat wH(Y-M)]\ge0\) yields
\begin{equation}
\label{eq:weighted-bellman-basic}
\frac12\|H\|_{\widehat w}^2
\le
\left|(\mathbb P_m-\mathbb P_0)
\{\widehat wH(Y-M)\}\right|
+
\left|\E_\mu[uH\{\mathcal T_\tau Q-U\}]\right|,
\end{equation}

Apply Lemma~\ref{lemma::empiricalprocess} with
\((Q_1,Q_2,V_1)=(\widehat Q^+,M,V_{\tau,Q})\) and confidence level
\(\eta_{\mathrm{reg}}\). Because \(\widehat Q^+-M=H/2\) and
\(\widehat w^2\le B_w\widehat w\), its conclusion bounds the empirical term
in \eqref{eq:weighted-bellman-basic} by
\[
C\{d_{\mathrm{reg}}^2+
d_{\mathrm{reg}}\|H\|_{\widehat w}\},
\qquad
d_{\mathrm{reg}}
:=
\delta_m+\sqrt{\frac{\log(e/\eta_{\mathrm{reg}})}{m}}.
\]
Cauchy--Schwarz and Condition~\ref{cond:finite-frozen-curvature} bound the
second term in \eqref{eq:weighted-bellman-basic} by
\[
\underline\lambda_{\mathcal F}^{-1/2}
\varepsilon_w\|\mathcal T_\tau Q-U\|_\infty
\|H\|_{\widehat w}.
\]
Substitution into \eqref{eq:weighted-bellman-basic}, followed by
\(2ab\le a^2/2+2b^2\), gives
\[
\|H\|_{\widehat w}
\le
C\{d_{\mathrm{reg}}
+\varepsilon_w\|\mathcal T_\tau Q-U\|_\infty\}.
\]
Applying the restricted-curvature inequality once more proves
\eqref{eq:weighted-bellman-regression}.
\end{proof}

\section{Finite-Sample Proofs for Positive-Temperature Annealing}
\label{app:finite-sample-annealing}

The proof of Theorem~\ref{thm:finite-sample-annealing} separates a conditional
one-update bound, its fixed-temperature consequence, and propagation along the
geometric schedule.

\subsection{Conditional One-Step Bound}

We analyze the positive-temperature procedure in
Section~\ref{sec:occupancy-reweighted-fqi} with the
schedule and statistical notation of Sections~\ref{sec:population-annealing}
and~\ref{sec::finitesample}. At update \((j,k)\), condition on the
history before the regression batch \(\mathcal D_{j,k}\) is used. The
fresh-batch construction makes
\((\widehat Q_{j,0},\widehat w_j,\widehat Q_{j,k})\) fixed under this
conditioning and leaves \(\mathcal D_{j,k}\) independent of that history.
The foldwise interpretation is the cross-fitting construction described in
Section~\ref{sec:occupancy-reweighted-fqi}; unrestricted reuse
of the full sample is not covered by this conditional argument.

The proof uses the population restricted-curvature bound in
Condition~\ref{cond:finite-frozen-curvature}; it does not require an empirical
Gram lower bound on each regression batch. The entropy, critical radius, and
ratio-error event are those defined in Section~\ref{sec::finitesample}.
For the one-step bounds below, abbreviate
\[
\varepsilon_{w,j}
:=
\left\|\frac{\widehat w_j}{w_j}-1\right\|_{2,\mu_j}.
\]

\begin{lemma}[One fitted update with fixed stage weights]
\label{lem:finite-stage-frozen-one-step}
Assume Conditions~\ref{cond::convex}, \ref{cond::supnorm}, and
\ref{cond:finite-frozen-curvature}. Let
\(C_{\mathrm{reg}}\) and \(C_w\) be the constants in
Lemma~\ref{lem:weighted-bellman-regression}. Once the common boundedness,
complexity, reference-policy, and restricted-curvature constants are fixed,
these constants do not otherwise depend on \(\tau_j\) or \(\gamma\);
for any \(N_{\mathrm{reg}}\) conditionally independent regression batches
of size \(m\),
the following holds simultaneously on an event of probability at least
\(1-\eta\), with
\[
d_{\mathrm{reg}}(N_{\mathrm{reg}},m,\eta)
:=
\delta_m+\sqrt{\frac{\log(eN_{\mathrm{reg}}/\eta)}{m}}.
\]
Let
\(U_{j,k}:=
\mathcal U_{\tau_j,\widehat Q_{j,0}}(\widehat Q_{j,k})\) and
\(e_{j,k}:=\|\widehat Q_{j,k}-Q_j^\star\|_{2,\mu_j}\). Suppose either the
global bound in Lemma~\ref{prop:global-high-temperature} applies, or
both \(\widehat Q_{j,0}\) and \(\widehat Q_{j,k}\) lie in the
radius-\(R_{\mathrm{loc}}\) neighborhood of
Lemma~\ref{prop:stage-frozen-affine-contraction} and its contraction
modulus is at most one. Then
\begin{equation}
\label{eq:finite-stage-frozen-one-step}
\|\widehat Q_{j,k+1}-U_{j,k}\|_{2,\mu_j}
\le
C_{\mathrm{reg}}d_{\mathrm{reg}}(N_{\mathrm{reg}},m,\eta)
+C_w(1+C_{\mathrm{int}})
\varepsilon_{w,j}(e_{j,k}+\varepsilon_j)^\alpha.
\end{equation}
\end{lemma}

\begin{proof}
Fix \(j,k\), condition on the pre-regression history, and abbreviate
\(\mu:=\mu_j\), \(Q:=\widehat Q_{j,k}\), and \(U:=U_{j,k}\).
Let \(Q_j^\circ\) be a best supremum-norm approximant to
\(Q_j^\star\). By \eqref{eq:stage-frozen-interpolation},
\[
\|Q-Q_j^\star\|_\infty
\le
C_{\mathrm{int}}(e_{j,k}+\varepsilon_j)^\alpha+\varepsilon_j.
\]
Equation~\eqref{eq:stage-frozen-global-contraction} at stage zero and
\eqref{eq:stage-frozen-affine-contraction} at every local stage, together
with the assumed upper bound of one on their contraction moduli, give
\[
\|U-Q_j^\star\|_{2,\mu}
\le e_{j,k}+2\varepsilon_j.
\]
Hence
\[
\|U-Q_j^\circ\|_{2,\mu}
\le e_{j,k}+3\varepsilon_j
\le3(e_{j,k}+\varepsilon_j).
\]
The supremum-norm contraction of \(\mathcal T_{\tau_j}\),
\eqref{eq:stage-frozen-interpolation} applied to
\(Q-Q_j^\circ\) and
\(U-Q_j^\circ\), and
\(\|Q_j^\circ-Q_j^\star\|_\infty=\varepsilon_j\)
give the explicit chain
\[
\begin{aligned}
\|\mathcal T_{\tau_j}Q-U\|_\infty
&\le
\|\mathcal T_{\tau_j}Q-\mathcal T_{\tau_j}Q_j^\star\|_\infty
+\|U-Q_j^\circ\|_\infty
+\varepsilon_j\\
&\le
\gamma\{C_{\mathrm{int}}(e_{j,k}+\varepsilon_j)^\alpha
+\varepsilon_j\}
+C_{\mathrm{int}}\{3(e_{j,k}+\varepsilon_j)\}^\alpha
+\varepsilon_j.
\end{aligned}
\]
After absorbing numerical constants, this proves
\begin{equation}
\label{eq:finite-frozen-residual-localization}
\|\mathcal T_{\tau_j}Q-U\|_\infty
\le C(1+C_{\mathrm{int}})(e_{j,k}+\varepsilon_j)^\alpha.
\end{equation}
The preceding inequalities apply directly when
\(e_{j,k}+\varepsilon_j\le1\); when this sum exceeds one, uniform
boundedness of \(\mathcal F\), the rewards, and the Bellman images gives the
same inequality after increasing \(C\).
For every \(\tau>0\),
\[
\min_a Q(s,a)
\le V_{\tau,Q}(s)\le
\max_a Q(s,a),
\qquad
|V_{\tau,Q}(s)-V_{\tau,Q'}(s)|
\le\|Q-Q'\|_\infty.
\]
Therefore the envelope and Lipschitz constants are uniform in
\(\tau>0\). The only reference-restart interpolation factor in residual
localization is the displayed \(1+C_{\mathrm{int}}\); no constant depends on a
positive lower temperature bound.

Apply Lemma~\ref{lem:weighted-bellman-regression} conditionally with
\(\eta_{\mathrm{reg}}=\eta/N_{\mathrm{reg}}\), and combine its conclusion with
\eqref{eq:finite-frozen-residual-localization}. A union bound over the
\(N_{\mathrm{reg}}\) batches gives, simultaneously,
\[
\|\widehat Q_{j,k+1}-U_{j,k}\|_{2,\mu_j}
\le
C_{\mathrm{reg}}d_{\mathrm{reg}}(N_{\mathrm{reg}},m,\eta)
+
C_w(1+C_{\mathrm{int}})
\varepsilon_{w,j}(e_{j,k}+\varepsilon_j)^\alpha.
\]
This is \eqref{eq:finite-stage-frozen-one-step}.
\end{proof}

\subsection{A Fixed Positive Temperature}

The one-step bound has the following fixed-temperature consequence. Fix
\(\tau>0\), let
\(Q^{\mathrm{init}}=\widehat Q_0\), and write
\[
\mu
:=\mu_{\tau,Q^{\mathrm{init}}},
\qquad
w
:=\frac{\mathrm d\mu}{\mathrm d\nu_b}.
\]
The ratio \(w\) is estimated once by \(\widehat w\) and held
fixed across the \(K\) Bellman regressions. Define
\[
e_k
:=\|\widehat Q_k-Q_\tau^\star\|_{2,\mu},
\qquad
\varepsilon_w
:=
\left\|\frac{\widehat w}{w}-1
\right\|_{2,\mu}.
\]
Fix a deterministic ratio-error bound \(\delta_w>0\).
For \(K\) independent regression batches of size \(m\), set
\[
\delta_{\mathrm{reg}}
:=
C_{\mathrm{reg}}
\left\{
\delta_m+\sqrt{\frac{\log(eK/\eta)}{m}}
\right\},
\qquad
\rho_{\mathrm{fs}}
:=
\frac{1+\rho_{\tau}(R_{\mathrm{loc}})}{2},
\]
where \(0<R_{\mathrm{loc}}<R_\tau^{\max}\). Define
\begin{equation}
\label{eq:finite-frozen-local-error}
\operatorname{err}_{\mathrm{fs}}
:=
\frac{\delta_{\mathrm{reg}}}
{1-\rho_{\mathrm{fs}}}
+
\frac{\{C_w(1+C_{\mathrm{int}})\delta_w\}^{1/(1-\alpha)}}
{(1-\rho_{\mathrm{fs}})^{1/(1-\alpha)}}.
\end{equation}

\begin{theorem}[Local finite-sample convergence with one ratio fit]
\label{thm:finite-frozen-local}
Assume Conditions~\ref{cond::convex}--\ref{cond::supnorm} and
Condition~\ref{cond:finite-frozen-curvature}. Suppose
\[
\max\{e_0,\operatorname{err}_{\mathrm{fs}}\}\le R_{\mathrm{loc}}<R_\tau^{\max},
\qquad
\Pr\{\varepsilon_w\le\delta_w\}
\ge1-\eta.
\]
Then, with probability at least \(1-2\eta\), all \(K\) fitted iterates
remain in the radius-\(R_{\mathrm{loc}}\) local contraction neighborhood and,
for
\(0\le k<K\),
\begin{align*}
e_{k+1}
&\le
\rho_{\tau}(R_{\mathrm{loc}})e_k
+\delta_{\mathrm{reg}}
+C_w(1+C_{\mathrm{int}})\delta_we_k^\alpha\\
&\le
\rho_{\mathrm{fs}}e_k
+(1-\rho_{\mathrm{fs}})\operatorname{err}_{\mathrm{fs}}.
\end{align*}
Consequently, for \(1\le k\le K\),
\begin{equation}
\label{eq:finite-frozen-local-rate}
e_k
\le
\rho_{\mathrm{fs}}^k e_0
+(1-\rho_{\mathrm{fs}}^k)\operatorname{err}_{\mathrm{fs}}
\le
\rho_{\mathrm{fs}}^k e_0+\operatorname{err}_{\mathrm{fs}}.
\end{equation}
\end{theorem}

\begin{proof}[Proof of Theorem~\ref{thm:finite-frozen-local}]
Work on the intersection of the regression event in
Lemma~\ref{lem:finite-stage-frozen-one-step}, applied with
\(N_{\mathrm{reg}}=K\), and the ratio event displayed in
Theorem~\ref{thm:finite-frozen-local}.
This intersection has probability at least \(1-2\eta\).
Condition~\ref{cond::fitted-realizability} gives
\(\varepsilon_\tau=0\). Write
\(\rho:=\rho_{\tau}(R_{\mathrm{loc}})\). Then
Theorem~\ref{thm:positive-stage-frozen-local} and the one-step lemma imply,
whenever \(e_k\le R_{\mathrm{loc}}\),
\[
e_{k+1}
\le
\rho e_k
+\delta_{\mathrm{reg}}
+C_w(1+C_{\mathrm{int}})\delta_we_k^\alpha.
\]
Set \(b:=1-\rho_{\mathrm{fs}}=(1-\rho)/2\) and
\(a_w:=C_w(1+C_{\mathrm{int}})\delta_w\). The bound
\[
a_we_k^\alpha
\le
be_k
+a_w^{1/(1-\alpha)}b^{-\alpha/(1-\alpha)}
\]
follows by maximizing \(a_wx^\alpha-bx\) over \(x\ge0\) and bounding the
resulting multiplier
\((1-\alpha)\alpha^{\alpha/(1-\alpha)}\) by one. The definition
\eqref{eq:finite-frozen-local-error} therefore gives
\[
e_{k+1}
\le
\rho_{\mathrm{fs}}e_k
+(1-\rho_{\mathrm{fs}})\operatorname{err}_{\mathrm{fs}}.
\]
Iteration gives
\[
e_k
\le
\rho_{\mathrm{fs}}^ke_0
+(1-\rho_{\mathrm{fs}}^k)\operatorname{err}_{\mathrm{fs}}
\le
\max\{e_0,\operatorname{err}_{\mathrm{fs}}\}
\le R_{\mathrm{loc}}.
\]
The bound \(e_k\le R_{\mathrm{loc}}\) verifies the hypothesis needed to
apply the one-step recursion at iteration \(k\). Induction therefore proves
invariance and
\eqref{eq:finite-frozen-local-rate}.
\end{proof}

Theorem~\ref{thm:finite-frozen-local} separates geometric decay of the
initial error from regression and ratio-estimation errors. Under
realizability, ratio error is multiplied by the current error through
\(\delta_we_k^\alpha\), whereas regression error is additive. The initializer
determines the fixed occupancy, but introduces no additional term because
\(e_0\le R_{\mathrm{loc}}\) controls the occupancy comparison in
\(\rho_{\tau}(R_{\mathrm{loc}})\). If
\(R_{\mathrm{loc}}=\theta R_\tau^{\max}\) for fixed
\(\theta\in(0,1)\), then
\(\rho_{\tau}(R_{\mathrm{loc}})=\gamma^{(1-\theta^\alpha)/2}\); hence the
contraction factor has no small-temperature dependence. If the boundedness
and curvature constants are uniform in \(\gamma\), then
\[
\operatorname{err}_{\mathrm{fs}}
\lesssim
\frac{\delta_{\mathrm{reg}}}{1-\gamma}
+
\frac{\delta_w^{1/(1-\alpha)}}
{(1-\gamma)^{(1+\alpha/2)/(1-\alpha)}}.
\]
Thus, under polynomial statistical rates, choosing \(K\) logarithmic in the
relevant per-batch sample sizes makes the optimization error no larger than the
statistical error bound, using one ratio estimate for the stage.

\subsection{Proof of the Geometric-Annealing Guarantee}

Set \(b_{\mathrm{ann}}:=1-\rho_{\mathrm{ann}}\) and
\(a_w:=C_w(1+C_{\mathrm{int}})\delta_w\). Young's inequality gives
\begin{equation}
\label{eq:finite-annealing-young}
a_wx^\alpha
\le
b_{\mathrm{ann}}x
+a_w^{1/(1-\alpha)}
b_{\mathrm{ann}}^{-\alpha/(1-\alpha)},
\qquad x\ge0.
\end{equation}

\begin{proof}[Proof of Theorem~\ref{thm:finite-sample-annealing}]
Intersect the empirical-process event in
Lemma~\ref{lem:finite-stage-frozen-one-step}, applied with
\(N_{\mathrm{reg}}=(J+1)K\), and the uniform ratio event in
\eqref{eq:finite-frozen-ratio-guarantee}. A union bound gives probability
at least \(1-2\eta\).

On this event, Lemmas~\ref{prop:global-high-temperature} and
\ref{prop:stage-frozen-affine-contraction},
Lemma~\ref{lem:finite-stage-frozen-one-step}, and subadditivity of
\(x\mapsto x^\alpha\) give, for \(j<J\),
\begin{align*}
e_{j,k+1}
&\le
\bar\rho e_{j,k}
+2\varepsilon_j
+\delta_{\mathrm{reg}}
+a_w(e_{j,k}+\varepsilon_j)^\alpha\\
&\le
\rho_{\mathrm{ann}}e_{j,k}
+2\varepsilon_{\mathrm{path}}
+\delta_{\mathrm{reg}}
+a_w\varepsilon_{\mathrm{path}}^\alpha
+a_w^{1/(1-\alpha)}
b_{\mathrm{ann}}^{-\alpha/(1-\alpha)}\\
&=
\rho_{\mathrm{ann}}e_{j,k}
+(1-\rho_{\mathrm{ann}})
\operatorname{err}_{\mathrm{ann}}(\varepsilon_{\mathrm{path}}).
\end{align*}
At the target,
Lemma~\ref{prop:stage-frozen-affine-contraction} and
Lemma~\ref{lem:finite-stage-frozen-one-step} first give
\begin{align*}
e_{J,k+1}
&\le
\bar\rho e_{J,k}
+2\varepsilon_J
+\delta_{\mathrm{reg}}
+a_w(e_{J,k}+\varepsilon_J)^\alpha\\
&\le
\rho_{\mathrm{ann}}e_{J,k}
+(1-\rho_{\mathrm{ann}})\operatorname{err}_{\mathrm{ann}}(\varepsilon_J).
\end{align*}
The second inequality uses subadditivity of \(x\mapsto x^\alpha\), followed
by \eqref{eq:finite-annealing-young} with \(x=e_{J,k}\), and the definition
of \(\operatorname{err}_{\mathrm{ann}}\) in \eqref{eq:finite-annealing-error-bound-main}.
Because \(\operatorname{err}_{\mathrm{ann}}\) is nondecreasing and
\(\varepsilon_J\le\varepsilon_{\mathrm{path}}\),
\(\operatorname{err}_{\mathrm{ann}}(\varepsilon_J)\le
\operatorname{err}_{\mathrm{ann}}(\varepsilon_{\mathrm{path}})\). Consequently, conditional on
a stage beginning within the radius-\(R_{\mathrm{loc}}\) neighborhood, its recursion
remains valid throughout that stage. It remains to verify this initialization
property at every
positive-temperature stage.

Choose a best approximant
\(Q_j^\circ\) to \(Q_j^\star\). At the end of stage zero,
\[
\|\widehat Q_{0,K}-Q_0^\circ\|_{2,\mu_0}
\le
\rho_{\mathrm{ann}}^KR_{\mathrm{init}}
+(1-\rho_{\mathrm{ann}}^K)
\operatorname{err}_{\mathrm{ann}}(\varepsilon_{\mathrm{path}})
+\varepsilon_{\mathrm{path}}.
\]
At every later intermediate stage, the within-stage recursion gives
\[
\|\widehat Q_{j,K}-Q_j^\circ\|_{2,\mu_j}
\le
\rho_{\mathrm{ann}}^KR_{\mathrm{loc}}
+(1-\rho_{\mathrm{ann}}^K)
\operatorname{err}_{\mathrm{ann}}(\varepsilon_{\mathrm{path}})
+\varepsilon_{\mathrm{path}},
\qquad 1\le j<J.
\]
Apply the decomposition
\[
\widehat Q_{j+1,0}-Q_{j+1}^\star
=
(\widehat Q_{j,K}-Q_j^\circ)
+(Q_j^\circ-Q_j^\star)
+(Q_j^\star-Q_{j+1}^\star).
\]
Because \(\mu_{j+1}\) is a probability measure, its
\(L^2\)-norm is bounded by the supremum norm. Applying
\eqref{eq:stage-frozen-interpolation} to
\(\widehat Q_{j,K}-Q_j^\circ\in\mathcal F-\mathcal F\),
using the definition of \(\varepsilon_{\mathrm{path}}\), and then applying
Lemma~\ref{lem:temperature-path-lipschitz} gives
\begin{align*}
\|\widehat Q_{j+1,0}-Q_{j+1}^\star\|_{j+1}
&\le
C_{\mathrm{int}}
\Bigl\{x_j+(1-\rho_{\mathrm{ann}}^K)
\operatorname{err}_{\mathrm{ann}}(\varepsilon_{\mathrm{path}})
+\varepsilon_{\mathrm{path}}\Bigr\}^\alpha
+\varepsilon_{\mathrm{path}}
\\
&\qquad
+\frac{\gamma L_{\mathrm{ref}}}{1-\gamma}
(\tau_j-\tau_{j+1}),
\end{align*}
with \(x_0=\rho_{\mathrm{ann}}^KR_{\mathrm{init}}\) and
\(x_j=\rho_{\mathrm{ann}}^KR_{\mathrm{loc}}\) thereafter. For
\(c=(\tau_{\mathrm{tar}}/\tau_0)^{1/J}\),
\[
\tau_j-\tau_{j+1}
=\tau_j(1-c)
\le\frac{\tau_0}{J}
\log\left(\frac{\tau_0}{\tau_{\mathrm{tar}}}\right),
\]
where the last inequality uses \(1-e^{-x}\le x\) for \(x\ge0\).
Moreover,
\(x_j\le\rho_{\mathrm{ann}}^K
\max\{R_{\mathrm{init}},R_{\mathrm{loc}}\}\) and
\((1-\rho_{\mathrm{ann}}^K)
\operatorname{err}_{\mathrm{ann}}(\varepsilon_{\mathrm{path}})
\le\operatorname{err}_{\mathrm{ann}}(\varepsilon_{\mathrm{path}})\).
The schedule condition in Theorem~\ref{thm:finite-sample-annealing} therefore
bounds the next-stage error by \(R_{\mathrm{loc}}\). This completes the
warm-start induction, so the within-stage recursions are valid at every
temperature. Finally, unrolling the target-stage recursion through \(k=K\) and
using \((1-\rho_{\mathrm{ann}}^K)
\operatorname{err}_{\mathrm{ann}}(\varepsilon_J)
\le\operatorname{err}_{\mathrm{ann}}(\varepsilon_J)\) gives
\eqref{eq:finite-annealing-target-main}, which proves
Theorem~\ref{thm:finite-sample-annealing}.
\end{proof}

\section{Finite-Sample Hard FQI with Refreshed Occupancy Weights}
\label{app:finite-sample-hard-fqi}

The local population geometry for the hard continuation described in
Section~\ref{sec:occupancy-reweighted-fqi} is established in
Appendix~\ref{app:hard-polishing}. This section adds regression and
occupancy-ratio estimation errors to prove
Theorem~\ref{thm:finite-sample-hard-continuation}. Using the statistical
notation of Section~\ref{sec::finitesample}, write
\(\widehat Q_k:=\widehat Q_k^{\mathrm h}\) and
\(\widehat w_k:=\widehat w_k^{\mathrm h}\). Conditional on the
history before \(\mathcal D_k^{\mathrm h}\) is used, these two objects are
fixed and the fresh regression batch is independent of that history.

Define
\[
e_k
:=
\|\widehat Q_k-Q_{\mathrm{hard}}^\star
\|_{2,\mu_{\mathrm{hard}}},
\qquad
\varepsilon_{w,k}
:=
\left\|\frac{\widehat w_k^{\mathrm h}}{w_k^{\mathrm h}}-1
\right\|_{2,\mu_k^{\mathrm h}}.
\]
Assume the uniform event in \eqref{eq:finite-hard-ratio-event}.
Lemma~\ref{cor:fore-stagewise-control} supplies this simultaneous
ratio-error bound. Condition~\ref{cond:finite-frozen-curvature}
supplies restricted curvature; Lemma~\ref{prop:ratio-curvature-linear}
verifies it from the same ratio error for linear classes. Let
\(p_{\mathrm{rem}}=\alpha(1+\beta/2)>1\), as in
Appendix~\ref{app:hard-polishing}, and recall that
\(\rho_{\mathrm{hard}}=(1+\sqrt\gamma)/2\).

\begin{lemma}[One-step finite-sample hard recursion]
\label{lem:finite-hard-one-step}
Assume Conditions~\ref{cond::convex}, \ref{cond::supnorm},
\ref{cond:finite-frozen-curvature}, and
\ref{cond::hard-realizability}--\ref{cond::hard-soft-margin}. There are
\(R_{\mathrm{hard,fs}}>0\) and finite
constants \(C_{\mathrm{reg,h}}\) and \(C_{w,\mathrm h}\), depending only on
the constants in these conditions. Recall \(\delta_{\mathrm{reg,h}}\) from
Section~\ref{sec::finitesample}.
There is a regression event of probability at least \(1-\eta\),
simultaneously over the \(K_{\mathrm{hard}}\) batches. On the intersection
of this event with the ratio event in
\eqref{eq:finite-hard-ratio-event}, whenever
\(e_k\le R_{\mathrm{hard,fs}}\),
\begin{equation}
\label{eq:finite-hard-one-step}
e_{k+1}
\le
\rho_{\mathrm{hard}}e_k
+\delta_{\mathrm{reg,h}}
+C_{w,\mathrm h}(1+C_{\mathrm{int}})
\varepsilon_{w,k}e_k^\alpha.
\end{equation}
The constants contain no positive-temperature factor.
\end{lemma}

\begin{proof}
Fix \(k\), condition on the pre-regression history, and abbreviate
\[
Q:=\widehat Q_k,\quad
\widehat Q^+:=\widehat Q_{k+1},\quad
\mu:=\mu_k^{\mathrm h},\quad
\widehat w:=\widehat w_k,\quad
U:=\mathcal U_0(Q).
\]
Let
\(e=\|Q-Q^\star\|_{2,\mu_\star}\), using the notation of
Appendix~\ref{app:hard-polishing}. Equations
\eqref{eq:hard-full-local-recursion},
\eqref{eq:hard-norm-transfer-forward}, and
\eqref{eq:hard-exploratory-interpolation} give, for sufficiently small
\(e\),
\[
\begin{aligned}
\|U-Q^\star\|_{2,\mu}
&\le
\sqrt2\|U-Q^\star\|_{2,\mu_\star}
+C_{\mathrm{norm}}e^{p_{\mathrm{norm}}}
\le Ce,\\
\|U-Q^\star\|_\infty
&\le
C_{\mathrm{int}}\|U-Q^\star\|_{2,\mu}^{\alpha}
\le CC_{\mathrm{int}}e^\alpha.
\end{aligned}
\]
The supremum-norm contraction of \(\mathcal T_0\) and
\eqref{eq:hard-exploratory-interpolation} also give
\(\|\mathcal T_0Q-Q^\star\|_\infty
\le\gamma C_{\mathrm{int}}e^\alpha\). Hence
\begin{equation}
\label{eq:finite-hard-residual-localization}
\|\mathcal T_0Q-U\|_\infty
\le C(1+C_{\mathrm{int}})e^\alpha.
\end{equation}

Apply Lemma~\ref{lem:weighted-bellman-regression} conditionally with
\(\tau=0\) and
\(\eta_{\mathrm{reg}}=\eta/K_{\mathrm{hard}}\), and take a union bound over
the hard regression batches. Combining
\eqref{eq:weighted-bellman-regression} and
\eqref{eq:finite-hard-residual-localization}, and enlarging the two
phase-specific constants if necessary, gives
\[
\|\widehat Q^+-U\|_{2,\mu}
\le
\delta_{\mathrm{reg,h}}
+C_{w,\mathrm h}(1+C_{\mathrm{int}})
\varepsilon_{w,k}e^\alpha.
\]
Applying the reverse norm comparison
\eqref{eq:hard-norm-transfer-reverse} to
\(\widehat Q^+-U\in\mathcal F-\mathcal F\) gives
\[
\begin{aligned}
\|\widehat Q^+-U\|_{2,\mu_\star}
&\le
\sqrt2\|\widehat Q^+-U\|_{2,\mu}
+C_{\mathrm{norm}}e^{p_{\mathrm{norm}}}\\
&\le
\delta_{\mathrm{reg,h}}
+C_{w,\mathrm h}(1+C_{\mathrm{int}})
\varepsilon_{w,k}e^\alpha
+Ce^{p_{\mathrm{norm}}},
\end{aligned}
\]
after increasing \(C_{\mathrm{reg,h}}\) and \(C_{w,\mathrm h}\).
The triangle inequality
\[
e_{k+1}
\le
\|U-Q^\star\|_{2,\mu_\star}
+\|\widehat Q^+-U\|_{2,\mu_\star}
\]
together with \eqref{eq:hard-full-local-recursion} now yields
\[
e_{k+1}
\le
\sqrt\gamma\,e+Ce^{p_{\mathrm{rem}}}
+\delta_{\mathrm{reg,h}}
+C_{w,\mathrm h}(1+C_{\mathrm{int}})
\varepsilon_{w,k}e^\alpha,
\]
because \(p_{\mathrm{norm}}>p_{\mathrm{rem}}\) and \(e\le1\). Choose
\(R_{\mathrm{hard,fs}}\le R_{\mathrm{hard}}\wedge1\) so that
\(Ce^{p_{\mathrm{rem}}}\le(\rho_{\mathrm{hard}}-\sqrt\gamma)e\) for
\(e\le R_{\mathrm{hard,fs}}\). This proves
\eqref{eq:finite-hard-one-step}.
\end{proof}

Recall \(\rho_{\mathrm{hard,fs}}\) and
\(\operatorname{err}_{\mathrm{hard}}\) from Section~\ref{sec::finitesample}. Since
\(\rho_{\mathrm{hard,fs}}=(1+\rho_{\mathrm{hard}})/2\),
\[
1-\rho_{\mathrm{hard,fs}}
=\frac{1-\sqrt\gamma}{4}
=\frac{1-\gamma}{4(1+\sqrt\gamma)},
\]
so the finite-sample hard contraction gap is of order \(1-\gamma\).

\begin{theorem}[Finite-sample convergence of hard FQI with refreshed occupancy weights]
\label{thm:finite-hard-fqi}
Under the conditions of Lemma~\ref{lem:finite-hard-one-step}, suppose
\eqref{eq:finite-hard-ratio-event} holds and
\[
\max\{e_0,\operatorname{err}_{\mathrm{hard}}\}
\le R_{\mathrm{hard,fs}}.
\]
Then, with probability at least \(1-2\eta\), for
\(0\le k<K_{\mathrm{hard}}\),
\begin{align}
\label{eq:finite-hard-recursion}
e_{k+1}
&\le
\rho_{\mathrm{hard}}e_k
+\delta_{\mathrm{reg,h}}
+C_{w,\mathrm h}(1+C_{\mathrm{int}})
\delta_{w,\mathrm h}e_k^\alpha\\
&\le
\rho_{\mathrm{hard,fs}}e_k
+(1-\rho_{\mathrm{hard,fs}})\operatorname{err}_{\mathrm{hard}}.
\end{align}
Consequently, for \(0\le k\le K_{\mathrm{hard}}\),
\begin{align}
\label{eq:finite-hard-unrolled}
e_k
&\le
\rho_{\mathrm{hard,fs}}^ke_0
+(1-\rho_{\mathrm{hard,fs}}^k)\operatorname{err}_{\mathrm{hard}},\\
\|\widehat Q_k-Q_{\mathrm{hard}}^\star\|_\infty
&\le C_{\mathrm{int}}e_k^\alpha,\\
\E_{\xi_{\mathrm{hard}}}
\|\pi_{0,\widehat Q_k}(\cdot\mid S)
-\pi_{0,Q_{\mathrm{hard}}^\star}(\cdot\mid S)\|_{\mathrm{TV}}
&\le C_G(2C_{\mathrm{int}})^\beta e_k^{\alpha\beta}.
\end{align}
\end{theorem}

\begin{proof}
Intersect the regression event in
Lemma~\ref{lem:finite-hard-one-step} and the ratio event in
\eqref{eq:finite-hard-ratio-event}. The intersection has probability at
least \(1-2\eta\). On this event, apply the Young bound
\[
ax^\alpha
\le bx+
a^{1/(1-\alpha)}b^{-\alpha/(1-\alpha)}
\]
with
\(a=C_{w,\mathrm h}(1+C_{\mathrm{int}})\delta_{w,\mathrm h}\) and
\(b=(1-\rho_{\mathrm{hard}})/2\). This gives
\[
\rho_{\mathrm{hard}}e_k+ae_k^\alpha
\le
\rho_{\mathrm{hard,fs}}e_k
+a^{1/(1-\alpha)}
(1-\rho_{\mathrm{hard,fs}})^{-\alpha/(1-\alpha)}.
\]
Adding the regression term and using the definition of
\(\operatorname{err}_{\mathrm{hard}}\) proves the second line of
\eqref{eq:finite-hard-recursion}. Its right side is at most
\(R_{\mathrm{hard,fs}}\) whenever
\(\max\{e_k,\operatorname{err}_{\mathrm{hard}}\}\le R_{\mathrm{hard,fs}}\).
Induction therefore keeps every iterate in the local neighborhood, and
unrolling the recursion proves the first line of
\eqref{eq:finite-hard-unrolled}. The interpolation inequality
\eqref{eq:hard-exploratory-interpolation} gives the supremum-norm bound.
Because both tie-broken greedy policies are deterministic, their total
variation distance equals the indicator of the mismatch set
\(D_{\widehat Q_k}\). Therefore
\eqref{eq:hard-greedy-mismatch-set},
Condition~\ref{cond::hard-soft-margin}, and the supremum-norm bound give
\[
\E_{\xi_{\mathrm{hard}}}
\|\pi_{0,\widehat Q_k}(\cdot\mid S)
-\pi_{0,Q_{\mathrm{hard}}^\star}(\cdot\mid S)\|_{\mathrm{TV}}
=\xi_{\mathrm{hard}}(D_{\widehat Q_k})
\le C_G(2C_{\mathrm{int}})^\beta e_k^{\alpha\beta},
\]
which proves the final line of \eqref{eq:finite-hard-unrolled}.

\end{proof}

\begin{proof}[Proof of
Theorem~\ref{thm:finite-sample-hard-continuation}]
Intersect the probability-\((1-2\eta)\) event from the
positive-temperature proof, on which
\eqref{eq:finite-annealing-target-main} holds, with the regression and ratio
events used in Theorem~\ref{thm:finite-hard-fqi}. A union bound gives
probability at least \(1-4\eta\). On this intersection,
\[
\|\widehat Q_{J,K}-Q_{\tau_{\mathrm{sw}}}^\star
\|_{2,\mu_J}
\le e_{\mathrm{sw}}.
\]
Let \(Q_{\mathrm{sw}}^\circ\in\mathcal F\) be a best
supremum-norm approximant to \(Q_{\tau_{\mathrm{sw}}}^\star\). Then
\[
\|\widehat Q_{J,K}-Q_{\mathrm{sw}}^\circ
\|_{2,\mu_J}
\le e_{\mathrm{sw}}+\varepsilon_J.
\]
Applying \eqref{eq:stage-frozen-interpolation} to the difference of these
two functions in \(\mathcal F\), and then using
\(\|Q_{\mathrm{sw}}^\circ
-Q_{\tau_{\mathrm{sw}}}^\star\|_\infty=\varepsilon_J\), gives
\[
\|\widehat Q_{J,K}-Q_{\tau_{\mathrm{sw}}}^\star\|_\infty
\le
C_{\mathrm{int}}(e_{\mathrm{sw}}+\varepsilon_J)^\alpha
+\varepsilon_J.
\]
Lemma~\ref{lem:general-reference-soft-hard-comparison} and
\eqref{eq:finite-hard-main-switch} therefore imply
\[
\|\widehat Q_{J,K}-Q_{\mathrm{hard}}^\star\|_
{2,\mu_{\mathrm{hard}}}
\le
\|\widehat Q_{J,K}-Q_{\mathrm{hard}}^\star\|_\infty
\le R_{\mathrm{hard,fs}}.
\]
Take \(\widehat Q_0=\widehat Q_{J,K}\) in
Theorem~\ref{thm:finite-hard-fqi}. The initialization bound just proved and
the first line of \eqref{eq:finite-hard-unrolled}, together with
\((1-\rho_{\mathrm{hard,fs}}^{K_{\mathrm{hard}}})
\operatorname{err}_{\mathrm{hard}}\le\operatorname{err}_{\mathrm{hard}}\), give
\eqref{eq:finite-hard-main-rate}. This proves
Theorem~\ref{thm:finite-sample-hard-continuation}.
\end{proof}

\end{document}